\documentclass{ieeeaccess}
\usepackage{cite}
\usepackage{amsmath,amssymb,amsfonts}
\usepackage{algorithmic}
\usepackage{graphicx}
\usepackage{textcomp}
\usepackage[hyphens,spaces,obeyspaces]{url}
\def\BibTeX{{\rm B\kern-.05em{\sc i\kern-.025em b}\kern-.08em
    T\kern-.1667em\lower.7ex\hbox{E}\kern-.125emX}}
    
\usepackage{hyperref}
\hypersetup{
    colorlinks=true,
    allcolors=blue
}
\renewcommand{\figureautorefname}{Fig.}
\renewcommand{\tableautorefname}{Table}
\usepackage{booktabs}
\usepackage{soul}

\usepackage{verbatim}
\usepackage[caption=false, subrefformat=parens, labelformat=parens]{subfig}
\usepackage{multirow}
\definecolor{ckgreen}{rgb}{0,0.56,0}

\usepackage{array}
\newcommand{\PreserveBackslash}[1]{\let\temp=\\#1\let\\=\temp}
\newcolumntype{C}[1]{>{\PreserveBackslash\centering}p{#1}}
\newcolumntype{R}[1]{>{\PreserveBackslash\raggedleft}p{#1}}
\newcolumntype{L}[1]{>{\PreserveBackslash\raggedright}p{#1}}
\newcolumntype{X}[1]{>{\PreserveBackslash\centering}m{#1}}
\newcolumntype{E}[1]{>{\PreserveBackslash\raggedleft}m{#1}}
\newcolumntype{K}[1]{>{\PreserveBackslash\raggedright}m{#1}}

\definecolor{spgreen}{rgb}{0,0.56,0}
\definecolor{spblue}{rgb}{0.05, 0.33, 0.58}

\graphicspath{{images/}}

\usepackage{scalerel}
\def\bnzero{\scalerel*{\includegraphics{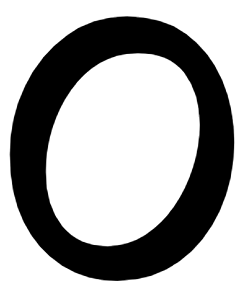}}{C}}
\def\bnone{\scalerel*{\includegraphics{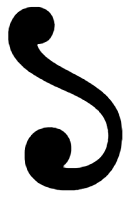}}{C}}

\def\bnthree{\scalerel*{\includegraphics{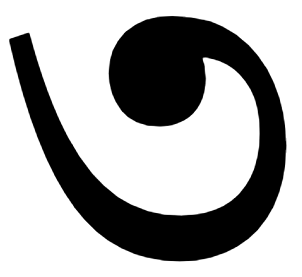}}{C}}
\def\bnfour{\scalerel*{\includegraphics{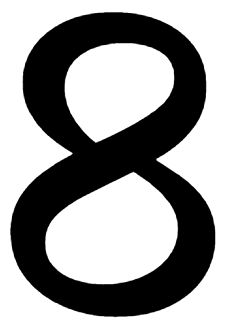}}{C}}
\def\bnfive{\scalerel*{\includegraphics{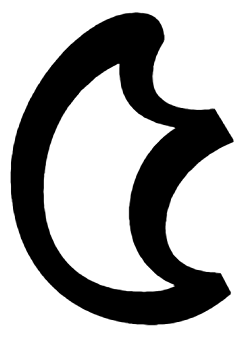}}{C}}
\def\bnsix{\scalerel*{\includegraphics{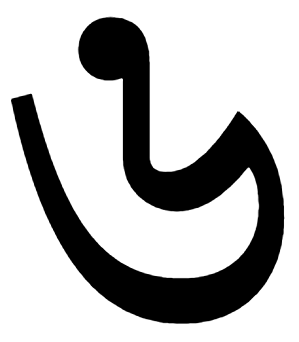}}{C}}

\def\bneight{\scalerel*{\includegraphics{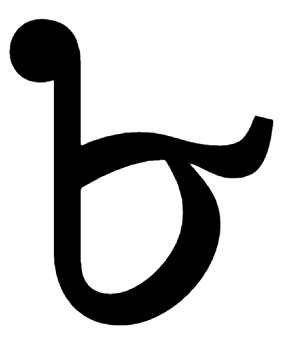}}{C}}
\def\bnnine{\scalerel*{\includegraphics{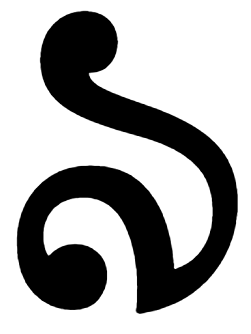}}{C}}

\usepackage{gensymb}
    
\begin{document}
\history{Date of publication yyyy 99, 9999, date of current version zzzz 88, 8888.}
\doi{01.2345/ZXCVBN.6789.ABC}

\title{Two Decades of Bengali Handwritten Digit Recognition: A Survey}
\author{\uppercase{A.B.M. Ashikur Rahman}\authorrefmark{1},
\uppercase{Md. Bakhtiar Hasan}\authorrefmark{1},
\uppercase{Sabbir Ahmed}\authorrefmark{1},
\uppercase{Tasnim Ahmed}\authorrefmark{1},
\uppercase{Md. Hamjajul Ashmafee}\authorrefmark{1},
\uppercase{Mohammad Ridwan Kabir}\authorrefmark{1},
\uppercase{And Md. Hasanul Kabir}\authorrefmark{1} \IEEEmembership{Member, IEEE}
}
\address[1]{Department of Computer Science and Engineering, Islamic University of Technology, Dhaka, Bangladesh}

\markboth
{Rahman \headeretal: Two Decades of Bengali Handwritten Digit Recognition: A Survey}
{Rahman \headeretal: Two Decades of Bengali Handwritten Digit Recognition: A Survey}

\corresp{Corresponding author: A.B.M. Ashikur Rahman (e-mail: \hyperlink{mailto:ashikiut@iut-dhaka.edu}{ashikiut@iut-dhaka.edu}).}

\begin{abstract}
Handwritten Digit Recognition (HDR) is one of the most challenging tasks in the domain of Optical Character Recognition (OCR). Irrespective of language, there are some inherent challenges of HDR, which mostly arise due to the variations in writing styles across individuals, writing medium and environment, inability to maintain the same strokes while writing any digit repeatedly, etc. In addition to that, the structural complexities of the digits of a particular language may lead to ambiguous scenarios of HDR. Over the years, researchers have developed numerous offline and online HDR pipelines, where different image processing techniques are combined with traditional Machine Learning (ML)-based and/or Deep Learning (DL)-based architectures. Although evidence of extensive review studies on HDR exists in the literature for languages, such as English, Arabic, Indian, Farsi, Chinese, etc., few surveys on Bengali HDR (BHDR) can be found, which lack a comprehensive analysis of the challenges, the underlying recognition process, and possible future directions. In this paper, the characteristics and inherent ambiguities of Bengali handwritten digits along with a comprehensive insight of two decades of state-of-the-art datasets and approaches towards offline BHDR have been analyzed. Furthermore, several real-life application-specific studies, which involve BHDR, have also been discussed in detail. This paper will also serve as a compendium for researchers interested in the science behind offline BHDR, instigating the exploration of newer avenues of relevant research that may further lead to better offline recognition of Bengali handwritten digits in different application areas.
\end{abstract}


\titlepgskip=-15pt

\maketitle

\section{Introduction}
\label{sec:introduction}
Communication between two or more human beings is defined as an exchange of information on a particular topic of interest, irrespective of their language, dialect, and cultural differences. Human communication has three basic modes: \textit{written}, \textit{verbal}, or \textit{sign-language}.

\begin{figure*}[htb]
    \centering
    \includegraphics[width=0.75\textwidth]{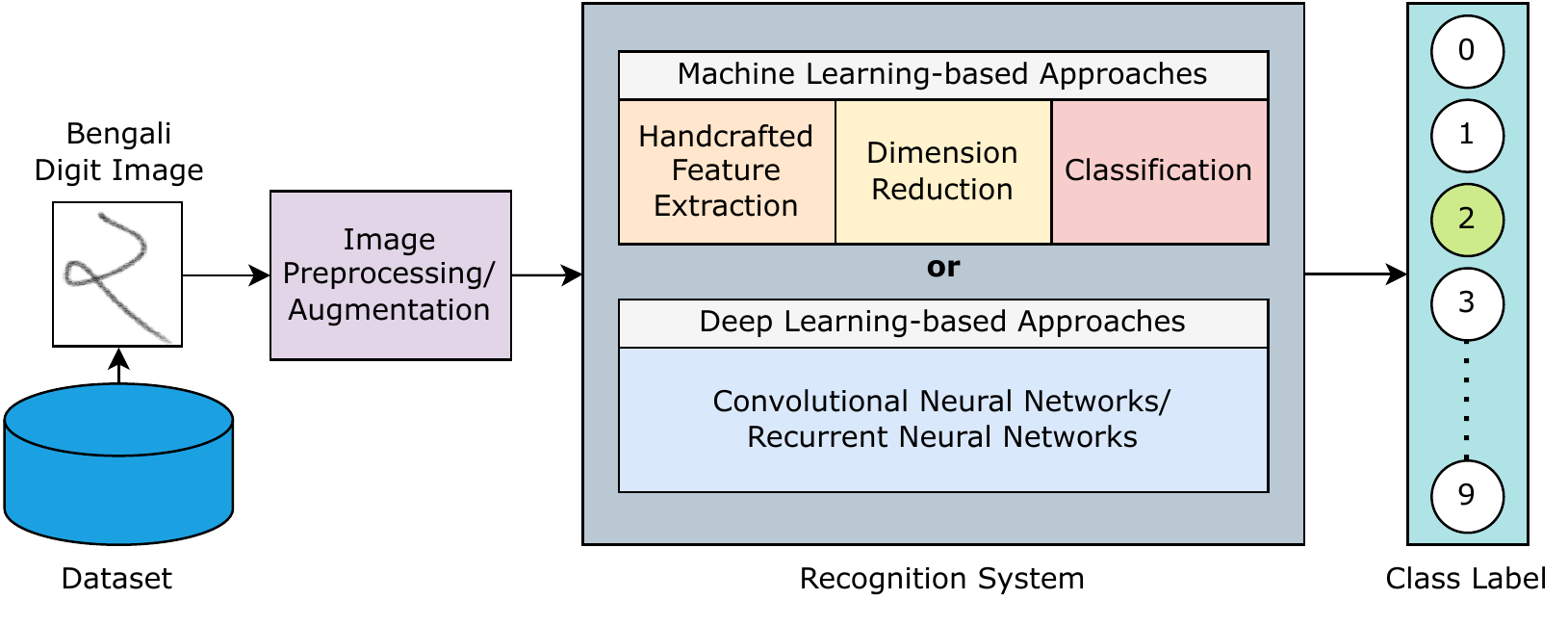}
    \caption{Overview of the Bengali Handwritten Digit Recognition Pipeline}
    \label{fig:pipeline}
\end{figure*}

Written means of communication include different symbols, such as: \textit{letters}, \textit{numerals}, and \textit{special characters}, combinations of which are used for visualizing words and sentences in human speech, conveying meaningful messages similar to communicating \textit{verbally}, or through \textit{braille} and/or \textit{sign language} \cite{1012khatun2021systematic}. Consequently, handwriting has been a ubiquitous means of communication and permanent information storage. However, these symbols are different across languages, and with time, different scripts and symbols belonging to the same language have emerged. For example, the Indian script itself has 9 major regional scripts: Bengali, Gujarati, Gurumukhi, Kannada, Malayalam, Nastaliq, Oriya, Tamil, and Telugu  \cite{0106pal2012handwriting}.

Written communication is mostly facilitated through \textit{letters}, \textit{digits}, \textit{symbols}, and \textit{special characters} in printed form using printing devices or in handwritten form using pen and paper, across different scripts and dialects of any language. However, with the evolution of technology, writing mediums and tools (digital tablets, touch screen PCs, etc.) have also been developed for digitally capturing human handwriting, apart from the pen and paper method. Although the unique shapes of the constituent symbols of different scripts can be maintained exactly when typed in using a computer, it is not the case when they are handwritten. This is because every individual has a unique writing style, which is often dependent on their state of mind, writing environment, writing medium, etc. \cite{0215singh2021online}. Due to these variations and ubiquitous applications, Handwritten Digit Recognition (HDR) has emerged as one of the most important research topics in the field of Optical Character Recognition (OCR) \cite{0172singh2018comprehensive}. The purpose of HDR systems is to encode handwritten digits (0-9 in English) into a computer interpretable format for creating a digital footprint of the information \cite{0179plamondon2000online, 0216silfverberg2007historical}. Even with the advances in computing technologies capable of achieving things that were previously considered impossible, the digit recognition process poses its own challenges. As evident from \cite{0172singh2018comprehensive, 0215singh2021online}, different factors such as the non-uniformity of the digits (size, shape, thickness, orientation, etc.), noise, and most importantly, the variation of writing styles from person to person introduce complexities in the process of HDR.


Researchers have applied HDR systems for recognizing digits of different languages using a combination of image processing tools with Machine Learning (ML) and/or Deep Learning (DL) techniques. Among the traditional ML techniques, Support Vector Machine (SVM), Genetic Algorithms, Decision Tree (DT), Random Forest (RF), $k$-Nearest Neighbor ($k$NN), Hidden Markov Models (HMM), etc. are noteworthy. On the other hand, DL techniques include Recurrent Neural Network (RNN), Convolutional Neural Network (CNN), Long Short–Term Memory (LSTM) network, Generative Adversarial Network (GAN), etc. \cite{0105parvez2013offline,  0104alhelali2017arabic, 0218singh2019wide, 0101baldominos2019survey, 0110memon2020handwritten, 0102cheng2020analysis, 0111nanehkaran2021analysis}.
Researchers from different origins have conducted extensive reviews on how different HDR systems have been used in the recognition of either offline (using scanned images) or online (using real-time input) handwritten digits of different scripts such as: English \cite{0232arica2001overview, 0220patel2015handwritten, 0101baldominos2019survey, 0102cheng2020analysis}, Chinese \cite{0223azmi2013review}, Arabic \cite{ 0223azmi2013review}, Indian Scripts \cite{0106pal2012handwriting, 0223azmi2013review, 0228yadav2018handwritten, 0103kumar2019character}, Farsi \cite{0111nanehkaran2021analysis}, and Roman \cite{0223azmi2013review} summarizing methodologies, tools, feature selection, and results. 
Compared to other languages, developing a robust Bengali Handwritten Digit Recognition (BHDR) pipeline, given the inherent morphological complexity of Bengali digits, remains a challenging task. However, the challenges of the intended task vary across researchers based on the adopted research methodologies, and most importantly, the datasets. To the best of our knowledge, there is a significant lack of review-based works on BHDR. Therefore, with an aim to clearly depict the progress of research in this domain along with providing useful directions for researchers in the future, we investigated the status quo of the existing literature on BHDR published in the last 20 years. 
Based on our findings, our contributions are as follows:
\begin{enumerate}
    \item Highlighted the characteristics of existing datasets on Bengali handwritten digits, exploited in different ML and DL-based frameworks.
    \item Provided an in-depth analysis of the existing data preprocessing and resampling techniques, feature extraction, and classification methods using ML and DL approach suitable for BHDR.
    \item Summarized the methodologies to find out the strengths and weaknesses of the existing BHDR-related works.
    \item Explored the broader scopes of different BHDR systems to provide a guideline for applying them in real-life scenarios.
    \item Created a reference for future researchers, allowing them to identify the challenges, and limitations, and provide research directions in the development of robust BHDR pipelines.
\end{enumerate}

To structure our discussion, we have divided the BHDR pipeline into multiple parts as shown in \figureautorefname~\ref{fig:pipeline}. Based on the division, the remaining sections are structured as follows: Section \ref{sec:characteristics} discusses the complex morphological characters of Bengali digits. Section \ref{sec:datasets} provides an overview of the datasets available for Bengali handwritten digit recognition. Then we go through the status quo of digit recognition research discussing preprocessing (Section \ref{sec:preprocessing}) and augmentation (Section \ref{sec:augmentation}), earlier Machine Learning-based approaches (Section \ref{sec:machine}), and recent shift to Deep Learning (Section \ref{sec:deep}). After that, we assess the studies related to Bengali handwritten digit recognition beyond the conventional classification task in Section \ref{sec:broader}. After in-depth analysis, we identified the research gaps and provided some future directions in section \ref{sec:futureworks}. Finally, Section \ref{sec:future} has drawn a conclusion to the work.


\section{Characteristics of Bengali Digits}\label{sec:characteristics}

Bengali, spoken by around 272 million people, is the second and the sixth most popular language in India and the world, respectively \cite{1005eberhard2022what}. Even though it is an ancient Indo-Aryans Language \cite{0587halder2013individuality}, the representations of different Bengali numerals, as we know of and use them today, are derived from the Hindu-Arabic Numeral System \cite{0589holme2010geometry}.

\begin{table}[htb]
    \centering
    \caption{Different representations of the base digits of Bengali numerals}
    \label{tab:banglaDigits}
    \includegraphics[width=\columnwidth]{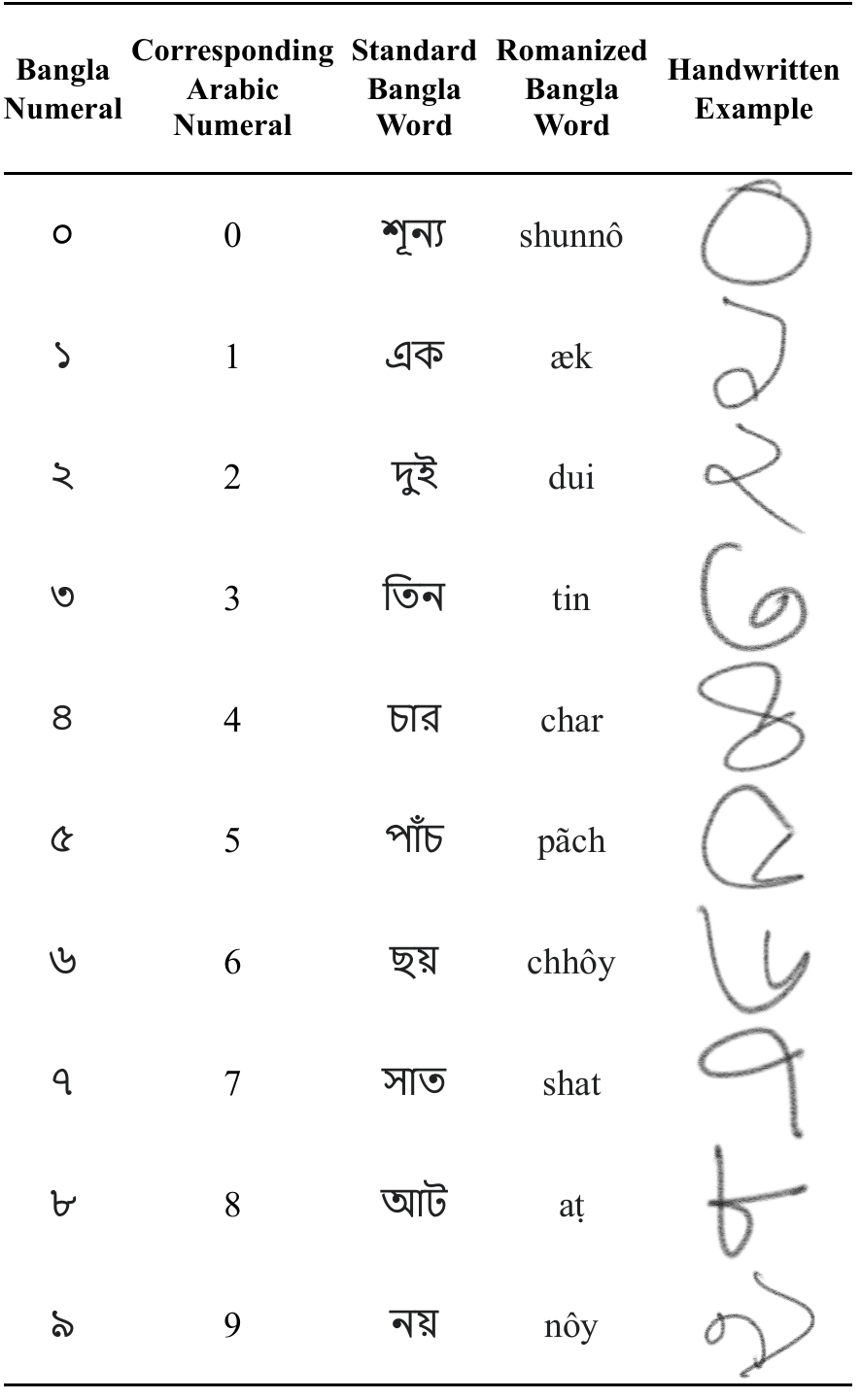}
\end{table}

Compared to the Arabic numerals, representations of the numerals \textit{four} and \textit{eight} in Bengali and Arabic scripts, respectively, are similar in nature \cite{0589holme2010geometry}. The same can be stated for the representation of the numerals \textit{zero}, \textit{two} and \textit{seven} in both scripts \cite{0588roy2004system}. The representations of the digits ``0'' to ``9'' in both scripts have been summarized in \tableautorefname~\ref{tab:banglaDigits} for better comprehension of such similarities.

The style of writing depends on various factors, such as state of mind, writing environment, writing medium, etc. It is intuitive that a person will not always be able to write a particular digit in the same manner. In other words, it is not possible to maintain uniformity in the size, shape, and orientation of a handwritten digit each time, the same or a different person writes it \cite{0587halder2013individuality}. In connection to this, unlike the Arabic numerals, several Bengali numerals share almost similar representations, which, if not written properly, may be misclassified as different numerals. Therefore, it is worth mentioning that multiple ambiguities may appear in a single handwritten Bengali numeral. A few potential factors behind such ambiguities resulting in misclassifications are discussed below:

\subsection{Similar Baseline Skeleton, with Minor Differentiating Strokes}
\begin{figure}[htb]
    \centering
    \subfloat[Typed digits]{\fbox{\includegraphics[height=1.5cm, width=1.3cm]{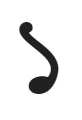}}\hspace{1cm}\fbox{\includegraphics[height=1.5cm, width=1.3cm]{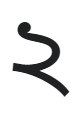}}}
    
    \subfloat[Handwritten digits]{\fbox{\includegraphics[height=1.5cm, width=1.3cm]{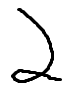}}\hspace{1cm}\fbox{\includegraphics[height=1.5cm, width=1.3cm]{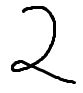}}}
    \caption{Bengali digits 1 and 2 having similar baseline skeleton with minor differentiating strokes}
    \label{fig:simBaseDiffStroke}
\end{figure}

Some of the digits of Bengali numerals share a similar baseline structure, where the difference is only a small stroke. Accidental strokes can make such pairs look alike, which is hard to differentiate, even for humans. For example, the digits \textit{one} and \textit{two} in Bengali have the same skeleton, with the digit \textit{two} having an extra stroke at the bottom, compared to \textit{one}. If these two numerals are not written properly, the BHDR model could suffer from misclassification. Few other pairs of such sort are: \textit{five} and \textit{six}, \textit{one} and \textit{nine}, \textit{zero} and \textit{three}, etc. An illustration showing the ambiguity can be seen in \figureautorefname~\ref{fig:simBaseDiffStroke}.

\subsection{Similar baseline skeleton, with different orientations} 
\begin{figure}[htb]
    \centering
    \subfloat[Typed digits]{\fbox{\includegraphics[height=1.5cm, width=1.3cm]{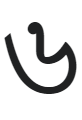}}\hspace{1cm}\fbox{\includegraphics[height=1.5cm, width=1.3cm]{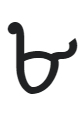}}}
    
    \subfloat[Handwritten digits]{\fbox{\includegraphics[height=1.5cm, width=1.3cm]{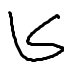}}\hspace{1cm}\fbox{\includegraphics[height=1.5cm, width=1.3cm]{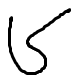}}}
    \caption{Bengali digits 6 and 8 having similar baseline skeleton with minor differentiating orientation}
    \label{fig:simBaseDiffOrientation}
\end{figure}

Real-life handwriting may contain slanted samples oriented in random directions. Even if a digit is properly written, a change in the orientation can make it very confusing for a machine to recognize. For example, the numeral \textit{eight} in Bengali, may be considered to share a similar baseline skeleton with the numeral \textit{six}. If the numeral \textit{six} is written in a way that appears to be rotated, it might lead to misclassification. An illustration showing the ambiguity can be seen in \figureautorefname~\ref{fig:simBaseDiffOrientation}.

\subsection{Incomplete Strokes}
\begin{figure}[htb]
    \centering
    \subfloat[Typed digits]{\fbox{\includegraphics[height=1.5cm, width=1.3cm]{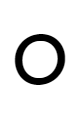}}\hspace{1cm}\fbox{\includegraphics[height=1.5cm, width=1.3cm]{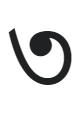}}}
    
    \subfloat[Handwritten digits]{\fbox{\includegraphics[height=1.5cm, width=1.3cm]{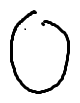}}\hspace{1cm}\fbox{\includegraphics[height=1.5cm, width=1.3cm]{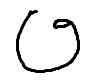}}}
    \caption{Bengali digits 0 and 3 having similarity due to incomplete stroke}
    \label{fig:incompleteStrokes}
\end{figure}

The numeral \textit{three} in Bengali is characterized by a circular shape at the top and a curve emanating from that shape to form an incomplete ellipse. Again, the numeral \textit{zero} is simply a circular shape. If \textit{zero} is written in an incomplete manner, i.e. the circle is not closed and if \textit{three} is written without the initial circular shape, it could be challenging for a classifier to differentiate between these two. The numeral pairs, such as:  \textit{five} and \textit{six}, \textit{zero} and \textit{seven}, etc. may also suffer from the issue of incomplete stroke. An illustration showing the ambiguity can be seen in \figureautorefname~\ref{fig:incompleteStrokes}.

\subsection{Redundant or Extended Strokes}
\begin{figure}[htb]
    \centering
    \subfloat[Typed digits]{\fbox{\includegraphics[height=1.5cm, width=1.3cm]{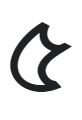}}\hspace{1cm}\fbox{\includegraphics[height=1.5cm, width=1.3cm]{6n.png}}}
    
    \subfloat[Handwritten digits]{\fbox{\includegraphics[height=1.5cm, width=1.3cm]{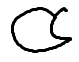}}\hspace{1cm}\fbox{\includegraphics[height=1.5cm, width=1.3cm]{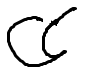}}}
    \caption{Bengali digits 5 and 6 having similarity due to extended stroke}
    \label{fig:extendedStrokes}
\end{figure}

If \textit{three} is written in a manner that the curved line emanating from the circular shape forms a closed loop, the BHDR model might confuse it with \textit{zero}. The numerals \textit{five} and \textit{six} may also suffer from the issue of redundant or extended stroke. An illustration showing the ambiguity can be seen in \figureautorefname~\ref{fig:extendedStrokes}.

\section{Datasets}\label{sec:datasets}
Well-organized datasets are always important for evaluating and benchmarking different methods and they should be consistent with certain quality and quantity standards. Besides, a relevant and well-balanced dataset is a must for smoother and faster training and better recognition \cite{0566roh2021survey}. 
A number of standard datasets are available for Bengali handwritten digits. Some of the datasets are created by taking samples using a specified form from a writer base. Banglalekha-Isolated (BLI) \cite{0052biswas2017banglalekha}, NumtaDB \cite{0011alam2018numtadb}, Ekush \cite{0073rabby2019ekush}, Bengali Handwritten Numerals Dataset (BHaND) \cite{0189chowdhury2016towards}, etc. are examples of this kind. In these datasets, along with the class labels, information regarding the gender, age of the writer, etc. are also available. These datasets are structured and samples are of uniform size and class distribution is balanced.  

Another type of dataset is generated from real-life handwritten documents such as forms, postcards, records, etc. As the sources are heterogeneous, writer information is often not available. However, this type of dataset provides more challenges in digit recognition. Some examples of this kind are ISI Handwritten Bangla Numeral (ISI-HBN) \cite{0044bhattacharya2009handwritten}, CMATERdb \cite{0214das2012statistical, 0155das2012genetic}, etc. 

\begin{figure*}
    \centering
    \subfloat[CMATERdb 3.1.1\label{fig:dataCMATERDB}]{
	\includegraphics[width=.45\textwidth]{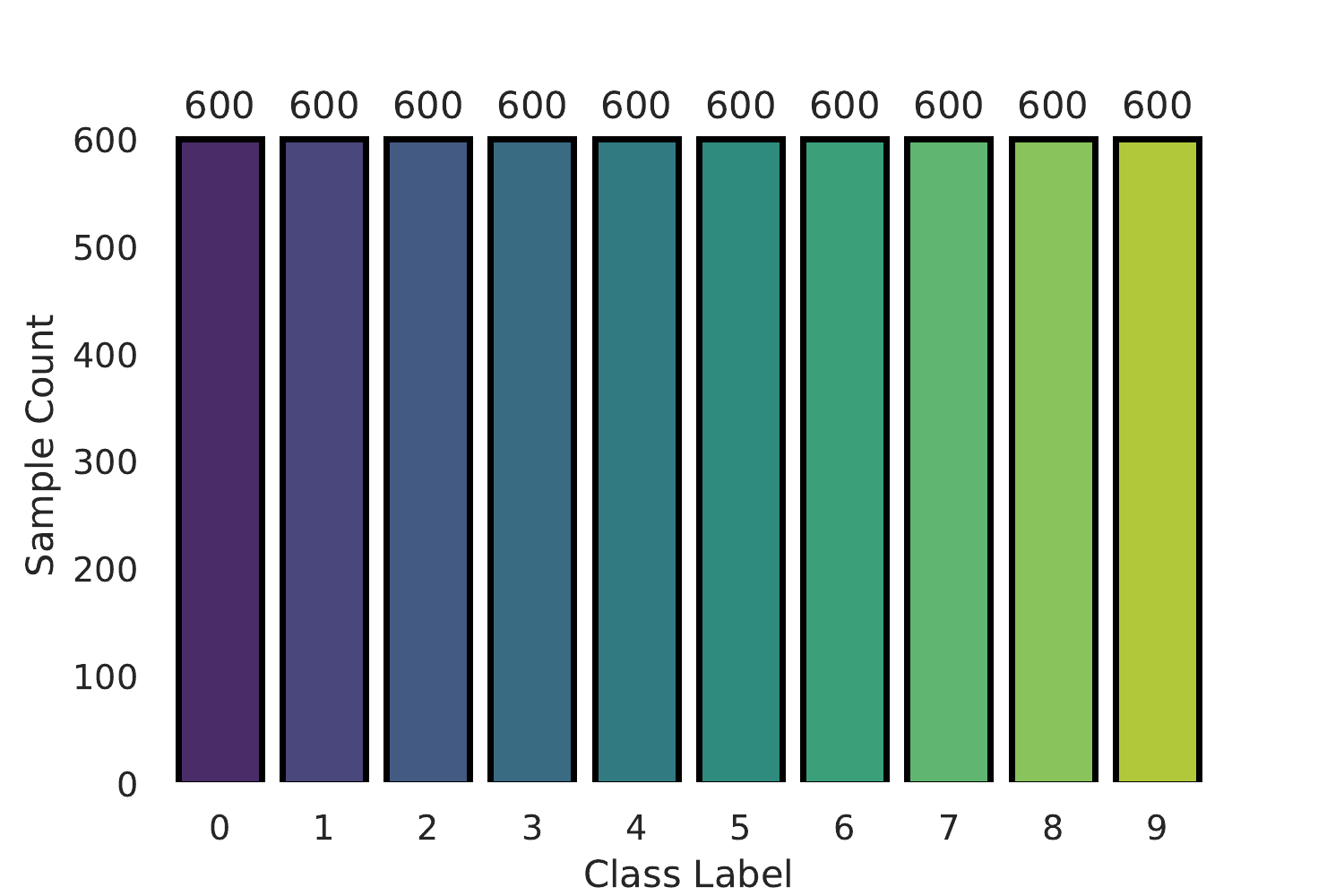}
	}
	\subfloat[ISI Handwritten Bangla Numeral\label{fig:dataISIHBN}]{
	\includegraphics[width=.45\textwidth]{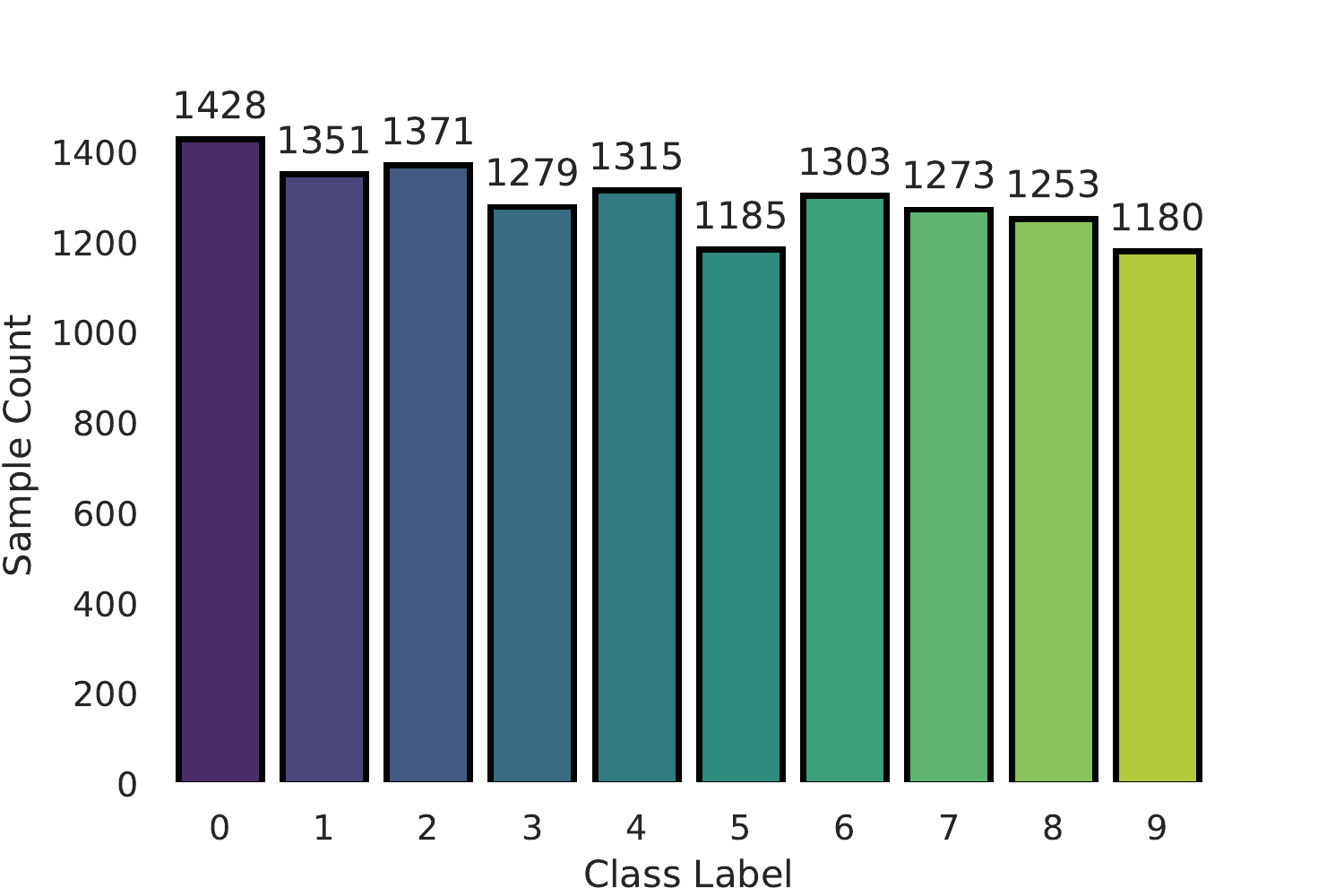}
	}\\
    \subfloat[NumtaDB\label{fig:dataNumtaDB}]{
	\includegraphics[width=.45\textwidth]{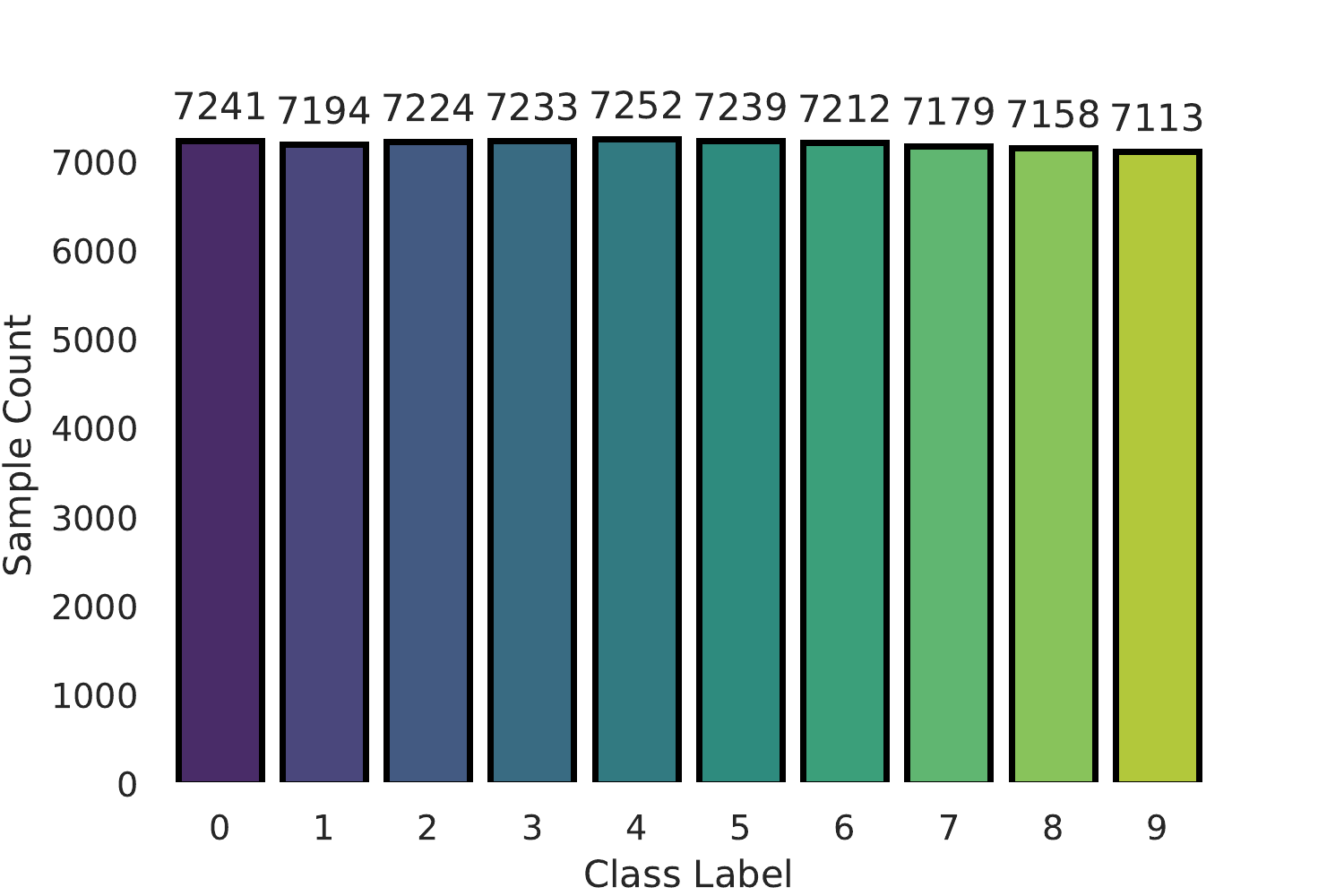}
	}
	\subfloat[Ekush\label{fig:dataekush}]{
	\includegraphics[width=.45\textwidth]{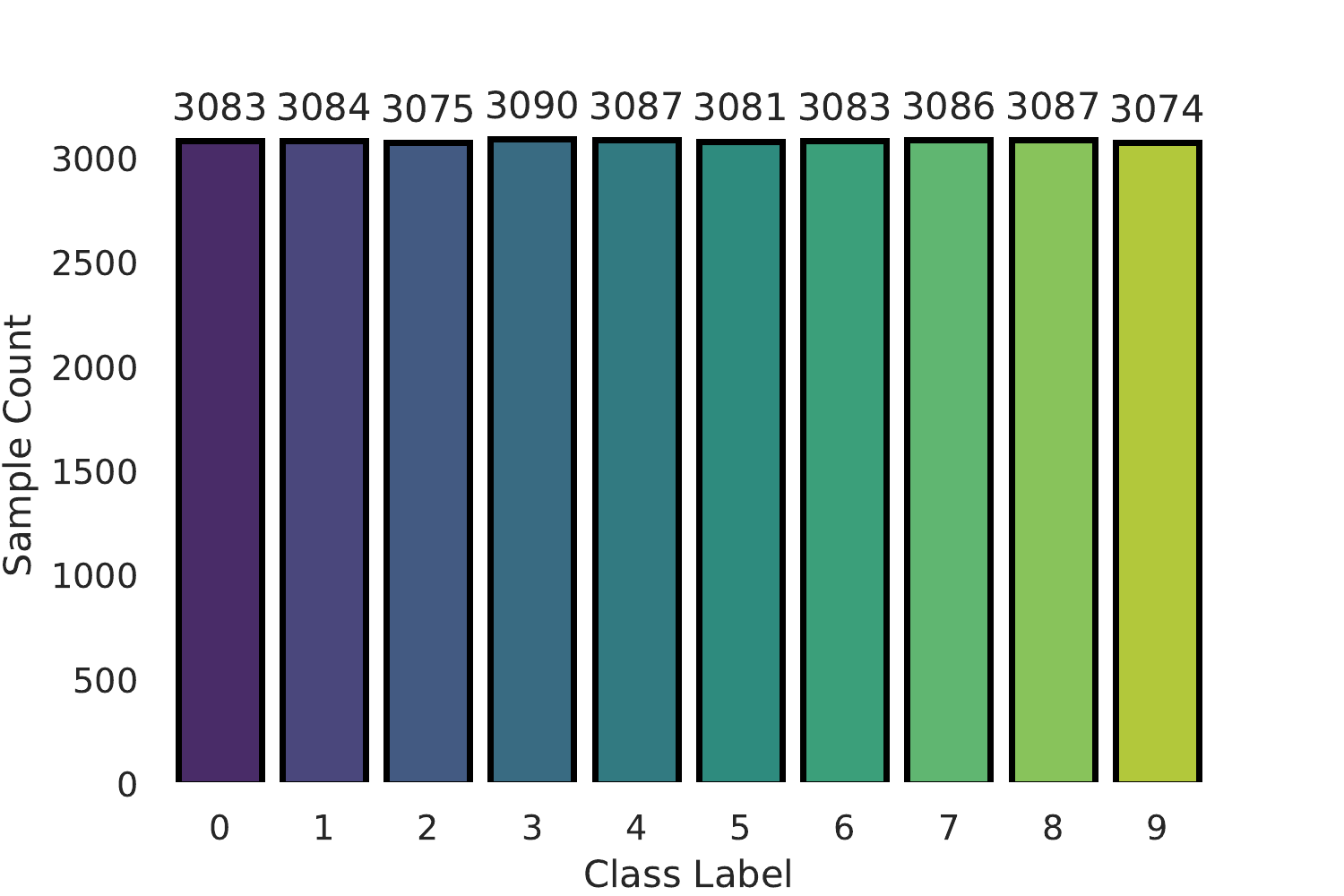}
	}\\
	\subfloat[Banglalekha-Isolated\label{fig:dataBLI}]{
	\includegraphics[width=.45\textwidth]{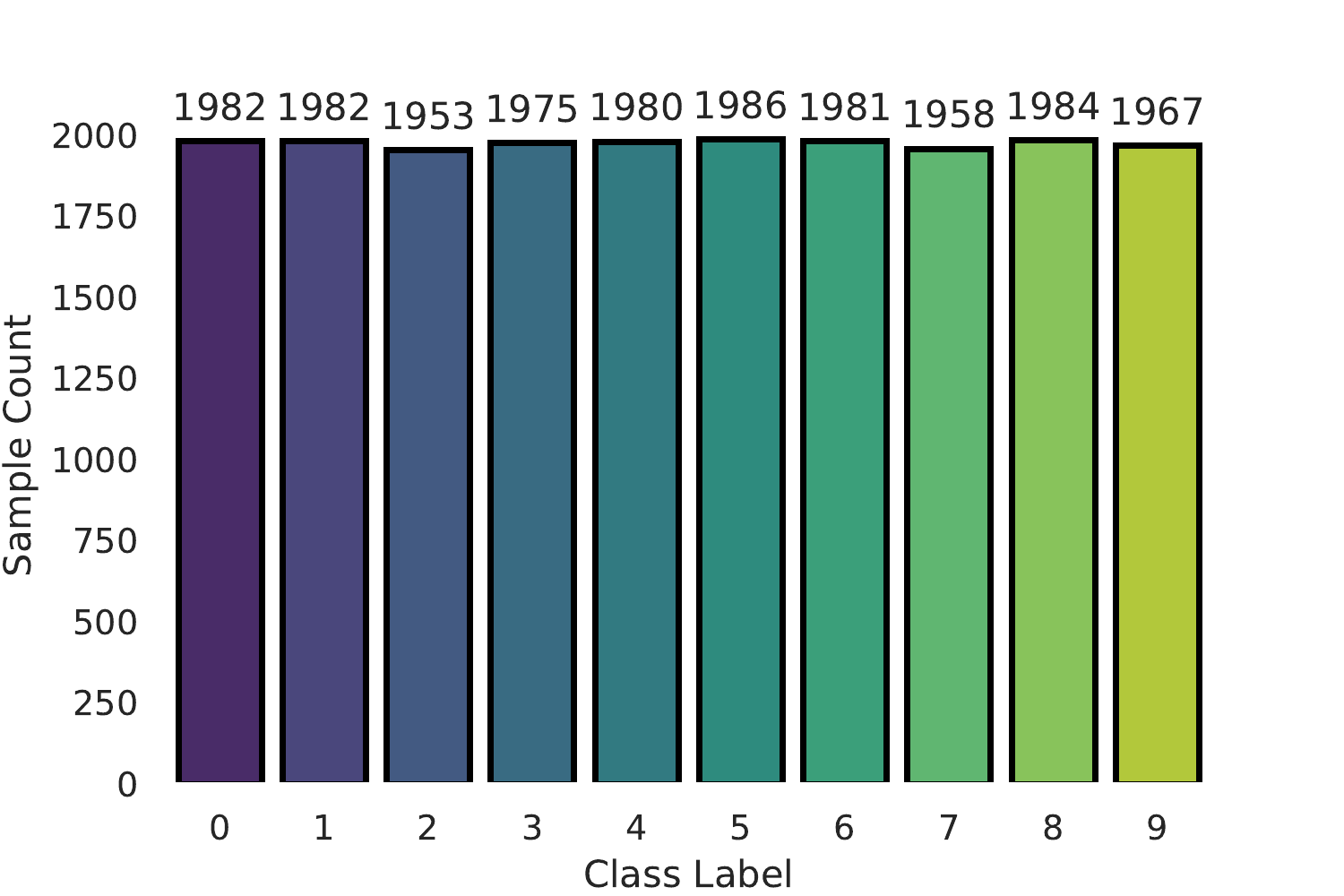}
	}
	\subfloat[Bangla Handwritten Numerals Dataset\label{fig:dataBHAND}]{
	\includegraphics[width=.45\textwidth]{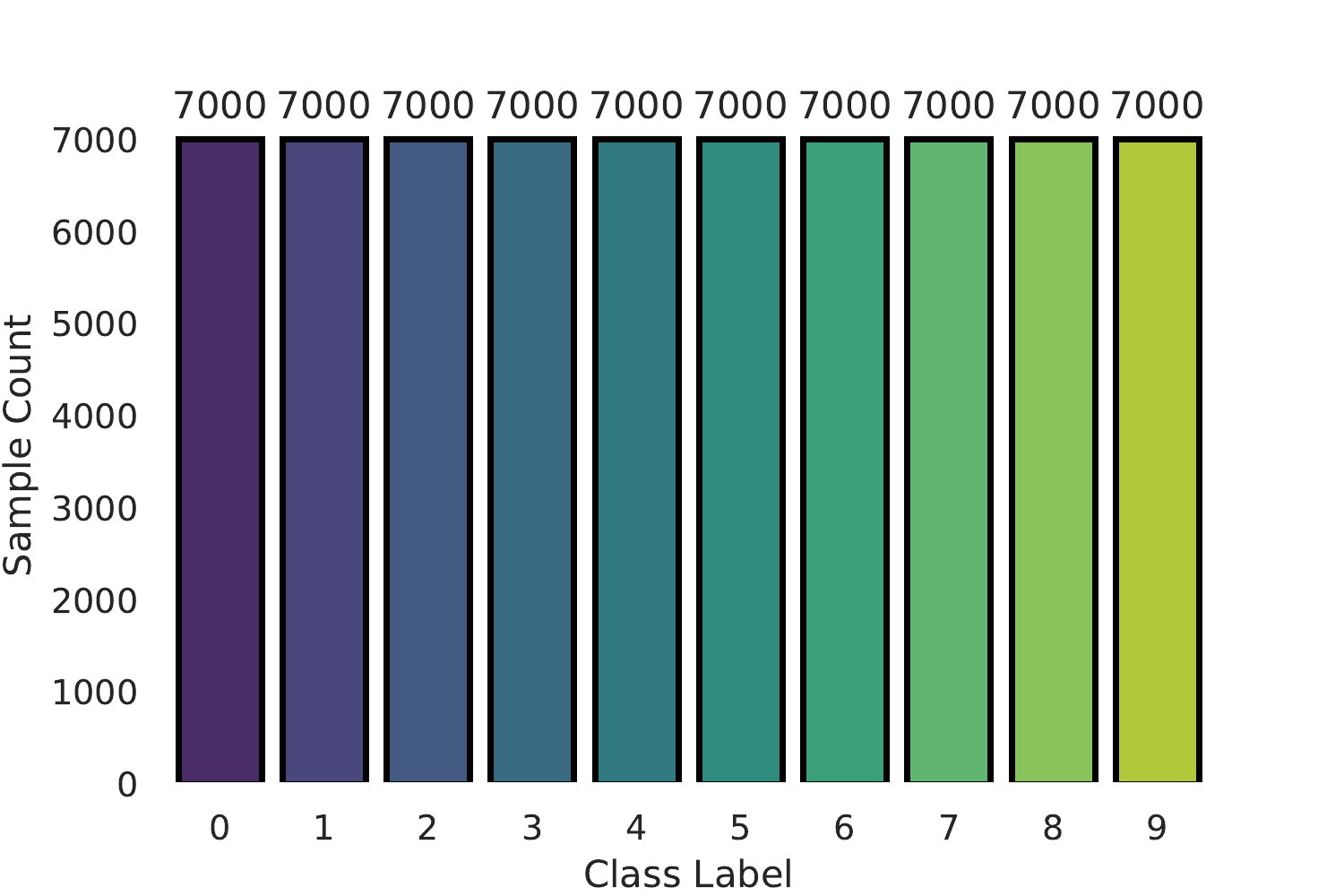}
	}
    \caption{Class distribution of different Bengali handwritten digits datasets}
    \label{fig:classDistribution}
\end{figure*}

The distribution of samples per class for each dataset is shown in \figureautorefname~\ref{fig:classDistribution}.

\begin{table*}[htb]
\centering
  \caption{Summary of Some of the Databases Used in Bengali Handwritten Digit Recognition}
  \label{tab:dataset_summary}
  \begin{tabular}{l c C{1.7cm} c c c c}
    \toprule
    Dataset & Source & Number of images & Writers & Male-Female ratio & Image format & Color format \\
    \midrule
    CMATERdb \cite{0214das2012statistical, 0155das2012genetic} & Real-life & 6000 & -& - & JPEG & RGB\\ 
    ISI Handwritten Bangla Numeral \cite{0044bhattacharya2009handwritten} & Real-life & 23392 & -& - & TIFF & RGB\\
    NumtaDB \cite{0011alam2018numtadb} & Curated & 85000+ & 2742 & Roughly 60-40 & JPEG & RGB \\
    Ekush \cite{0073rabby2019ekush} & Curated & 30688 & 3086& 50-50 & JPEG & RGB\\
    Banglalekha-Isolated \cite{0052biswas2017banglalekha} & Curated &  19748 & Roughly 2000 & 59.4-40.6 & JPEG & RGB\\
    BHaND \cite{0189chowdhury2016towards} & Curated & 70000 & 1750 & 56-44 & JPEG & Gray-scale\\ 
  \bottomrule
\end{tabular}
\end{table*}

\subsection{CMATERdb}
One of the earliest datasets for Bengali handwritten numerals is CMATERdb, which was created at the Center for Microprocessor Applications for Training Education and Research (CMATER) research lab, Jadavpur University, India. The original dataset contains many samples of handwritten documents. Along with different handwritten characters, it contains numerals for 3 different languages: Bengali, Devanagari, and Tamil. Among the different versions of the dataset, CMATERdb 3.1.1 contains Bengali handwritten numerals consisting of a total of 6000 images: 600 RGB color images of size $32\times32$ per class.

\subsection{ISI Handwritten Bangla Numeral (ISI-HBN)}
ISI Handwritten Bangla Numeral dataset was created at the Computer Vision and Pattern Recognition Unit laboratory of the Indian Statistical Institute (ISI), Kolkata. It has both online and offline samples for isolated handwritten digits. The offline samples are labeled and stored in TIFF image format. It has more than 23,000 samples in total, among those, around 19,000 are for training and 4,000 samples are for testing.

\subsection{NumtaDB}
NumtaDB is one of the most popular datasets in Bengali handwritten digit recognition, developed by Bengali-AI in 2018 \cite{0010kamran2018ailearns}. This dataset is a combination of six different databases which are labeled `a' to `f'. It contains more than 85000 images from around 2700 writers both males and females with a ratio of roughly 60\%/40\%. All the sources have separate training and testing partitions in such a way that samples from the same contributor do not exist on both. To make the testing process even more difficult, different challenging image artifacts such as noises, occlusions, rotations, etc. were added.

\subsection{Ekush}
Ekush is one of the largest datasets known so far for Bengali handwritten characters, with 367,018 samples in 122 classes. Among those, more than 30 thousand samples are for digits with an equal male/female writer ratio. This dataset also comes with information about the age, gender, location, and educational status of the writers. This makes it useful for various applications like forensic investigation of detecting these modalities.

\subsection{Banglalekha-Isolated (BLI)}
Although Banglalekha-Isolated is a dataset mainly for isolated Bengali characters, it also contains nearly twenty thousand samples of handwritten digits from 2,000 writers of diverse ages, gender, locations, and educational background. Moreover, the dataset also includes the age and gender labels for each sample, which makes it suitable for the investigation of age and gender influence on handwriting. It also contains an aesthetic quality index for each of the handwriting, marked by several experts. This unique feature provides new directions for investigation.

\subsection{Bangla Handwritten Numerals Dataset (BHaND)}
The Bangla Handwritten Numerals Dataset (BHaND) has 70,000 samples, the same as MNIST \cite{0502lecun1998gradient}, which has a 5:1:1 split for train, validation, and test. One basic difference with MNIST is the size of the images is $32\times32$ instead of $28\times28$. Authors claimed to have done this in order to make the size a multiple of two which may come in handy if down-sampling is needed while working with deep CNN-based architecture \cite{0189chowdhury2016towards}.

The metadata of the discussed datasets are listed in \tableautorefname~\ref{tab:dataset_summary}.




\section{Preprocessing}\label{sec:preprocessing}
\begin{figure}[htb]
    \centering
    \includegraphics[width=\columnwidth]{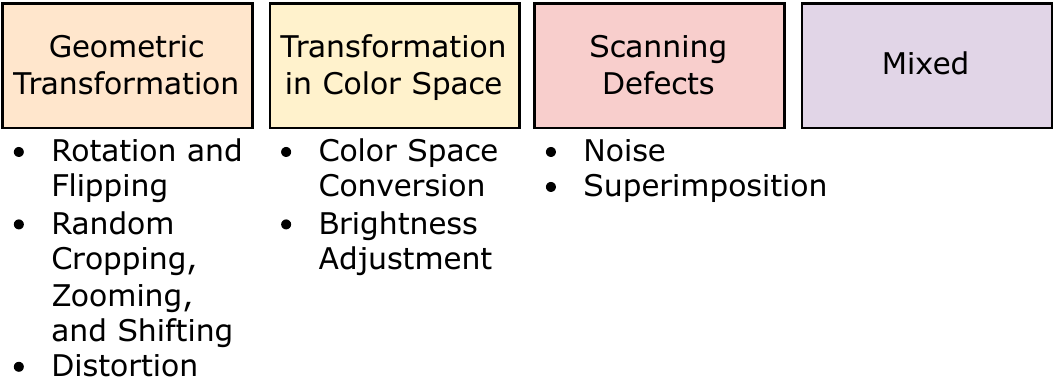}
    \caption{Taxonomy of the preprocessing methods used in Bengali Handwritten Digit Recognition}
    \label{fig:preprocTaxonomy}
\end{figure}

Data Preprocessing, also known as data cleaning, is one of the most common components in a BHDR pipeline with the lowest level of abstraction since both the input and output of this process are image intensity values. The purpose of image preprocessing is to remove distortions and/or enhance image features necessary before further processing without introducing undesirable artifacts. Various types of preprocessing methods seen in BHDR include image transformation, noise correction, geometric and morphological operations, image filtering, segmentation, etc. A taxonomy of the preprocessing techniques is shown in \figureautorefname~\ref{fig:preprocTaxonomy}.

\subsection{Geometric Transformation}

Geometric transformation deals with altering the coordinates of the pixels without changing the intensity values. A set of common geometric transformation techniques used in BHDR is illustrated in \tableautorefname~\ref{tab:pretransform}.

\begin{table}[htb]
    \centering
    \caption{Common geometric transformation techniques used in the existing literature. Original image taken from NumtaDB dataset \cite{0011alam2018numtadb}.}
    \label{tab:pretransform}
    \begin{tabular}{X{2cm} X{2.5cm} X{2.5cm}}
    \toprule
    \textbf{Technique} & \textbf{Original Image} & \textbf{Preprocessed Image}\\\midrule
        Image Resizing & \includegraphics[width=1.66cm]{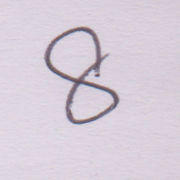} & \includegraphics[width=0.59cm]{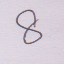}\\
        Padding & \includegraphics[width=1.66cm]{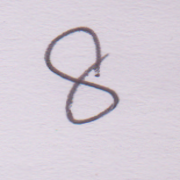} & \includegraphics[width=2.4cm]{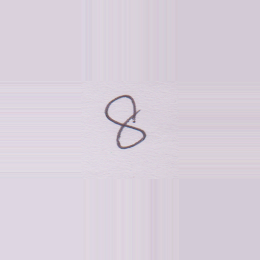}\\
         Cropping & \includegraphics[width=1.66cm]{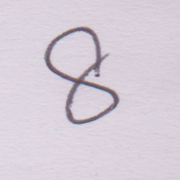} & \includegraphics[width=0.68cm]{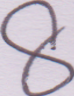}\\
         Slant Correction & \includegraphics[width=1.66cm]{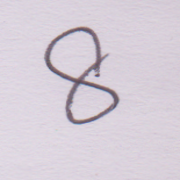} & \includegraphics[width=1.66cm]{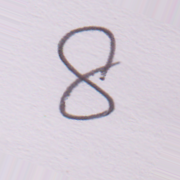}\\
        \bottomrule
    \end{tabular}
\end{table}
\subsubsection{Resizing}
Since vision-based models tend to focus on extracting image textures, reducing the size of the input image does not hurt the performance of the model \cite{0564geirhos2019imagenet}. Rather, these sorts of resizing techniques are often recommended in mini-batch learning to reduce the time required to train the learning model \cite{0567goodfellow2016deep}. Again, some deep neural network-based architectures might have strict requirements for input image dimensions in order to utilize the pretrained weights. For these reasons, in BHDR systems, the digit images are often resized to reduce the input dimensions.

Most of the existing literature on BHDR either use $28\times28$ \cite{0043akhand2015bangla, 0050sharif2016hybrid, 0038akhand2016convolutional, 0029akhand2016convolutional, 0152sharif2017evil, 0008paul2018image, 0006rabby2019bangla, 0012islam2019sankhya, 0166wahid2021classical} or $32\times32$ \cite{0153basu2005handwritten, 0195basu2005mlp, 0194xu2008handwritten, 0045khan2014handwritten, 0157hassan2015handwritten, 0190chowdhury2016towards, 0004alom2017handwritten, 0005aziz2017bangla, 0192rehana2017bangla, 0049choudhury2018handwritten, 0001shawon2018bangla, 0016saha2019bangla, 0020hasan2020new, 0015sikder2020bangla, 0021sufian2020bdnet, 0236singh2021anew} input dimensions. Other dimensions include $16\times16$ \cite{0161wen2007handwritten}, $48\times48$ \cite{0009zunair2018unconventional}, $64\times64$ \cite{0003noor2018handwritten}, and $128\times128$ \cite{0044bhattacharya2009handwritten, 0013mahmud2020deepbanglanet}. The popularity of smaller input dimensions indicates that digit images can be reduced in size without affecting the performance of the learning model. Various interpolation techniques are utilized in this regard to preserve the details \cite{0573siu2012review}. Most of the existing literature uses Bilinear interpolation. Bicubic interpolation is also used in some of the works \cite{0201majumdar2006mlp, 0165surinta2013comparison}.

\subsubsection{Padding}
Padding is used to meet the input dimension requirement for the learning model. Digit images from the same dataset can have different sizes. On the other hand, resizing smaller images to larger ones may introduce unwanted artifacts. To avoid these issues, padding can be performed to increase the size of the smaller images. For rotated images, padding can further ensure the preservation of the original aspect ratio of the digit \cite{0009zunair2018unconventional}. To fill up the additional pixels padded to the digit images, constant values, neighboring values, mirrored values, etc. are used.

\subsubsection{Cropping}
Cropping the input image can help the model to focus only on the digit portion, ignoring background noise.
Again, based on the requirement of input dimensions of different models, the cropped images can be resized afterward. References \cite{0008paul2018image, 0236singh2021anew} achieved this by cropping the digit images based on the largest contour of the digits. On the other hand, some of the existing literature used manual cropping \cite{0038akhand2016convolutional, 0039akhand2016multiple, 0029akhand2016convolutional, 0026akhand2018convolutional, 0037ahmed2019recognizing}. However, this is not feasible for large datasets and even from an application perspective, it should be avoided so that the performance of the model is not dependent on the manual cropping of digits.


\subsubsection{Slant Correction}
Slanted handwritten digits affect the accuracy of the learning model. For example, a slightly tilted six ($\bnsix$) might be considered as eight ($\bneight$) as they have similar baseline skeletons with different orientations. KSC algorithm \cite{0501kimura1993improvements} is used to fix individual slanted digits \cite{0045khan2014handwritten, 0157hassan2015handwritten}.

\subsection{Transformation in Color Space}
Color space transformation modifies the intensity values of the pixels in a digit image. A set of such techniques used in BHDR is illustrated in \tableautorefname~\ref{tab:prenoise}.

\begin{table}[htb]
    \centering
    \caption{Common color space transformation techniques used in the existing literature. Original image taken from NumtaDB dataset \cite{0011alam2018numtadb}.}
    \label{tab:prenoise}
    \begin{tabular}{X{2cm} X{2.5cm} X{2.5cm}}
    \toprule
    \textbf{Technique} & \textbf{Original Image} & \textbf{Preprocessed Image}\\\midrule
    
         Color Inversion & \includegraphics[width=1.66cm]{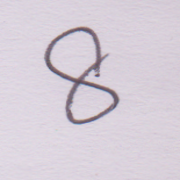} & \includegraphics[width=1.66cm]{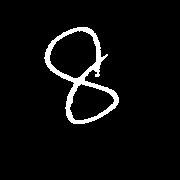}\\
         
        Grayscale Conversion & \includegraphics[width=1.66cm]{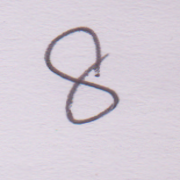} & \includegraphics[width=1.66cm]{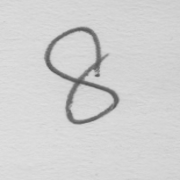}\\
        \bottomrule
    \end{tabular}
\end{table}

\subsubsection{Color Inversion}
Since a majority of the dataset curators collect data from users/annotators on white paper using a black pen, scanned digit images contain white background and dark text in most cases. In 8-bit images, for example, the dark pixels have an intensity value of 0 and the white pixels have an intensity value of 255. That means, all the information regarding digits is stored as 0 values. It may be computationally expensive since models have to work with larger values corresponding to the background.

To avoid this, images are converted to binary form via thresholding techniques (such as Otsu's method \cite{0001shawon2018bangla, 0003noor2018handwritten, 0008paul2018image}, adaptive thresholding \cite{0045khan2014handwritten, 0007mamun2018bangla}) and then inverted, so that the digits have an intensity value of 255 and the background has an intensity value of 0. This also helps get rid of random noises that might be present in the background of the digits. 
Although earlier machine learning-based approaches benefitted from grayscale images \cite{0174chenglin2009new}, having binary intensity does not affect the accuracy of convolutional neural networks \cite{0002hasan2018recognition}. This can be attributed to the invariance of the filters used in the convolutional layers that can easily identify edges regardless of the color space of the image \cite{0002hasan2018recognition}. Furthermore, it can reduce the computational complexity since most of the convolution is performed over pixels with an intensity value of zero \cite{0037ahmed2019recognizing}.

\subsubsection{Grayscale Conversion}
Since the color information of the handwritten digit image has no impact on the performance of the deep learning-based model, the input is often converted to grayscale. This reduces the number of channels in the input image, reducing the training time by decreasing the computational cost \cite{0006rabby2019bangla, 0020hasan2020new, 0035haque2019shonkhanet}. It also ensures the unanimity of the samples collected from multiple sources \cite{0020hasan2020new, 0008paul2018image}. To convert to grayscale images, usually the intensity values for each pixel are considered in the existing literature.

\subsection{Morphological Operations}
Morphological operations were mostly popular with machine learning-based techniques, as these approaches may utilize pixel count to generate features. A set of common morphological operations is illustrated in \tableautorefname~\ref{tab:pregeo}.

\begin{table}[htb]
    \centering
    \caption{Common morphological operations used in the existing literature. Original image taken from NumtaDB dataset \cite{0011alam2018numtadb}.}
    \label{tab:pregeo}
    \begin{tabular}{X{2cm} X{2.5cm} X{2.5cm}}
    \toprule
    \textbf{Technique} & \textbf{Original Image} & \textbf{Preprocessed Image}\\\midrule
        Thinning & \includegraphics[width=1.66cm]{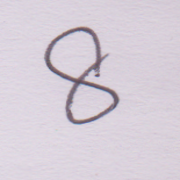} & \includegraphics[width=1.66cm]{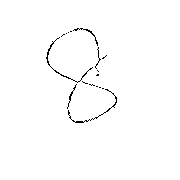}\\
        Thickening & \includegraphics[width=1.66cm]{original.png} & \includegraphics[width=1.66cm]{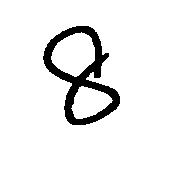}\\
        \bottomrule
    \end{tabular}
\end{table}


\subsubsection{Thinning}
Based on the stroke, the thickness of the handwritten digit can vary. Digits with thicker strokes contain a larger number of pixels, requiring greater computational cost in feature extraction. To alleviate the issue, the thinning operation is performed to make the strokes have the same thickness \cite{0161wen2007handwritten, 0165surinta2013comparison, 0197hashem2014handwritten}. Machine learning-based models that focus on the number of pixels to extract handcrafted features mostly use this technique.

\subsubsection{Thickening}
Thickening is the opposite of the thinning operation. Due to the artifacts introduced during the scanning of the handwritten digits, some unintended gaps might be introduced between strokes. Considering the similarities between Bengali digits, this can hamper the performance of the model. For example, if any gap is introduced in the middle portion of digit four ($\bnfour$), it can be perceived as two separate zeros ($\bnzero$). Reference \cite{0152sharif2017evil} used thickening operation to remove those gaps.

\subsection{Miscellaneous}
Other common preprocessing techniques are illustrated in \tableautorefname~\ref{tab:premisc}.

\begin{table}[htb]
    \centering
    \caption{Other preprocessing techniques used in the existing literature. Original image taken from NumtaDB dataset \cite{0011alam2018numtadb}.}
    \label{tab:premisc}
    \begin{tabular}{X{2cm} X{2.5cm} X{2.5cm}}
    \toprule
    \textbf{Technique} & \textbf{Original Image} & \textbf{Preprocessed Image}\\\midrule
        Noise Removal & \includegraphics[width=1.66cm]{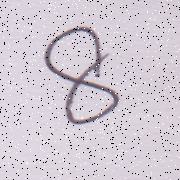} & \includegraphics[width=1.66cm]{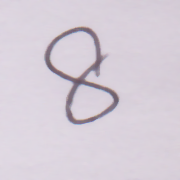}\\
        Deblurring & \includegraphics[width=1.66cm]{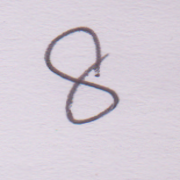} & \includegraphics[width=1.66cm]{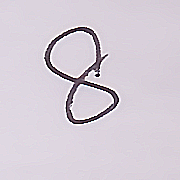}\\
        Coarse Dropout Removal & \includegraphics[width=1.66cm]{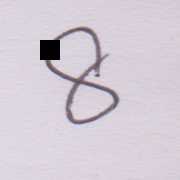} & \includegraphics[width=1.66cm]{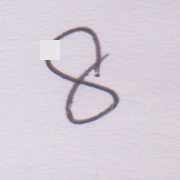}\\
        \bottomrule
    \end{tabular}
\end{table}

\subsubsection{Noise Removal}
Noise can distort images to the point that they become unrecognizable. This can result in the learning models fitting to the noise, reducing the overall performance. Adopting a proper noise removal technique can solve this problem. For example, to remove salt and pepper noise from the handwritten digits, median blur is used \cite{0192rehana2017bangla, 0001shawon2018bangla, 0008paul2018image, 0020hasan2020new}. To remove Gaussian noise, the Gaussian filter seems to be the popular choice \cite{0161wen2007handwritten, 0045khan2014handwritten, 0197hashem2014handwritten, 0157hassan2015handwritten, 0050sharif2016hybrid, 0020hasan2020new, 0042shovon2020recognition, 0166wahid2021classical, 0236singh2021anew}. Finally, in order to remove small artifacts, caused due to resizing the digit images, a combination of morphological opening and closing operations can be performed \cite{0236singh2021anew}.

\subsubsection{Deblurring}
Having sharp curves and edges in the digit portion of the images can help improve the performance of the models that rely on shapes and texture information as features \cite{0001shawon2018bangla, 0042shovon2020recognition}. On the contrary, blurry images can hamper recognition performance. To reduce the blurring effect, a copy of the original image is blurred using Gaussian blur and subtracted from the original image to create an unsharp mask. That mask is again added to the original image to reduce the blurring effect \cite{0001shawon2018bangla}. Another way is to use a sharpening kernel and perform 2D convolution on the image. For example, \cite{0042shovon2020recognition} and \cite{0036islam2018deep} used the Laplacian filter to extract edge information, which is then added to the original image to sharpen blurred samples.

\subsubsection{Removal of Coarse Dropout}
During the scanning of digits, artifacts like coarse dropout can be introduced in images. This can hamper the training process of the learning models by obscuring the digit to be recognized. To remove coarse dropout, contour approximation is used, which detects and converts the dropout pixels to white in order to match the background of the digit \cite{0008paul2018image}.

\subsubsection{Normalization}
Normalization is performed to reduce the effect of illumination differences in images. Considering the maximum and minimum intensity value of pixels in all images, the intensity values are scaled between 0 and 1. It also helps learning models converge faster. Min-max normalization is used to perform this operation \cite{0006rabby2019bangla, 0169shuvo2021mathnet}.

\section{Augmentation}\label{sec:augmentation}
Data augmentation techniques are applied to artificially increase the size of the dataset by creating a modified version of the samples. Considering the huge amount of data required to train deep learning models, various image augmentation techniques can be applied to handwritten digits to generate multiple copies of the same image with slight variation(s) \cite{0575perez2017theEffectiveness}. These techniques also reduce the dependency of the model on preprocessing, since the model learns to recognize samples with imperfections during training. Not considering common augmentation techniques such as perspective transform, blurring, shearing, hue shifting, etc. can significantly affect the performance of the models \cite{0010kamran2018ailearns}.

Another benefit of augmentation is to fix the class imbalance issue. Class imbalance issue occurs when the distribution of samples among the known classes is skewed. This class imbalance can cause a few problems. First, the model fails to learn generalized features as it only sees a few instances of the classes with a lower number of samples \cite{0500chawla2002smote}. Additionally, due to the small contribution of the small-sized classes in the overall accuracy, a model might achieve high accuracy even when it fails to learn about the small-sized classes \cite{0503leevy2018survey}.

\begin{figure}[htb]
    \centering
    \includegraphics[width=\columnwidth]{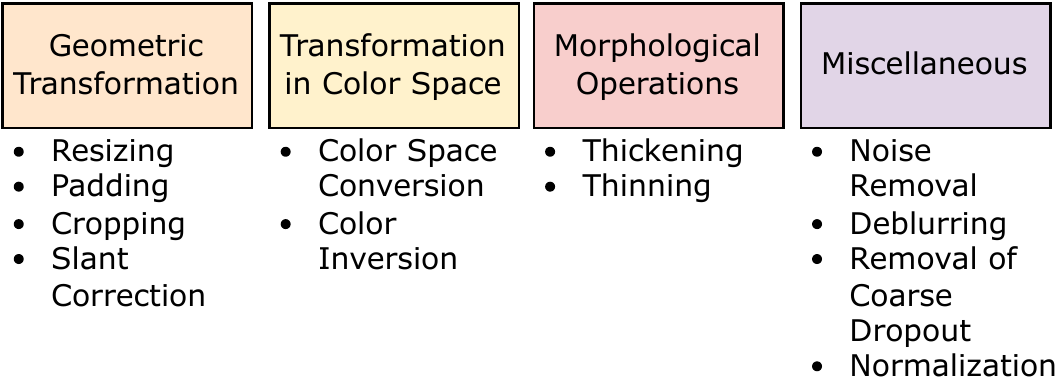}
    \caption{Taxonomy of the augmentation techniques used in Bengali Handwritten Digit Recognition}
    \label{fig:augTaxonomy}
\end{figure}

Several augmentation techniques can be found in the existing literature that emulates real-life scenarios. A taxonomy of the augmentation techniques are shown in \figureautorefname~\ref{fig:augTaxonomy}.

\begin{figure*}[bt]
	\centering
	\subfloat[Original Image\label{fig:augorig}]{
	\includegraphics[width=.24\textwidth]{original.png}	
	}
	\subfloat[Random Rotation\label{fig:augrot}]{
	\includegraphics[width=.24\textwidth]{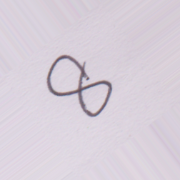}	
	}\hfill
	\subfloat[Random Cropping\label{fig:augcrop}]{
	\includegraphics[width=.2\textwidth]{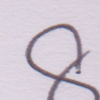}	
	}\hfill
	\subfloat[Random Zoom\label{fig:augzoom}]{
	\includegraphics[width=.24\textwidth]{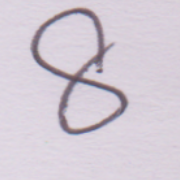}	
	}\\
	\subfloat[Shifting\label{fig:augshift}]{
	\includegraphics[width=.24\textwidth]{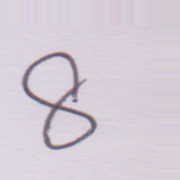}	
	}\hfill
	\subfloat[Shearing\label{fig:augshear}]{
	\includegraphics[width=.24\textwidth]{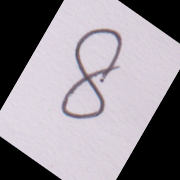}	
	}\hfill
	\subfloat[Distortion\label{fig:augdistort}]{
	\includegraphics[width=.24\textwidth]{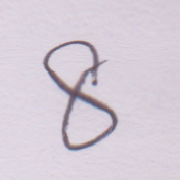}	
	}\hfill
	\subfloat[HSV Shift\label{fig:aughsv}]{
	\includegraphics[width=.24\textwidth]{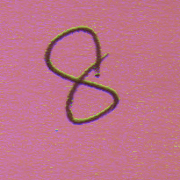}	
	}\\
	\subfloat[Brightness Shift\label{fig:augbright}]{
	\includegraphics[width=.24\textwidth]{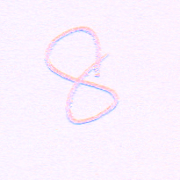}
	}\hfill
	\subfloat[Noise\label{fig:augnoise}]{
	\includegraphics[width=.24\textwidth]{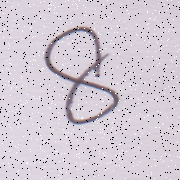}	
	}\hfill
	\subfloat[Coarse Dropout\label{fig:augcoarse}]{
	\includegraphics[width=.24\textwidth]{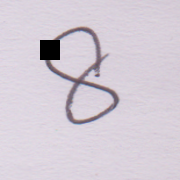}}\hfill
	\subfloat[Superimpose\label{fig:augsuper}]{
	\includegraphics[width=.24\textwidth]{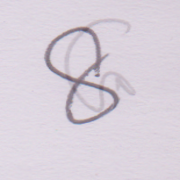}}
	\caption{Common data augmentations techniques used in the existing literature. Original image taken from NumtaDB dataset \cite{0011alam2018numtadb}}
	\label{fig:aug}
\end{figure*}

\subsection{Geometric Transformation}
\subsubsection{Rotation and Flipping}
To help the models recognize slanted texts, rotation can be performed with respect to a certain pixel (\figureautorefname~\ref{fig:aug}\subref{fig:augrot}). Hence, the existing literature on BHDR rotated the digit images by a randomly chosen rotation angle within a range of 0 to 50 degrees on either direction \cite{0002hasan2018recognition, 0003noor2018handwritten, 0009zunair2018unconventional, 0012islam2019sankhya, 0013mahmud2020deepbanglanet, 0030shopon2016bangla, 0036islam2018deep, 0050sharif2016hybrid, 0152sharif2017evil}. One key consideration is that, after performing rotation, additional pixels are padded to preserve the original dimension of the image \cite{0002hasan2018recognition}.
As an extension of rotation, flipping can be performed to flip the image horizontally or vertically along the centerline of the image.

\subsubsection{Random Cropping, Zooming, and Shifting}
Cropping random portions of the image or zooming into it can enable the model to recognize digits by only seeing a portion of them (\figureautorefname~\ref{fig:aug}\subref{fig:augcrop}). Removing the discriminative portion of the digits (For example, the bottom part of `$\bnone$' and `$\bnnine$') can help models learn to ignore non-important portions of the image. It also helps the model recognize occluded characters. Reference \cite{0009zunair2018unconventional} applied random cropping to remove 0 to 20\% of the digit images. A similar effect can be achieved using zooming and shifting. Zooming is performed via pixel replication while keeping the image dimension same (\figureautorefname~\ref{fig:aug}\subref{fig:augzoom}). In BHDR, 0 to 30\% zooming augmentation can be seen \cite{0002hasan2018recognition, 0003noor2018handwritten, 0013mahmud2020deepbanglanet, 0020hasan2020new, 0036islam2018deep, 0169shuvo2021mathnet}. Shifting requires translating each pixel of a sample by a constant factor chosen randomly within a certain range (\figureautorefname~\ref{fig:aug}\subref{fig:augshift}). The common values for the constant factor range from 0 to 0.30 \cite{0002hasan2018recognition, 0003noor2018handwritten, 0009zunair2018unconventional, 0012islam2019sankhya, 0013mahmud2020deepbanglanet, 0020hasan2020new, 0030shopon2016bangla, 0036islam2018deep, 0169shuvo2021mathnet}. In this process, the pixels that are shifted outside the image boundary are discarded and the pixels that become empty are filled with constant values or values of the nearest pixel.

\subsubsection{Distortion}
Shearing can be performed to replicate the effect of distorted digit images (\figureautorefname~\ref{fig:aug}\subref{fig:augshear}). It moves each pixel in a specific direction. The magnitude of the movement is determined by the distance of the corresponding pixels from a certain baseline. The magnitude can also be controlled using a constant factor. Common values for constant factors are often in the range of 0 to 0.25 \cite{0003noor2018handwritten, 0009zunair2018unconventional, 0036islam2018deep}. Shearing in this manner can help the model learn to recognize slanted texts.

Apart from that, application elastic transform \cite{0003noor2018handwritten} and grid distortions \cite{0152sharif2017evil} are also used to distort the digit images (\figureautorefname~\ref{fig:aug}\subref{fig:augdistort}).

\subsection{Transformation in Color Space}
Different color channels found in different color spaces can convey various information regarding the sample. To exploit this information, color space transformation techniques are used to augment the training samples.

\subsubsection{Color Space Conversion}
Reference \cite{0009zunair2018unconventional} used samples in both RGB and HSV color space to extract the latent information of each color channel (\figureautorefname~\ref{fig:aug}\subref{fig:aughsv}). On the other hand, reference \cite{0002hasan2018recognition} randomly shifted the value of the hue and saturation channel. Reference \cite{0007mamun2018bangla} inverted the images by converting the image to grayscale color space and then inverting the intensity values. Even though this technique is popular in deep learning-based models, it is expected that the models themselves should use a set of convolution layers to learn from different color channels, since these transformations can be represented using linear or nonlinear equations.

\subsubsection{Brightness Adjustment}
Considering the heterogeneity of the lighting conditions during the scanning of digits, models can be trained using the same image with different brightness (\figureautorefname~\ref{fig:aug}\subref{fig:augbright}). This can be achieved by shifting the value of the intensity channel in the HSI color space. Reference \cite{0013mahmud2020deepbanglanet} adjusted the brightness of the samples within the range $0.75$ to $1.25$. Again, \cite{0003noor2018handwritten} used a washed-out version of the original samples to achieve a similar effect. This effect can also be achieved using histogram equalization which also enhances the image without any loss of details \cite{1014wadud2007dynamic, 1015kabir2010brightness}.

\subsection{Scanning Defects}
To ensure that the model learns to recognize digits captured with lens blur, data can be augmented using blurred versions of the original samples. To achieve this effect, a Gaussian filter can be applied to the samples \cite{0002hasan2018recognition, 0003noor2018handwritten, 0009zunair2018unconventional}.

\subsubsection{Noise}
Considering the adverse effect of noise in digit recognition, models can be trained to recognize noisy digits (\figureautorefname~\ref{fig:aug}\subref{fig:augnoise}). Consequently, during data augmentation, Salt and Pepper noise \cite{0007mamun2018bangla, 0002hasan2018recognition, 0003noor2018handwritten}, Gaussian noise \cite{0009zunair2018unconventional}, random noise \cite{0007mamun2018bangla}, etc. can be added. Additionally, to incorporate the effects of coarse dropout, random squared portions of the image can be selected and replaced with $0$s (\figureautorefname~\ref{fig:aug}\subref{fig:augcoarse}) \cite{0007mamun2018bangla, 0002hasan2018recognition, 0003noor2018handwritten}.

\subsubsection{Superimposition}
While scanning a digit, if there is something written on the opposite side of the page, a horizontally flipped and blurred version of the text can appear superimposed on the original digit (\figureautorefname~\ref{fig:aug}\subref{fig:augsuper}). To ensure that the model learns to recognize the original digit in these scenarios, multiple training images can be superimposed that achieve the same effect. To do that, a weighted Gaussian blur of one image is added with another image \cite{0002hasan2018recognition, 0003noor2018handwritten, 0036islam2018deep}. This process is known as superimposition.

\begin{figure*}[htb]
    \centering
    \includegraphics[width=0.75\textwidth]{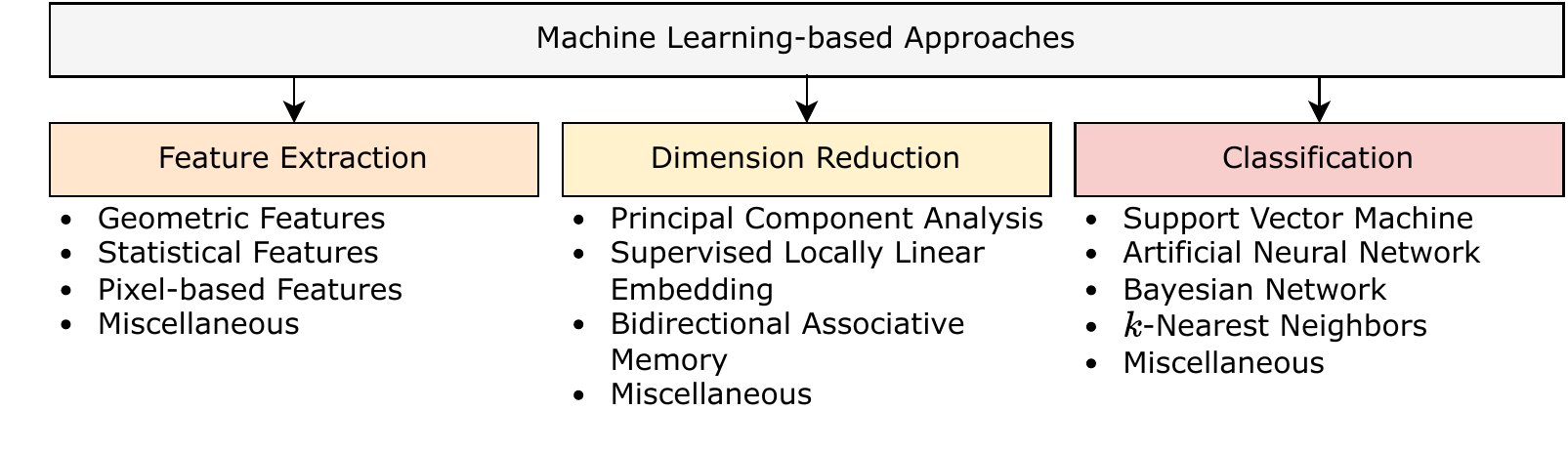}
    \caption{Taxonomy of the machine learning-based techniques used in Bengali Handwritten Digit Recognition}
    \label{fig:ml_hierarchy}
\end{figure*}

\subsection{Mixed}
To add further variety to the augmented samples, multiple augmentations can be combined to generate additional samples. For example, \cite{0002hasan2018recognition} randomly picked 1-3 augmentations and applied them simultaneously on their dataset to create 5 to 7 new images from each image of the training set.

\section{Machine Learning-based Approaches}\label{sec:machine}
Machine Learning-based approaches mostly focus on feature extraction and classification. In such works, handcrafted features are extracted from the digit images and then fed to a classifier to generate the prediction. Similar to other tasks involving the classification of images, the development of a conventional machine learning-based system for recognizing handwritten Bengali digits typically entails the completion of three phases. The first thing that we do is extract crucial information from the image that we are given, and the information that is extracted is referred to as features. In a later step, the feature space is shrunk down to a smaller dimensional size. This can be beneficial for lowering the required level of computing complexity or getting rid of features that are superfluous or biased and have the potential to confuse a model. In the last step, the characteristics that were chosen are sent into a classifier so that it may be trained. Previous research has demonstrated that it is possible to successfully use a variety of approaches to each of these processes. \figureautorefname~\ref{fig:ml_hierarchy} illustrates a schematic representation of the taxonomy of the machine learning-based approach used for this particular task.

\subsection{Feature Extraction}
One of the most challenging tasks of handwritten digit classification in earlier machine learning-based approaches was to find a suitable feature extraction technique that can uniquely represent the individual digits. For Bengali handwritten digit recognition, these descriptors can be divided into four broad categories:
\begin{enumerate}
    \item Geometric Features
    \item Statistical Features
    \item Pixel-based Features
    \item Miscellaneous 
\end{enumerate}

\subsubsection{Geometric Feature Extraction Methods}
\begin{figure}[htb]
    \centering
    \includegraphics[width=\columnwidth]{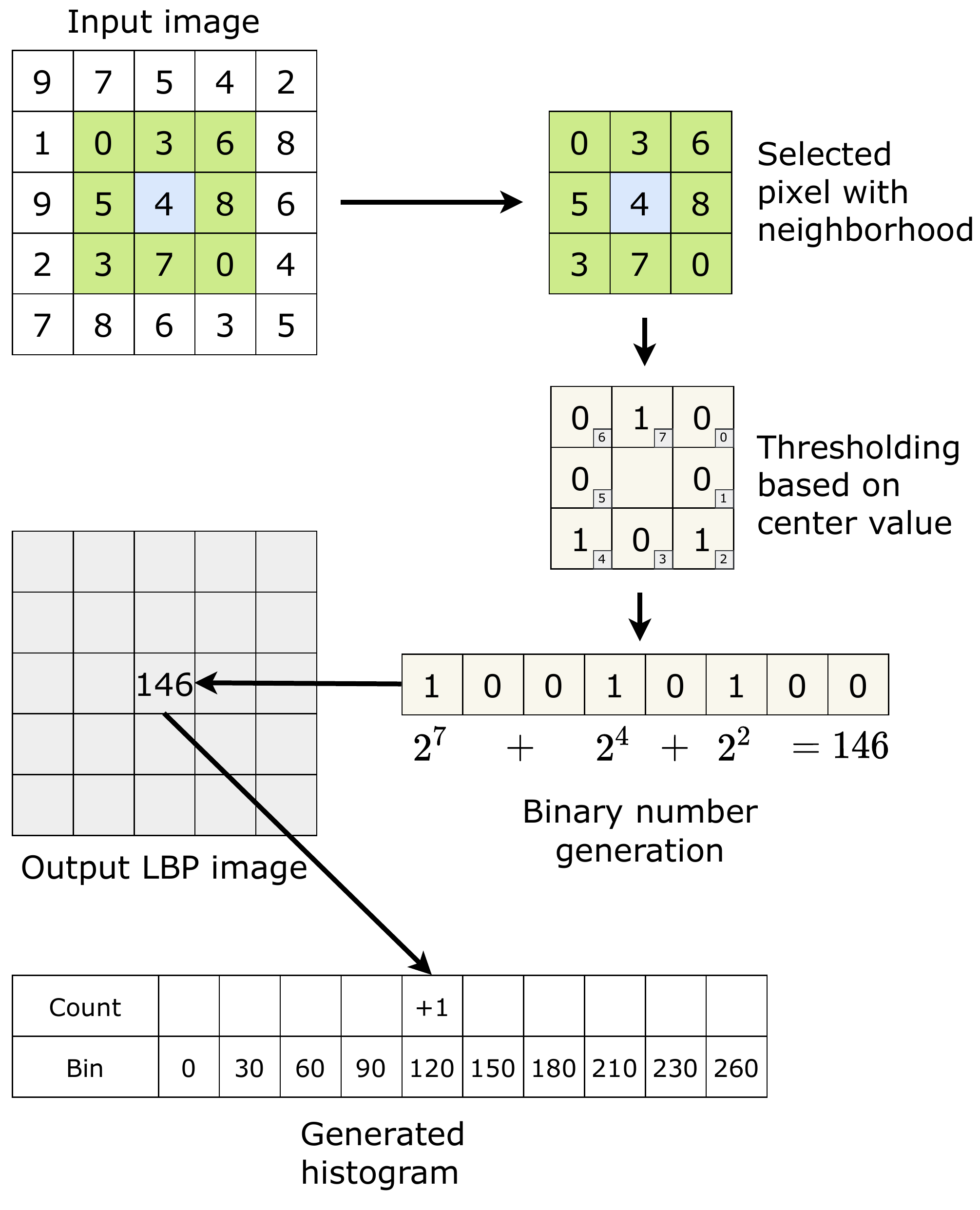}
    \caption{Local Binary Pattern (LBP) generation from a pixel. This process is repeated for all the pixels in the input image.}
    \label{fig:mllbp}
\end{figure}

\paragraph{Local Binary Pattern (LBP)} 
Local Binary Pattern is an efficient texture operator, with an easy and robust calculation method \cite{0509wang1990texture}. It iterates through each pixel of the image and generates a label by comparing each neighboring pixel with the pixel itself. Each label is either `0' or `1' based on whether it is bigger or smaller than the center pixel. All the labels of the neighboring pixels are concatenated to form a binary number. Later, it creates a histogram using the values of those binary numbers. An illustration of calculating LBP is shown in \figureautorefname~\ref{fig:mllbp}. 
Reference \cite{0157hassan2015handwritten} used the basic, uniform, and simplified variations of LBP. Each of them produced an accuracy of around 96.5\% while simplified LBP slightly underperformed. On the other hand, \cite{0166wahid2021classical} used LBP with 10 neighbors with a radius of 3 and reported an accuracy of 95.3\% on the CMATERdb dataset. However, the performance of LBP was comparatively worse in datasets containing challenging samples such as NumtaDB and Ekush. This can be attributed to the fact that LBP is prone to variations in illumination and random noises \cite{1019jabid2010robust}, which are present in digit samples of those datasets.

\paragraph{Histogram of Oriented Gradient (HOG)}
The Histogram of Oriented Gradient (HOG) feature descriptor exploits the count of the occurrence of the gradient orientation in localized regions of an image \cite{0506dalal2005histograms}. It determines whether a pixel is on an edge or not, along with the gradient and the direction of that pixel. It calculates the gradient for every pixel along both axes of the image. There are different filters that can be used to calculate the gradient such as Sobel \cite{0568sobel1968isotropic}, Kirsch \cite{0570kirsch1971computer}, Roberts \cite{0569roberts1963machine}, etc. Then the magnitude and the orientation are calculated using \equationautorefname~\ref{eqn:mlhogmag} and \ref{eqn:mlhogori}, respectively.
\begin{align}
    \text{Magnitude} &= \sqrt{G_x^2 + G_y^2}\label{eqn:mlhogmag}\\
    \text{Orientation} &= \tan^{-1}\left(\frac{G_y}{G_x}\right)\label{eqn:mlhogori}
\end{align}
where $G_x$ and $G_y$ are the difference between the neighboring pixel values on the x-axis and the y-axis, respectively.

\begin{figure}[htb]
    \centering
    \includegraphics[width=0.8\columnwidth]{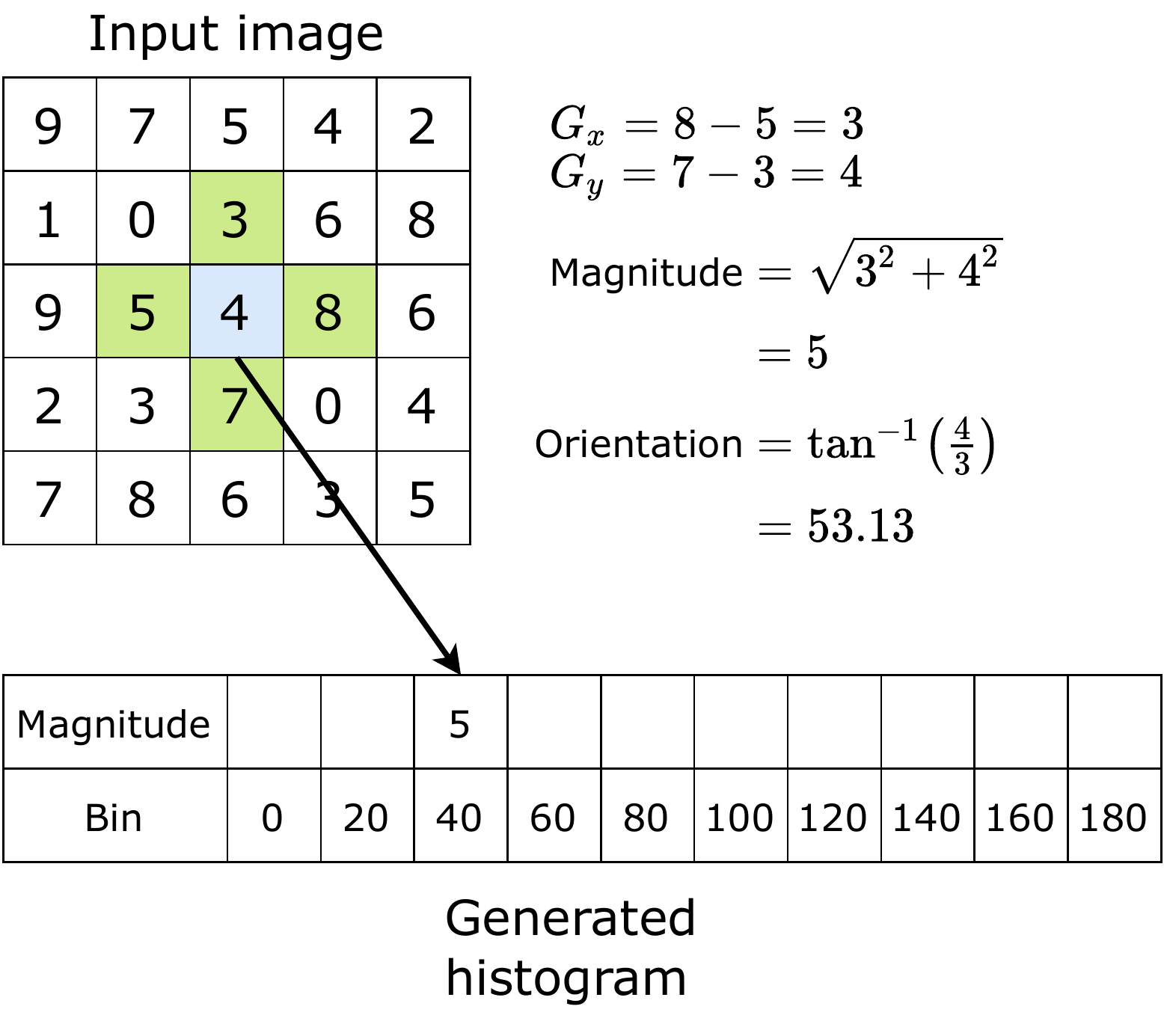}
    \caption{Histogram of Oriented Gradients (HOG)-based histogram generation from a pixel. This process is repeated for all the pixels in the image.}
    \label{fig:mlhog}
\end{figure}

After that, a histogram is generated, storing the magnitude into several bins based on the value of the orientation. The size of the filter and the number of bins in the histogram depends on the implementation. This also controls the number of features generated.  An illustration of calculating HOG is shown in \figureautorefname~\ref{fig:mlhog}.

Reference \cite{0049choudhury2018handwritten} used $8\times8$ cell to create a feature vector along with the color histogram and reported 98.05\% accuracy on the CMATERdb dataset with SVM as the classifier. In another work, the authors used $2\times2$ cell to create a feature vector of length 6084 \cite{0166wahid2021classical}. On the CMATERdb dataset, they reported 98.08\% accuracy. However, the performance on other datasets was relatively low: 93.32\% on NumtaDB and 95.68\% on Ekush. Reference \cite{0174chenglin2009new} used Sobel and Roberts filters to create the histogram. They presented a comparative analysis of the performances of the ISI-HBN dataset from different viewpoints. The authors claimed that the Sobel operator performs slightly better than Roberts with a maximum accuracy of 99.40\%. Reference  \cite{0192rehana2017bangla} also used HOG features with normalization.

\paragraph{Directional Pattern} 
A number of directional patterns, such as: Local Directional Pattern (LDP) \cite{0510jabid2010local}, Gradient Directional Pattern (GDP) \cite{0511ahmed2012gradient}, etc. have been used as feature descriptor over the years.

LDP is similar to LBP, except that it creates the binary code using 8 different Kirsch filters instead of direct thresholding. As it considers edge responses instead of the pixel density of the neighboring pixels, it can provide more consistency in the presence of noise compared to LBP \cite{1020jabid2010local}. On the other hand, GDP applies operators like the Sobel masks to the image to find the gradient along both the X and Y-axis. The gradient orientation uses the same equation as the orientation of HOG features \equationautorefname~\ref{eqn:mlhogori}. After that, the orientation of a pixel is compared with the neighboring pixels, and based on a threshold, it is coded as either `0' or `1', creating a binary number. Finally, a histogram is obtained from the values of the binary number similar to LDP. The benefit of using GDP is that the produced codes are consistent for identical and near-identical regions of digit images, unlike the ones produced by LDP \cite{0511ahmed2012gradient}.

Reference \cite{0005aziz2017bangla} investigated the performance of both GDP and LDP. Authors claimed that an ensemble of both features yields maximum accuracy of 95.62\% on the CMATERdb dataset.


\paragraph{Wavelet filter and Chain Code}
Wavelet filters are a well-established tool for multi-resolution analysis of handwritten digit images \cite{00572kingsbury1999image}. Wavelet filters decompose an image into a hierarchy of several levels of resolution, making a coarse-to-fine recognition scheme possible. There exist many sets of wavelets used in image analysis. Usually, the input image is split into two components: a smooth component and a detail component by using low-pass and a high-pass filter, respectively. Both components are then down-sampled by a factor of 2. The low-pass component is then split further into low-pass and high-pass components as above for the second time and they are again down-sampled by a factor of 2. This process of splitting and down-sampling, also known as the `pyramidal algorithm', is continued as far as required.

Reference \cite{0044bhattacharya2009handwritten} proposed a system using a wavelet filter and chain-code-based feature extraction technique. The authors used Daubechies-4 wavelets \cite{00571daubechies1990filter}, which works better in both the time domain and frequency domain. After that, chain code histogram features are obtained corresponding to each of the detailed image components. The bounding box is divided into equal sizes of blocks and for each block, a local histogram of chain code is computed. The size of the feature vector is reduced to 64 by down-sampling using a Gaussian filter. Later, an MLP classifier is used to classify. A wavelet filter, consisting of linear operations, is computationally fast and suitable for real-life applications. The authors achieved 98.2\% test accuracy on the ISI-HBN dataset.

\paragraph{Gabor Feature}
The Gabor filter can approximate the characteristics of a certain cell in an image in the visual cortex of a mammal \cite{0573mehrotra1992gabor}. It is a variant of band-pass filters that allows passing a band of frequency in a certain orientation. It analyzes whether there is any specific frequency content in the image in specific directions within a localized region around the point or region of analysis. Reference \cite{0166wahid2021classical} used a 2D Gabor filter with default parameters for digit recognition and compared the performance of Gabor features with HOG and LBP using a variety of classifiers on NumtaDB, CMATERdb, and Ekush datasets. They concluded that Gabor features perform poorly compared to other features in all the datasets used.

\paragraph{Projection Histogram}
To generate a projection histogram, the image is iterated row-wise and the number of pixels that are part of the digit is counted for each row. The same process is repeated for each column. Then, considering the rows and columns as bins, a histogram is created. This feature is also used in digit segmentation. However, \cite{0164hossain2011rapid} concluded that this feature performs poorly in digit recognition with an accuracy of around 82\% on a self-curated dataset. This was further improved by considering a celled version of projection, where the image is divided into multiple cells and the length of the projection is considered. Celled projection can improve the accuracy up to 92.60\% on the same dataset \cite{0164hossain2011rapid}. In another work, \cite{0230singh2017recognition} exploited Mojette transformation that converts the 2D image into a set of discrete 1D projections. It generates a set of vectors, where each of the elements is calculated using the sum of the value of pixels that lie on the projection line.

\begin{figure}[htb]
    \centering
    \includegraphics[width=0.9\columnwidth]{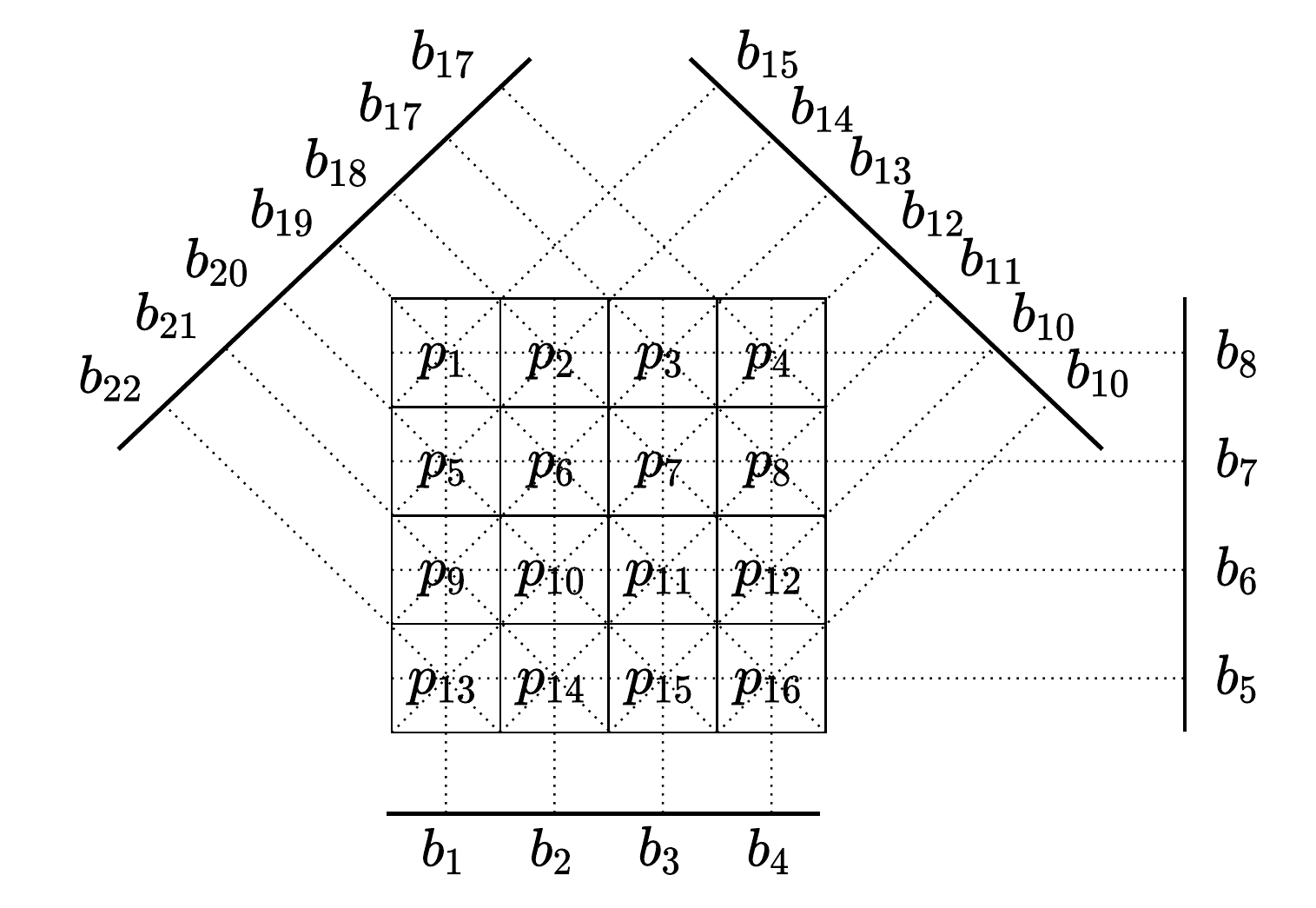}
    \caption{Projection histogram generated from a $4\times4$ image. Adapted from \cite{0230singh2017recognition}.}
    \label{fig:mlprojection}
\end{figure}

\begin{figure}[htb]
	\centering
	\subfloat[\label{fig:shadow1}]{
	\includegraphics[width=.45\columnwidth, height=0.45\columnwidth]{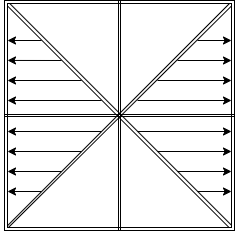}
	}
	\subfloat[\label{fig:shadow2}]{
	\includegraphics[width=.45\columnwidth, height=0.45\columnwidth]{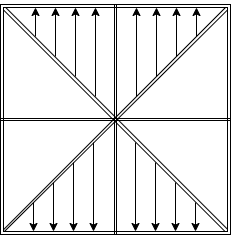}	
	}\\
	\subfloat[\label{fig:shadow3}]{
	\includegraphics[width=.45\columnwidth, height=0.45\columnwidth]{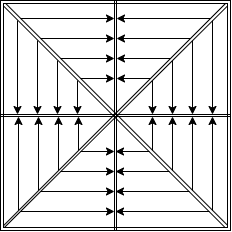}
	}
	\subfloat[\label{fig:shadow4}]{
	\includegraphics[width=.45\columnwidth, height=0.45\columnwidth]{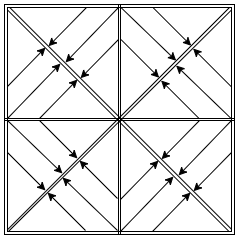}	
	}\\
	\subfloat[\label{fig:shadow5}]{
	\includegraphics[width=.45\columnwidth, height=0.45\columnwidth]{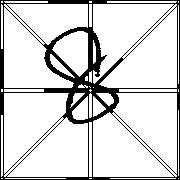}
	}
	\caption{Extraction of Shadow-based Features. The arrows show the direction of light for generating the shadow. Digit image taken from NumtaDB dataset \cite{0011alam2018numtadb}}
	\label{fig:mlshadow}
\end{figure}

\paragraph{Shadow-based Features}
Shadow features are calculated using the length of the shadow that is supposed to project on a surface. As the shape of Bengali digits varies a lot, the shadow length also varies from different lighting positions. The images are divided into different zones to extract multiple shadow features. References \cite{0153basu2005handwritten, 0155das2012genetic, 0195basu2005mlp} proposed zone-based shadow features dividing the image into 8 triangular shaped zones extracting 24 shadow lengths (\figureautorefname~\ref{fig:mlshadow}).

\begin{figure*}[htb]
    \centering
    \includegraphics[width=0.6\textwidth]{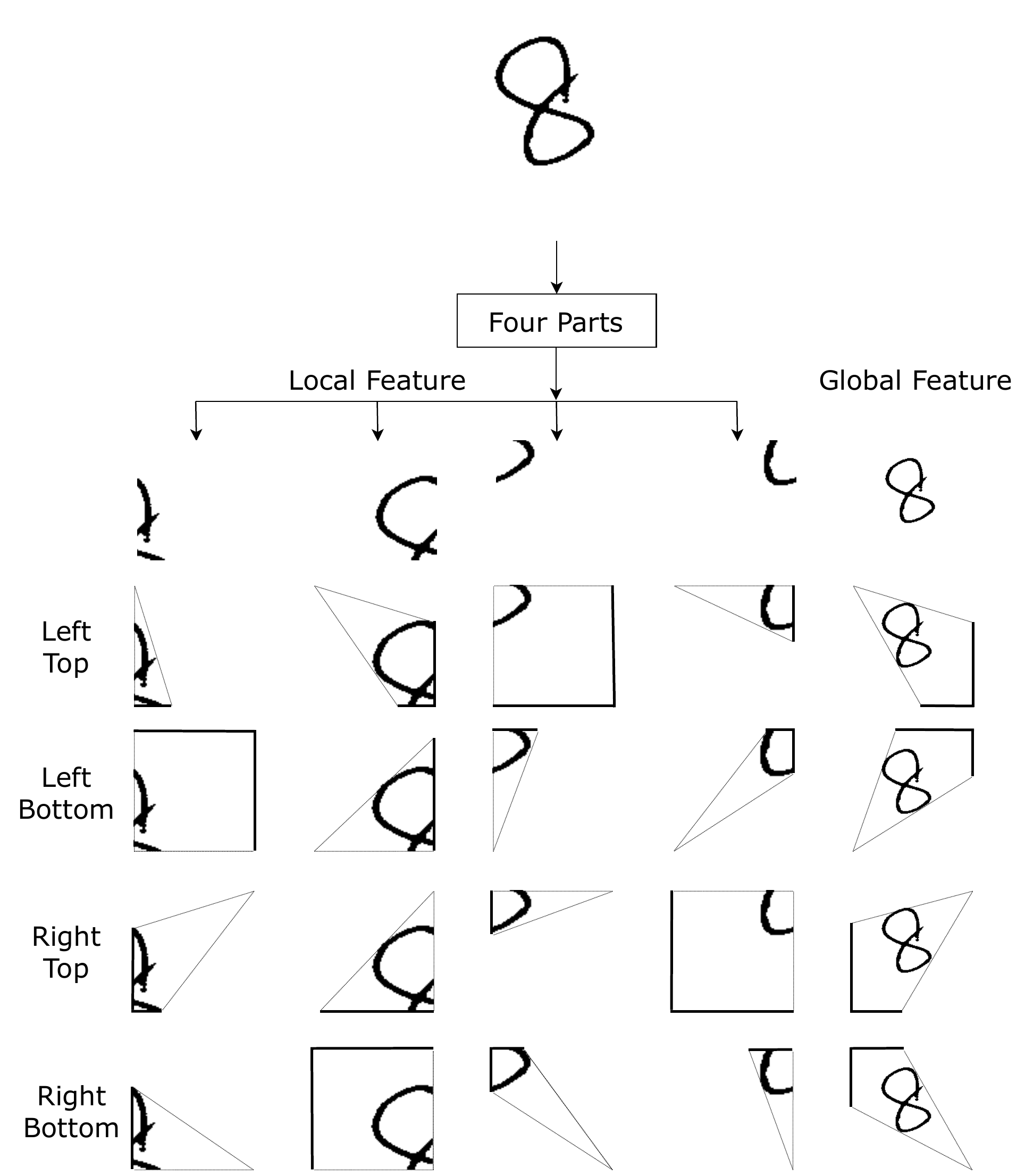}
    \caption{Light source-based shadow feature extraction. Digit image taken from NumtaDB dataset \cite{0011alam2018numtadb}}
    \label{fig:shadowlocal}
\end{figure*}

In a more recent work, \cite{0162ghosh2021language} proposed a single light source-based shadow feature named Point Light Source-based Shadow (PLSS). The shadow changes if the position of the light source is changed (\figureautorefname~\ref{fig:shadowlocal}). They obtained 20 shadow lengths from four different source positions for four local regions and one global image. 
All the methods combined shadow-based features with other features, i.e. centroids, pixel density, and crossing to construct the feature vector.

\paragraph{Contour Angular Technique}
Reference \cite{0165surinta2013comparison} proposed a system named Contour Angular Technique (CAT) to capture the aspects of the curvature in the numeral image and angular co-occurrence. CAT is a fast implementation of quantized angle co-occurrence computation similar to the Hinge feature \cite{0532bulacu2007text}. In this method, the original image is divided into $4\times4$ non-overlapping blocks. Then 8-directional code for identifying the contour of the neighboring pixels of a starting point is calculated for each of the blocks. Along with this, it also calculates an angular co-occurrence histogram with 64 elements that approximates the angular co-occurrence probability along the contours. By combining both outputs, a feature vector of size 192 is extracted which is then fed to SVM classifiers. Authors reported 96\% accuracy on a self-made dataset.

\subsubsection{Statistical Feature Extraction Methods}

\paragraph{Moments}
Moments calculate the center of gravity of the image considering the pixel distribution to capture global shape information \cite{1016hausdorff1921summationsmethoden}. They can be used to describe quantities at a distance from an axis or a point. They were popular due to their invariance to rotation, translation, and scaling. Reference \cite{0164hossain2011rapid} constructed a feature vector with fifteen translation-invariant central moments and achieved a maximum accuracy of 86.7\% with a Feed Forward Back Propagation Neural Network. In another work, \cite{0231singh2016study} extracted a feature vector containing 130 features that combines six types of moments: Complex moment \cite{0590yaser1984recognitive}, Zernike moment \cite{0591khotanazad1990invariant}, Legendre moment \cite{0592yap2005Legendre}, Geometric moment \cite{0593gonzalez2009digital}, Affine moment invariant \cite{0594petrou2004affine}, and Moment invariant \cite{0512hu1962visual}. They reported 99.5\% accuracy on the CMATERdb dataset with MLP as the classifier.

\begin{figure}[htb]
    \centering
    \includegraphics[width=0.5\columnwidth]{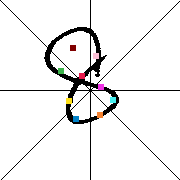}
    \caption{Centroid features extracted from each octant. The red dot denotes the global centroid and the rest are local. Original image taken from NumtaDB dataset \cite{0011alam2018numtadb}. (Best viewed in color)}
    \label{fig:mlcentroid}
\end{figure}

\paragraph{Centroids}
Centroid is the middle point of an object. In image analysis, centroids are obtained as the weighted average of the pixel coordinates belonging to the digit in a region of interest. Global centroids are calculated considering all the pixels of the image, whereas local centroids are calculated for different zones. \figureautorefname~\ref{fig:mlcentroid} shows the centroids for each octant along with the global centroid.

Reference \cite{0203azmi2013exploiting} used the global centroid to divide the image vertically into two regions. A scalene triangle was created using the three centroids (one global centroid and two centroids of two regions) as the corner points. Then 9 different features were obtained using triangular geometry.


\paragraph{Pixel Density}
Pixel density is another statistical feature that is often used in different image classification tasks. It counts the number of black pixels within a region of interest; be it the whole image or a zone division within the image. In some methods, it is normalized by the size of the zone to obtain the density.

\begin{figure}[htb]
	\centering
	\subfloat[Octant zone density\label{fig:mloctzone}]{
	\includegraphics[width=.45\columnwidth]{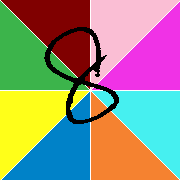}
	}
	\subfloat[Rectangular zone density\label{fig:mlsqzone}]{
	\includegraphics[width=.45\columnwidth]{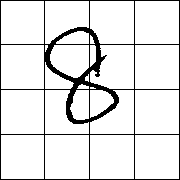}	
	}
	\caption{Extraction of zone-based density features. Original image taken from NumtaDB dataset \cite{0011alam2018numtadb}}
	\label{fig:mlzone}
\end{figure}

Reference \cite{0162ghosh2021language} divided the image into eight octants and created a histogram called Histogram of Oriented Pixel Positions (HOPP), counting the number of black pixels within the region. The HOPP feature is combined with the PLSS feature to achieve 98.5\% accuracy on CMATERdb 3.1.1 dataset. In other works, the whole image was divided into rectangular zones, and zone-wise local density was taken as features \cite{0201majumdar2006mlp, 0164hossain2011rapid, 0045khan2014handwritten}. Reference \cite{0164hossain2011rapid} used $4\times4$ zoning and achieved 90.30\% accuracy on self-curated dataset. Reference \cite{0201majumdar2006mlp} combined pixel density with other statistical features and reported 95.7\% accuracy on their own dataset. Reference \cite{0045khan2014handwritten} reported 94.0\% accuracy on ISI-HBN dataset using $8\times8$ zoning.

\begin{figure}[htb]
    \centering
    \includegraphics[width=0.48\columnwidth]{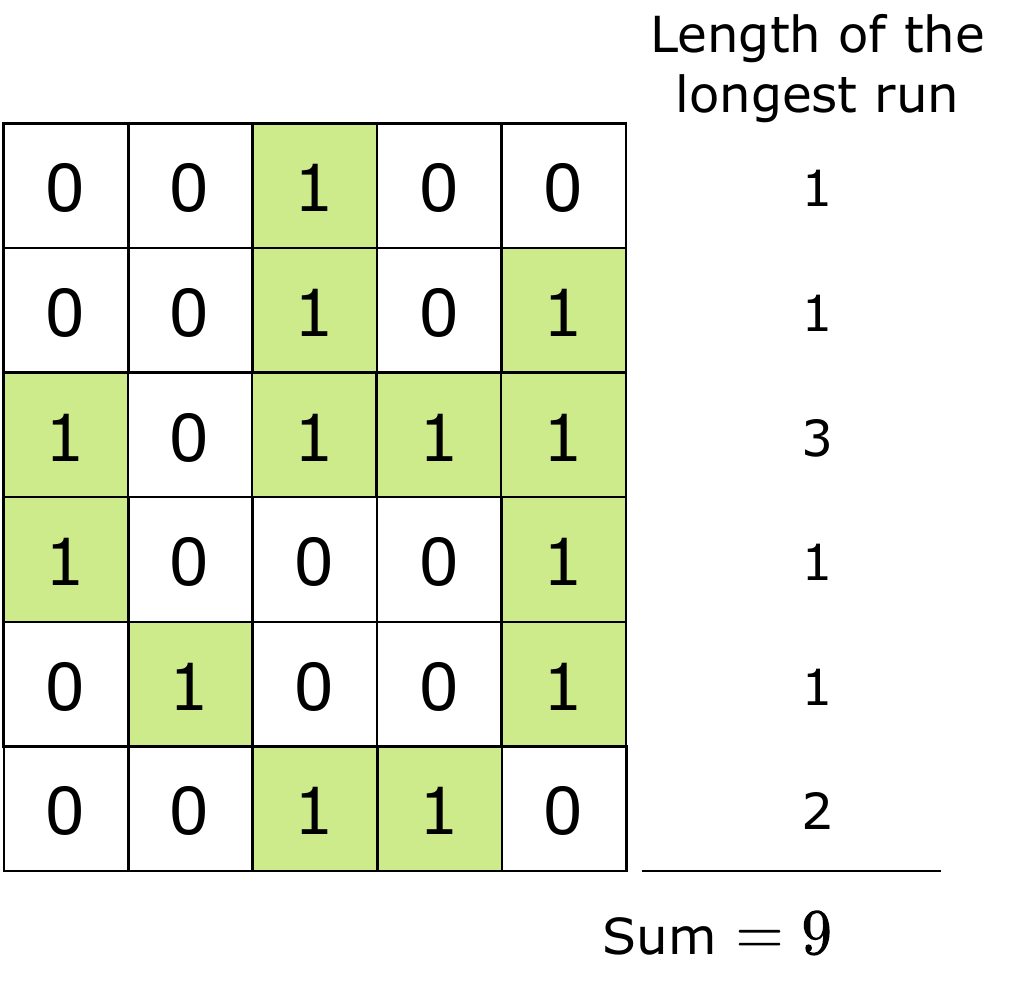}
    \caption{Longest run feature extraction in the horizontal direction of a binary image of the digit six ($\protect\bnsix$).}
    \label{fig:mllongest}
\end{figure}

\paragraph{Longest Run}
The longest run of a numeral image is the maximum number of consecutive black pixels along with any directions. The longest run is usually calculated considering four directions: horizontal, vertical, diagonal, and anti-diagonal. Reference \cite{0155das2012genetic} used the length of the longest run in four directions. As illustrated in \figureautorefname~\ref{fig:mllongest}, the longest run along the row is calculated considering the length of the longest bar that fits the maximum consecutive black pixel is computed for each row. Then the sum of the lengths for all the rows is calculated to get the horizontal longest run. The same method is repeated for the vertical and the two diagonals. Each of the values is then normalized based on the length of the direction. References \cite{0195basu2005mlp, 0153basu2005handwritten, 0202roy2012new} used a normalized longest run for 9 overlapping windows.

\paragraph{Crossing}
Crossing is obtained by counting the number of transitions between the background to the foreground or vice versa in any direction in the image. In simple words, it counts the number of changes between black and white pixels along each row or column. It can be obtained along the diagonals as well. Unlike other features, it is invariant to the width of the strokes and does not need thinning. Reference \cite{0164hossain2011rapid} computed crossings for every column and row to construct the feature vector of the image and reported 86.40\% accuracy on a self-curated dataset of 12000 images. Reference \cite{1004singh2018script} divided the image into 16 zones and a mask of 4 regions surrounding each other was applied to each of the neighboring zones. Crossing was counted for each of the regions along 4 different directions: vertical, horizontal, and the two diagonals. Authors claimed that a feature set of length 144 yielded 97.85\% accuracy on a self-curated dataset of 10,000 images.

\paragraph{The Hotspot Technique}
Reference \cite{0165surinta2013comparison} proposed a feature extraction method to represent the numeral image that considers the distance between evenly spaced hotspots in the image and the nearest black pixels in any direction. The distance is set to a maximum threshold value if the black pixel is not found in any direction. The authors used 25 hotspots and 4 directions to create a feature vector of length 100. As this is a global feature and lacks the ability to uniquely represent the numerals, this feature accumulated maximum accuracy of 92.7\% on a self-curated dataset.

\begin{figure}[htb]
    \centering
    \includegraphics[width=0.6\columnwidth]{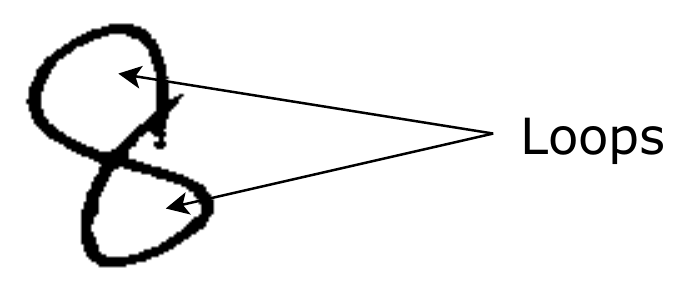}
    \caption{Bengali digit 4 containing 2 loops}
    \label{fig:mlloop}
\end{figure}

\paragraph{Loop Features}
The area of the completely enclosed portion of the digit pattern is considered a loop. In Bengali digits, the number and the size of the loop vary from digit to digit (\figureautorefname~\ref{fig:mlloop}). This feature was utilized by \cite{0155das2012genetic}. In other works, the relative positions of the loops were considered in addition to the area and the number \cite{0180pal2000automatic, 0201majumdar2006mlp}.

\begin{figure}[htb]
    \centering
    \includegraphics[height=2cm]{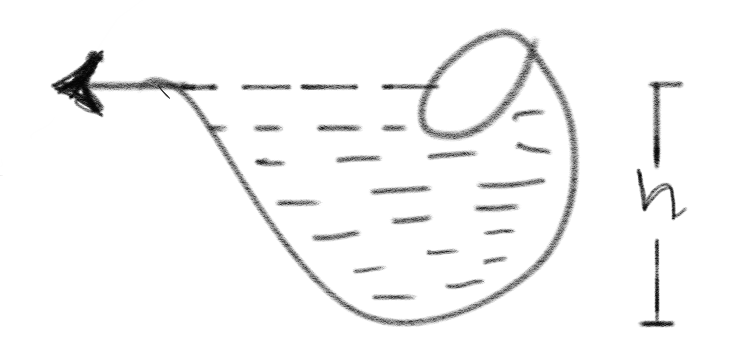} \includegraphics[height=2cm]{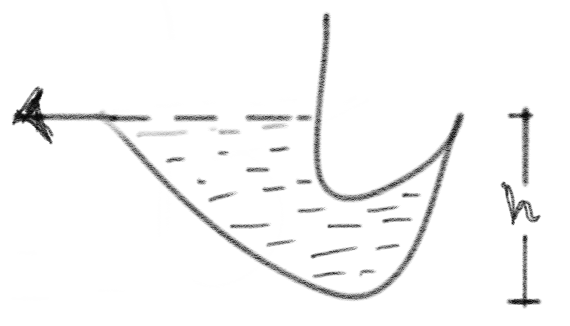}
    \caption{Water overflow features for digit 3 and 6}
    \label{fig:waterflow}
\end{figure}

\paragraph{Water overflow from reservoir}
Water overflow from reservoir takes advantage of the rounded nature of Bengali digits (\figureautorefname~\ref{fig:waterflow}). It considers that the closed portion of the digits was filled with water, and extracts features based on that. Its use was first seen in \cite{0180pal2000automatic}, who used it as an extension of loop features. If there was no loop to be found, this feature was used. In another work, \cite{0173pal2006complete} extracted overflow features such as the direction of the overflow, the height of the water during overflow; reservoir features such as position and shape of the reservoir, etc. These features were particularly useful in the segmentation of touching numerals.

\subsubsection{Pixel-based Feature Extraction Methods}
Instead of calculating the features through statistical or geometric methods, the pixel-based method directly uses the intensity of the pixel values from either the original image or sub-images. The benefit of using this method is the avoidance of error due to the miscalculation of features, which can occur in complex images. As a result, the performance of these methods can be better than that of geometric/statistical feature-based methods. These pixel values are often fed into artificial neural networks \cite{0190chowdhury2016towards}. Reference \cite{0197hashem2014handwritten} used a combination of Back-Propagation Neural Network (BPNN) and Bidirectional Associative Memory (BAM) to accomplish this task. Reference \cite{0194xu2008handwritten} proposed the use of Hierarchical Bayesian Network (HBN). It took $32\times32$ images that are divided into patches of $4\times4$ pixels. The input layer of the network consisted of $16$ nodes, each corresponding to one pixel of the patch. The hidden layer had $4$ children nodes for each of the nodes in the input layer. These nodes are fed into a single node in the output layer. Another work on pixel-based feature extraction methods depended on Supervised Locally Linear Embedding (SLLE) to reduce the dimension of the feature vector before feeding it to an SVM classifier \cite{0198ahmed2009bangla}.

\subsubsection{Miscellaneous}
\paragraph{Image Distortion Model Distance}
Image Distortion Model Distance (IDMD), proposed in \cite{0531keysers2007deformation}, is a well-known feature used for image recognition or matching. Despite having high accuracy in recognition, the computation of this distance is quite expensive while handling a large dataset. Reference \cite{0211cecotti2015handwritten} proposed a method of selecting a limited number of images to keep the computation time within reach while maintaining high accuracy. They proposed an ensemble of multi-classifier approaches to achieve an accuracy of  98.55\% on the ISI-HBN dataset.

\paragraph{Fourier Transform}
Fourier transform is a popular method in signal processing. It provides information about the frequencies present in a signal; image in our case. Low frequency carries shape and structure information, whereas high frequency provides finer details. In digit recognition, shape information is much more important than finer details. Reference \cite{0164hossain2011rapid} proposed the use of 64 low-frequencies as a feature vector since it reduces the feature dimension while also keeping the shape information intact. As subtle changes in the time domain do not yield a significant change in the frequency domain, the performance of this feature is not worth mentioning. 

\begin{table*}[t]
    \centering
    \caption{Performances of the traditional machine learning-based approaches in self-curated datasets}
    \label{tab:mlperfself}
    \begin{tabular}{C{1.4cm} C{0.7cm} L{4.85cm} L{4.15cm} C{1.6cm} C{1.3cm}}
        \toprule
        \textbf{Reference} & \textbf{Year} & \textbf{Feature Extraction} & \textbf{Classifiers} & \textbf{Number of Images} & \textbf{Accuracy (\%)}\\
        \midrule
        \cite{0180pal2000automatic} & 2000 & Water resorvoir & Decision tree & 10000 & 91.98\\
        \cite{0173pal2006complete} & 2006 & Water overflow from resorvoir, topological and structural features of the numerals & Binary tree classification & 12000 & 92.80\\
        \cite{0201majumdar2006mlp} & 2006 & Pixel and shape features: curves, holes, terminating and intersecting points & Multilayer Perceptron & 10500 &  97.20\\
        \cite{0200hoque2006efficient} & 2006 & Fuzzy frequency using histogram analysis & Rule-based & N/A  &  85.00\\
        \cite{0161wen2007handwritten} & 2007 & $3\times3$ Kirsch Mask for edge detection & Support Vector Machine & 16000 & 95.05\\
        \cite{0194xu2008handwritten} & 2008 & Not applicable & Hierarchical Bayesian Network & 2000 & 87.50\\
        \cite{0199pal2008bangla} & 2008 & Histogram of direction chain code of the contour points and gradient based features & Modified quadratic discriminant function & 2692 & 99.08\\
        \cite{0044bhattacharya2009handwritten} & 2009 & Orthogonal wavelet and chain code histogram & Multilayer perceptron & 23392 & 98.20\\
        \cite{0164hossain2011rapid} & 2011 & Celled projections & Probabilistic Neural Network & 12000 & 94.12\\
        \cite{0154wen2012classifier} & 2012 & Polynomial and Gaussian kernels & Kernel and Bayesian Discriminant-based Classifier & 45000 & 96.91\\
        \cite{0165surinta2013comparison} & 2013 & Contour Angular Technique & Support Vector Machine & 10920 & 96.80\\
        \cite{0197hashem2014handwritten} & 2014 & Pixel values & Backpropagation Neural Network & 1400 & 91.10\\
        \cite{0190chowdhury2016towards} & 2016 & Pixel values & Multilayer Perceptron & 70000 & 96.05\\
        \bottomrule
    \end{tabular}
\end{table*}

\begin{table*}[tb]
    \centering
    \caption{Performances of the traditional machine learning-based approaches in the commonly used datasets}
    \label{tab:mlperfother}
    \begin{tabular}{C{1.4cm} C{0.7cm} L{4.15cm} L{4.15cm} C{1.6cm} C{1.6cm} C{1.3cm} }
        \toprule
        \textbf{Reference} & \textbf{Year} & \textbf{Feature Extraction} & \textbf{Classifiers} & \textbf{Dataset} & \textbf{Number of Images} & \textbf{Accuracy (\%)}\\
        \midrule
        
        \cite{0192rehana2017bangla} & 2017 & Histogram of Oriented Gradient, Histogram generation, and block normalization & Support Vector Machine & BLI & 6000 & 97.08\\
        
        \cite{0198ahmed2009bangla} & 2009 & Pixel features based on supervised locally linear embeddings & Support Vector Machine & ISI-HBN & 70000 & 97.06\\
        \cite{0174chenglin2009new} & 2009 & 8-directional gradient histogram extracted using Kirsch and Roberts mask & Discriminative Learning Quadratic Discriminant Function & ISI-HBN & 23392 & 99.40\\
        \cite{0202roy2012new} & 2012 & Quad tree-based directional gradient and longest run & Support Vector Machine & ISI-HBN & 4200 & 93.34\\
        \cite{0153basu2005handwritten} & 2005 & Octant-based (shadow and centroid) and longest run & Multilayer Perceptron & ISI-HBN & 6000 & 95.10 \\
        
        \cite{0195basu2005mlp} & 2005 & Shadow, centroid and longest run features & Multilayer Perceptron & CMATERdb 3.1.1 & 6000 & 96.67 \\
        \cite{0155das2012genetic} & 2012 & Genetic Algorithm for Local Region Sampling & Support Vector Machine & CMATERdb 3.1.1 & 6000 & 97.00\\
        \cite{0045khan2014handwritten} & 2014 & Zone density & Sparse Representation Classifier & CMATERdb 3.1.1 & 6000 & 94.00 \\
        \cite{0157hassan2015handwritten} & 2015 & Local Binary Pattern with Basic, Uniform and Simplified Zoning & $k$-Nearest Neighbor & CMATERdb 3.1.1 & 6000 & 96.70 \\
        \cite{0231singh2016study} & 2016 & Moments & Multilayer Perceptron & CMATERdb 3.1.1 & 6000 & 99.50\\
        \cite{0005aziz2017bangla} & 2017 & Local and Global Directional Pattern & Ensemble of $k$-Nearest Neighbor and Support Vector Machine & CMATERdb 3.1.1 & 6000 & 95.62\\
        \cite{0049choudhury2018handwritten} & 2018 & Histogram of Oriented Gradients & Support Vector Machine & CMATERdb 3.1.1 & 6000 & 98.05\\
        \cite{0162ghosh2021language} & 2021 & Point Light Source-based shadow and Histogram of Oriented Pixel Positions & Random Forest & CMATERdb 3.1.1 & 6000 & 98.50\\
        \cite{0166wahid2021classical} & 2021 & \multirow{4}{3.4cm}{Histogram of Oriented Gradient} & \multirow{4}{3.4cm}{Support Vector Machine} & CMATERdb 3.1.1 & 6000  & 98.08\\
                                      &      &                                                     &                                             & NumtaDB        & 85596 & 93.32\\
                                      &      &                                                     &                                             & Ekush          & 30688 & 95.68\\
                                      &      &                                                     &                                             & BDRW           & 1393  & 89.68\\
        \bottomrule
    \end{tabular}
\end{table*}

\subsection{Dimension Reduction}
Feature dimension reduction is applied to reduce the relatively redundant features resulting in an increase in the speed of the recognition process. Principal Component Analysis (PCA) is used to extract uncorrelated components by linearly transforming a high-dimensional input vector into a low-dimensional one \cite{0528pearson1901lines}. This technique was applied to reduce feature dimension and tested on eight different classifiers, resulting in the reduction of processing speed and improvement in performance \cite{0154wen2012classifier}. In another work \cite{0161wen2007handwritten}, the features, reduced using PCA, were fed to $k$NN ($k=1$) and SVM classifiers where SVM provided better performance.
PCA was also found to be useful in reducing different histogram-based features \cite{0174chenglin2009new}.

Among other feature reduction techniques, Supervised Locally Linear Embedding (SLLE) \cite{0557ridder2003supervised} was used to project the input features in a lower-dimensional feature space, which reduced the training time \cite{0198ahmed2009bangla}. However, this resulted in a slight reduction in accuracy compared to the original feature space. In another work, Bidirectional Associative Memory (BAM) \cite{0558kosko1988bidirectional} was used to reduce input features for Artificial Neural Network (ANN), which may help reduce overfitting problems by decreasing the number of hidden layers and their neurons \cite{0197hashem2014handwritten}. Reference \cite{1003Guha2019mhmoga} proposed a feature selection method named Memory-Based Histogram-Oriented Multi-objective Genetic Algorithm (M-HMOGA). This method applied Genetic Algorithm (GA) for feature selection that ensures adequate exploration of the entire search space by the use of histograms and storing the best set of candidate solutions throughout the generations. Authors claimed their method improved the performance of different classifiers and used only half of the feature set.

One interesting prospect that was seen in the existing literature is that the authors often selected the feature extraction technique in such a way that the number of features will be limited. For example, after experimenting with different cell sizes, \cite{0166wahid2021classical} selected $4\times4$ HOG cell size that maintains a significantly high accuracy using less number of features. Similarly, other literature compared different feature extraction techniques and selected zone density of dimension $8\times8$ (as opposed to zone dimensions $32\times32$, $16\times16$ and $4\times4$) \cite{0045khan2014handwritten}, uniform LBP (as opposed to basic and simplified LBP) \cite{0157hassan2015handwritten}, Histogram of Oriented Pixel Positions (as opposed to Point-Light Source-based Shadow and their combined approach) \cite{0162ghosh2021language}, celled projection (as opposed to crossings, Fourier transforms, moments, projection histograms, zoning) \cite{0164hossain2011rapid}, and contour angle technique (as opposed to gray pixel-based method, hotspot technique, black and white down scaled method) \cite{0165surinta2013comparison} to keep the number of features minimum while also achieving a respectable accuracy.

\subsection{Classification}
Majority of the ML-based works achieved the best results using Support Vector Machine \cite{0155das2012genetic, 0165surinta2013comparison, 0198ahmed2009bangla, 0005aziz2017bangla, 0161wen2007handwritten, 0166wahid2021classical, 0174chenglin2009new, 0202roy2012new, 0192rehana2017bangla, 0049choudhury2018handwritten, 0198ahmed2009bangla} and Artificial Neural Networks \cite{0190chowdhury2016towards, 0195basu2005mlp, 0197hashem2014handwritten, 0153basu2005handwritten, 0201majumdar2006mlp, 0044bhattacharya2009handwritten}. Apart from these two, other works on Bengali handwritten digit recognition employed Probabilistic Neural Networks \cite{0164hossain2011rapid}, Hierarchical Bayesian Network \cite{0194xu2008handwritten}, Bayesian Discriminant \cite{0154wen2012classifier}, $k$-Nearest Neighbor \cite{0157hassan2015handwritten}, Sparse Representation Classifier \cite{0045khan2014handwritten}, Random Forests \cite{0162ghosh2021language}, Decision Tree (DT) \cite{0173pal2006complete, 0180pal2000automatic}, Modified Quadratic Discriminant Function (MQDF) \cite{0199pal2008bangla}, etc.

\subsubsection{Support Vector Machine (SVM)}
Before the popularization of deep learning techniques, SVM \cite{0576cortes1995support} was one of the most robust techniques for handwritten digit recognition. SVM divides a dataset into two classes based on hyperplanes, maximizing the distance between those classes. For multiclass problems, such as handwritten digit recognition, One Versus One (OVO) or One Versus All (OVA) SVM is used \cite{0595duan2005which, 0596hsu2002comparison}. Considering the digit classes may not be linearly separable, a nonlinear kernel can be used to map the data into a new, higher-dimensional space in which they can be linearly separated, keeping the margin as wide as possible. There are different variants of kernel functions that can be used to build the hyperplane \cite{0155das2012genetic}. Before applying SVM, it is important to normalize the feature vector to avoid the attributes having a very large numeric range so that all features get equal importance \cite{0165surinta2013comparison}. The most commonly used kernels in Bengali digit classification are linear kernel \cite{0155das2012genetic, 0198ahmed2009bangla}, Gaussian kernel \cite{0155das2012genetic}, Polynomial kernel \cite{0005aziz2017bangla, 0155das2012genetic, 0161wen2007handwritten}, and RBF kernel \cite{0165surinta2013comparison, 0166wahid2021classical, 0174chenglin2009new, 0202roy2012new}. Reference \cite{0192rehana2017bangla} showed that linear kernels work better for a large number of features. On the other hand, if the feature set is smaller than the number of samples, then RBF and polynomial kernels were preferred. RBF kernels can outperform polynomial kernels, however, it requires higher computational resources \cite{0174chenglin2009new}.

In BHDR, SVM has performed the best using HOG features \cite{0049choudhury2018handwritten, 0166wahid2021classical, 0192rehana2017bangla} compared to other features such as LBP and Gabor. This is because HOG features preserve local pixel interactions which cannot be captured using LBP due to its limited capabilities. Combining global features with a genetic algorithm, simulated annealing, and hill-climbing approaches to extract local features have also been seen to provide good accuracy \cite{0155das2012genetic}. Reference \cite{0174chenglin2009new} showed that SVM classifiers can provide similar accuracy as other classifiers such as polynomial network classifier, class-specific feature polynomial classifier, and discriminative learning quadratic discriminant function. SVM has also been shown to provide comparable accuracy using directional edge features \cite{0161wen2007handwritten}, gradient and longest run features \cite{0202roy2012new}, global and local directional pattern \cite{0005aziz2017bangla}, curvature features \cite{0165surinta2013comparison}, and even pixel-values of the image samples \cite{0198ahmed2009bangla}. Reference \cite{0165surinta2013comparison} was the only study to consider the ensembling of four different SVM classifiers using the Unweighted Majority Voting (UMV) method to combine their results. To break the tie between the classes they used a random method. Their accuracy increased from 96.4\% to 96.8\% when UMV based ensemble technique was used.

\subsubsection{Artificial Neural Network (ANN)}
An Artificial Neural Network (ANN) works similar to the human brain, consisting of many interconnected neurons that are used to transfer information \cite{0526rosenblatt1958perceptron, 0577mcculloch1943logical}. The connections carry weights that are used to approximate a function estimating the output class from the input features. These weights are updated via adaptive learning \cite{0527rumelhart1985learning}. A commonly used variant of ANN is the multilayer perceptron (MLP), which consists of input, hidden, and output layers. The number of nodes in the input layer and the output layer corresponds to the number of features and the number of classes, respectively. Hence, in digit recognition, the number of nodes in the output layer is 10. The number of nodes in the hidden layer can be identified using grid search to find the optimal configuration \cite{1017chicco2017ten}. However, increasing the number of nodes generally results in an increase in computational complexity during training and inference.

One of the earliest works on ANN experimented with different choices of hyperparameters such as activation function, number of hidden layers, number of neurons, and batch sizes \cite{0190chowdhury2016towards}. The authors found that a single hidden layer containing 700 neurons with a ReLU activation function can achieve the highest accuracy (96.05\%) on a self-curated dataset of 70,000 images. Another work also experimented with the number of neurons in a single hidden layer \cite{0195basu2005mlp}. They achieved the highest accuracy (96.65\%) on a self-curated dataset containing 6,000 images using 65 neurons. This reduction in neurons is due to the lower number of samples in the training set. A similar pattern can be seen in \cite{0197hashem2014handwritten}, which required 52 neurons in the hidden layer in the proposed ANN architecture, trained using only 800 digit samples.

Reference \cite{0153basu2005handwritten} utilized Dempster-Shafer's (DS) theory \cite{0574dempster2008upper} to propose an ensemble technique containing two MLPs trained using different sets of features. The ensembled architecture was able to increase the accuracy of digit recognition by at least $1.4\%$ compared to the individual MLPs ($93.7\%$ and $92.62\%$). On the other hand, \cite{0201majumdar2006mlp} proposed a multi-layer NN trained using both printed and handwritten Bengali numerals, which achieved 95.7\% accuracy in recognizing handwritten digits and 99.2\% accuracy in recognizing printed digits. Reference  \cite{0044bhattacharya2009handwritten} employed three cascaded MLP classifiers provided with images of different resolution levels for digit recognition.  If the confidence of the output label is below a certain threshold, another MLP is used to combine the outputs of the three MLPs to provide a final prediction.

\subsubsection{Bayesian Network}
Bayesian Network (BN) simplifies the Bayes theorem given that both the probability distributions for the random variables and the interactions among them are subjectively described and the model can be regarded to represent ``belief" about a complex domain \cite{0578pearl1985bayesian}. Bayesian probability is the study of subjective probabilities or confidence in an outcome, as opposed to the statistical inference method. One drawback of BNs is that finding the optimal solution is NP-Hard, requiring the implementation of approximation algorithms to achieve a polynomial time solution \cite{1018cooper1990computational}.
Reference \cite{0164hossain2011rapid} implemented a Bayesian classification-based Probabilistic Neural Network (PNN) using a rapid feature extraction method, Celled Projection to compute the projections of each section as features from an image. This PNN architecture had two layers: the radial basis layer and the competitive layer. Some training is required for the estimation of the probability density function associated with each class. The performance of PNN depended on the spread factor which was chosen not more than 3 by experiment. They have also explored Feed-forward Back Propagation Neural Networks (FBPN) and $k$-NN along with PNN which achieved higher accuracy with less training time.

Reference \cite{0194xu2008handwritten} proposed a hierarchical Bayesian network where the original digit images were used directly as input rather than extracting the feature vectors. Reference \cite{0154wen2012classifier} utilized Polynomial and Gaussian kernels to extract features and a Bayesian discriminant to classify the digits selecting the optimal parameters to minimize the classification error in a given database. This technique made full use of characteristics of each class distribution such as the class mean and covariance even though it did not average the covariance matrix of all classes. In all the eigenvalues of each class, a small value threshold ($\lambda_0$) was used to substitute the smaller eigenvalues to overcome the problem of the non-existing inverse matrix for covariance matrix so that the classification error in a given database was minimized.

\subsubsection{$k$-Nearest Neighbours ($k$NN)}
$k$-Nearest Neighbours ($k$NN) \cite{0517fix1989discriminatory} is a similarity-based image classification approach that is based on the nearest training samples in the feature space. During the training phase, this algorithm stores and labels all of the training images' feature vectors. Then it assigns a label or class to a test image based on the majority vote of its $k$ nearest neighbors. The performance of this classifier in digit recognition was found to be highly dependent on the choice of $k$ and the distance metric used to calculate the distances between neighbors \cite{0164hossain2011rapid}. The distance parameter is frequently chosen as Euclidean distance, City Block distance, or Manhattan distance, and the number of neighbors is typically 3, 5, and 7 \cite{0005aziz2017bangla, 0014hoq2020comparative}. One disadvantage of $k$NN is that complex features may require significantly more time to perform inference \cite{0111nanehkaran2021analysis}. Reference \cite{0005aziz2017bangla} demonstrated that $k$NN with Euclidean distance is effective in recognizing digits from local and global directional patterns. The authors also experimented with different values of $k$ and found $k = 3$ to be the most effective, achieving an accuracy of $95.62\%$ on the CMATERdb dataset. On the same dataset, \cite{0157hassan2015handwritten} incorporated different variations of local binary patterns extracted from digits with $k$NN. Here, considering only one neighbor provided the best accuracy ($96.70\%$). Despite that, the performance of $k$NN was poor compared to SVM and ANN in more complex datasets such as NumtaDB and Ekush \cite{0166wahid2021classical, 0164hossain2011rapid}. As a result, even though $k$NN requires lesser computational resources compared to SVM and ANN, the latters are often preferred.


\subsubsection{Miscellaneous}

The zone density features of an unknown Bengali digit sample can be represented as a linear combination of the features extracted from the training samples from the same class \cite{0045khan2014handwritten}. Taking advantage of this property, \cite{0045khan2014handwritten} utilized Sparse Representation Classifiers \cite{0579wright2009robust} to build a dictionary of features from the training samples to represent Bengali digits. The authors used L1-minimization to efficiently compute the most sparse linear representation of the feature vectors of the test samples using the dictionary. Based on the representing samples, the class label was decided. The authors achieved $94\%$ accuracy on the CMATERdb dataset.


Reference \cite{0180pal2000automatic} utilized Decision Tree (DT) \cite{0600hunt1966} to generate a hierarchy of rules in order to classify Bengali digits using water overflow from reservoir-based features. The authors achieved $91.98\%$ accuracy on a self-curated dataset with 10,000 images. An improvement can be seen in the performance of the classifier if the number of digit samples is increased \cite{0173pal2006complete}. Recently, \cite{0162ghosh2021language} constructed a Random Forest Classifier \cite{0581ho1995random} by ensembling a set of decision trees that outperformed other classifiers such as Sequential Minimal Optimization (SMO) \cite{0580platt1998sequential}, Artificial Neural Network (ANN), and Logistic Regression (LR) \cite{0582verhulst1838notice} on the CMATERdb dataset achieving $98.50\%$ accuracy. The authors performed two post hoc tests, namely Nemenyi test \cite{0598nemenyi1963distribution} and Bonferroni–Dunn test \cite{0584dunnett1955multiple, 0585dunn1961multiple} to further consolidate the superiority of the classifier.

\section{Deep Learning-based Approaches}\label{sec:deep}
Traditional approaches of BHDR depended highly on extreme feature engineering, which limited the performance of the model with several constraints. However, to achieve high performance on large and challenging datasets under noisy conditions, researchers have found the remarkable generalization capacity of different Deep Learning (DL)-based algorithms extremely useful \cite{0550alom2018thehistory}. 
The DL-based techniques are capable of creating a more abstract representation of the data as the network grows deeper, by which the model can learn highly meaningful features that the ML-based methods often fail to grasp. Again, the high inter-class similarities along with diversity in writing styles make it extremely difficult to design highly efficient hand-crafted feature extractors for ML methods; which the DL-based methods can automatically learn. For these reasons, a major shift in paradigm is observed in recent works and almost all of them have focused on the DL-based pipelines.

\begin{figure}[htb]
    \centering
    \includegraphics[width=0.8\columnwidth]{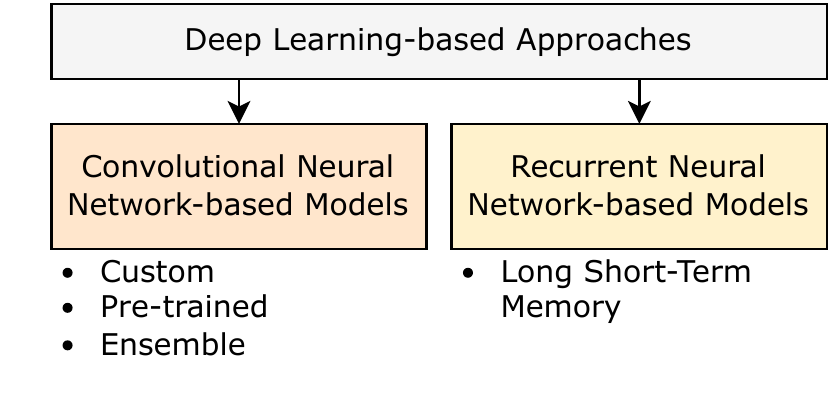}
    \caption{Taxonomy of the deep learning-based techniques used in Bengali Handwritten Digit Recognition}
    \label{fig:dl_hierarchy}
\end{figure}

This section discusses the Deep learning-based approaches of BHDR under two major categories: Convolutional Neural Networks (CNN) and Recurrent Neural Networks (RNN). These two basic categories and their subcategories are shown schematically in \figureautorefname~\ref{fig:dl_hierarchy}. Along with the discussion of the status quo, the trends in choices of different hyperparameters have been analyzed.

\subsection{CNN-Based Architectures}
\begin{figure}[htb]
    \centering
    \includegraphics[width=\columnwidth]{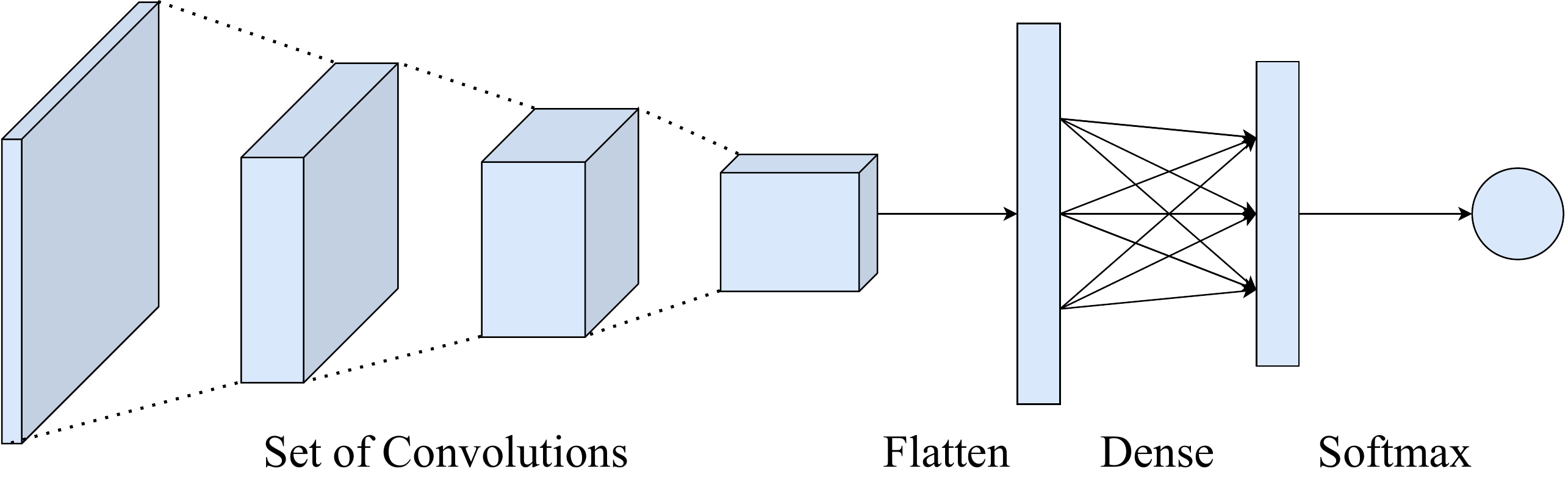}
    \caption{Generalized convolutional neural network-based architecture used in handwritten digit recognition}
    \label{fig:basicCNN}
\end{figure}

A Convolutional Neural Network (CNN) based architecture consists of different types of layers such as convolution, activation, pooling, batch normalization, dense, etc. (\figureautorefname~\ref{fig:basicCNN}).
The convolution operation produces activation maps containing meaningful patterns by applying a set of filters throughout the image. 
The output of this layer is passed through activation functions to introduce non-linearity. Moreover, the activation maps can be downsampled using pooling layers to reduce computational costs. The fully-connected layers receive a flattened activation map from the convolution block and consider further meaningful combinations of the learned features. Finally, a softmax layer is used to produce the class label.

The ability of Convolutional Neural Networks (CNN) to extract features utilizing the spatial dimension of images has been proven to be extremely useful in image classification tasks. Thus in recent times, most of the works on BHDR have been revolving around it one way or another and researchers have utilized this in different ways, such as, providing custom configurations of CNN-based models, and applying transfer learning by fine-tuning pretrained architectures, ensembling multiple models, etc. 

\subsubsection{Custom CNN-based Models}

In this section, we discuss the custom CNN-based approaches proposed in the literature. A summary of the proposed architectures is shown in \tableautorefname~\ref{tab:customCNNArchitecture}.

Reference \cite{0043akhand2015bangla} proposed one of the earliest works in Bengali Handwritten Digit Recognition that showed how the handcrafted feature extraction phase can be replaced with a CNN architecture. The authors achieved 97.93\% accuracy on a self-curated dataset consisting of 17,000 images using a CNN model consisting of two convolution layers and one fully-connected (FC) layer. However, despite working with a relatively simple dataset, the model failed to differentiate the inter-class similarities of certain classes such as ($\bnzero$, $\bnthree$), ($\bnfive$, $\bnsix$), etc. 
A similar CNN model was used by \cite{0189chowdhury2016towards} on a self-curated dataset of 70,000 images containing samples from different age and gender groups. An exhaustive grid search was applied to find the appropriate set of hyperparameters. Although the authors achieved a low validation error of 1.22\% with such a simple model with two convolution layers, the likely reason might be due to the absence of challenging images in the dataset.

In another work, \cite{0029akhand2016convolutional} applied the same architecture proposed in  \cite{0043akhand2015bangla} with the same set of hyperparameters on 22,000 images of the ISI-HBN dataset and achieved 98.98\% accuracy. They utilized rotation-based artificially generated patterns, which resulted in the improvement of performance compared to other works on the dataset till that time. However, the merit of this technique is questionable, since the authors applied a fixed degree of rotation to the images to create the patterns and reported the best result observed by repeating several such experiments. This is equivalent to hand fitting for the testing set. Moreover, the effect of other augmentation techniques was unexplored. Using the same architecture, the authors published another work exploring the possibility of Bangla-English mixed numeral recognition \cite{0038akhand2016convolutional}. However, the authors did not provide any directions on separating similar Bengali and English digits (For example, Four in Bengali `$\bnfour$' and Eight in English `$8$').

In \cite{0039akhand2016multiple}, the authors experimented with model ensembling techniques to determine its effect on the recognition performance. They trained three separate instances of the same CNN architecture used in \cite{0043akhand2015bangla}; one using the original images of the ISI-HBN dataset; one using the clockwise rotated version of the dataset (within a range of 10\degree~to 40\degree) and one using counterclockwise rotated version (within a range of 10\degree~to 40\degree). The authors reported an accuracy of 98.80\% using the ensembled model. However, instead of training the three instances, a single model could be trained with all three of the variations of the dataset, which has been found to produce slightly higher accuracy in \cite{0026akhand2018convolutional}. This helps to reduce the time complexity during training and inference. Moreover, the effects of other augmentation techniques were unexplored.

Reference \cite{0152sharif2017evil} proposed an image augmentation method on the ISI-HBN dataset with a CNN model, which achieved 99.42\% validation accuracy. The augmentation method downsamples the images into lower dimensions and then upsamples them again into the original dimension. This introduces a set of blurred images into the training set, reducing the overall quality, and enabling the model to learn to recognize poor-quality images. The best result was achieved by applying their proposed augmentation along with random rotation and shifting.
Applying the proposed augmentation method, the authors achieved an increment of 0.17\% accuracy compared to the 99.25\% achieved by only applying random rotation and shifting. However, the effectiveness of this augmentation technique is debatable considering that a similar effect can be achieved by applying random blur on images, which will have a lesser computational requirement.

Recently, \cite{0021sufian2020bdnet} proposed a densely connected deep CNN architecture, inspired by DenseNet architecture \cite{0280huang2017densely}, called `BDNet', which achieved 99.78\%  accuracy on the ISI-HBN dataset. The input is passed through a series of `Dense' and `Transition' blocks consisting of several 2D convolutions (Conv2D), ReLU, and Batch Normalization (BN) layers. The performance of this model was also tested on a self-curated dataset of 1,000 images where the model achieved an accuracy of 98.80\%.


Reference \cite{0050sharif2016hybrid} presented a hybrid model combining both hand-crafted and automatic feature extraction techniques that used the ISI-HBN dataset for training and validation sets, and the CMATERdb 3.1.1 dataset as the test set. The images were passed through a HOG feature extractor in one stream and through a two-layer CNN architecture in another. Then, the extracted features from both streams were combined in a larger feature vector and passed through an FC layer. This hybrid architecture successfully achieved 99.02\% and 99.17\% accuracy on the validation and test set, respectively. At that time, this was the state-of-the-art result. Since the test set consisted of a different dataset, the achieved result proved the generalization capability of the model in a completely unseen scenario. However, even though the hybrid model outperformed the CNN-only model mentioned in the paper, the CNN architecture was fairly simple. And this simple CNN-only architecture alone was able to achieve an accuracy of 98.87\% compared to the 99.02\% of the hybrid model on the ISI-HBN dataset. That means HOG had a minor contribution in increasing the accuracy, which might have been achieved by tweaking the layers in the CNN stream. Moreover, working with the HOG features might not turn out to be useful in challenging datasets like NumtaDB which contains lots of noise and variation \cite{0529aslam2020cnn}.

Authors in \cite{0030shopon2016bangla} experimented with both the ISI-HBN and CMATERDB 3.1.1 datasets, where they formed four combinations of the train-test sets. The proposed pipeline consisted of Autoencoders and CNN which was able to provide high performance on all four of the combinations. The autoencoders helped in unsupervised pretraining and the CNN part utilized the learned weights for further extraction of meaningful features.
In another work, the authors used the same architecture along with a newly introduced augmentation technique named `Blocky Artifact' and achieved slight improvement in the performance \cite{0033shopon2017image}. A similar Autoencoder-CNN-based approach was proposed by \cite{0183ahmed2016stacked}. However, since the encoder consisted of simple feed-forward layers, the model produced lower performance compared to the CNN-only approaches.

Another work on CMATERdb 3.1.1 dataset introduced a CNN model consisting of two convolutions and two FC layers \cite{0004alom2017handwritten}. The ability of the convolution layers in generating robust features invariant to shifting, rotation, scaling, etc. enabled the model to achieve an accuracy of 98.78\%. Inspired by LeNet architecture \cite{0505lecun1989backpropagation}, authors in \cite{0016saha2019bangla} proposed a model consisting of three convolution layers, three average pooling, and one fully connected layer. The model achieved $97.60\%$ accuracy on the same dataset. However, the accuracy can be improved with deeper networks and by applying more data augmentation techniques.

\begin{table*}[tb]
    \centering
    \caption{Performances of the Custom Convolutional Neural Network-based Architectures}
    \label{tab:customCNNArchitecture}
    \begin{tabular}{C{1.4cm}  C{0.7cm} L{5cm} C{2.5cm}  C{1.3cm} }
        \toprule
        \textbf{Reference} & \textbf{Year} & \textbf{Number of Layers and Filters} & \textbf{Dataset} & \textbf{Accuracy (\%)}\\
        \midrule
        \cite{0043akhand2015bangla} & 2015  & 2x Conv (6, 12) and 1x FC &  Self-curated & $97.93$\\
        \cite{0189chowdhury2016towards} & 2016 & 2x Conv (35, 45) and 1x FC &  Self-curated & $98.78$\\
        \cite{0029akhand2016convolutional} & 2016 & 2x Conv (6, 12) and 1x FC &  ISI-HBN & $98.98$ \\
        \cite{0038akhand2016convolutional} & 2016 & 2x Conv (6, 12) and 1x FC &  ISI-HBN & $98.45$ \\
        \cite{0039akhand2016multiple} & 2016 & 2x Conv (6, 12) and 1x FC &  ISI-HBN & $98.80$      \\
        \cite{0152sharif2017evil} & 2017  & 5x Conv (32, 32, 32, 32, 32) and 1x FC &  ISI-HBN & $99.42^*$\\
        \cite{0021sufian2020bdnet} & 2020 & Transition blocks and Dense Blocks & ISI-HBN & 99.78 \\
        
        \cite{0050sharif2016hybrid} & 2016  & HOG Features with 2x Conv (32, 32) and 1x FC &  ISI-HBN & $99.02^*$\\
                                    &       &                                              & CMATERdb 3.1.1 & $99.17$\\
        \cite{0030shopon2016bangla} & 2016 & Autoencoders with 3x Conv (32, 32) and 1x FC & ISI-HBN & $98.29$\\
                                    &      &                                              & CMATERdb 3.1.1 & $99.50$\\
        \cite{0026akhand2018convolutional} & 2017 & 2x Conv (6, 12) and 1x FC &  ISI-HBN & $98.98$ \\
                                           &      &                           & CMATERdb 3.1.1 & $99.65$\\

        \cite{0004alom2017handwritten} & 2017 & 2x Conv (32, 64) and 2x FC & CMATERdb 3.1.1 & $97.60$\\
        \cite{0016saha2019bangla} & 2019 & 3x Conv (32, 64, 128) and 2x FC & CMATERdb 3.1.1 & $97.60$\\
        \cite{0008paul2018image} & 2018 & 2x Conv (32, 64) and 1x FC & NumtaDB & $91.30$\\
        \cite{0001shawon2018bangla} & 2018 & 6x Conv (2x 32, 2x 128, 2x 256) and 2x FC & NumtaDB & $92.72$\\
        \cite{0003noor2018handwritten} & 2018 & Model 1: 12x Conv (2x 32, 2x 64, 2x 128, 3x 256, 521, 2x 512) and 2x FC. Model 2: 9x Conv (2x 32, 2x 64, 2x 128, 3x 256) and 2x FC & NumtaDB & $96.79$\\
        \cite{0013mahmud2020deepbanglanet} & 2020 & Series of Analogous and Changer blocks (13x Conv) and 2x FC & NumtaDB & $99.43$\\
        \cite{0025razik2017sustbhand} & 2017 & 2x Conv (32, 64) and 1x FC & Self-curated & $99.40$\\
        \cite{0015sikder2020bangla} & 2020  & 13x Conv (2x 32, 3x 64, 2x 128, 2x 256, 2x 384, 2x 512) and 3x FC & BHAND & $99.44^*$\\
        \cite{0006rabby2019bangla} & 2019 & 4x Conv (2x 32, 2x 64) and 2x FC & ISI-HBN & $99.58$\\
                                   &      &                                  & CMATERdb 3.1.1 & $92.65$\\
                                   &      &                                  & BLI & $98.65$\\
        \cite{0012islam2019sankhya} & 2019 & 6x Conv (2x 28, 2x 128, 2x 256) and 2x FC & Ekush & $99.71$\\
                                    &      &                                           & CMATERdb 3.1.1 & $99.25$\\
                                    &      &                                           & NumtaDB & $98.94$\\
                                    &      &                                           & BDRW & $96.65$\\

        \bottomrule
        \multicolumn{4}{l}{\footnotesize ${}^*$ denotes validation accuracy}
    \end{tabular}
\end{table*}

The introduction of the NumtaDB dataset presented the researchers with new challenges to handle highly diversified and heavily augmented noisy images. Reference \cite{0008paul2018image} demonstrated how careful preprocessing techniques like cropping, noise removal, thresholding, color inversion, etc. can boost the accuracy and also enable lighter architectures to achieve better performance in this dataset. The preprocessed dataset was fed into several ML and DL-based algorithms and resulted in improved performance compared to using the original dataset. The work also showed that CNN outperforms traditional ML-based techniques. 
On the same dataset, \cite{0001shawon2018bangla} proposed a six-layer CNN architecture to tackle the challenging NumtaDB dataset and achieved an accuracy of 92.72\%. The performance of the network was validated in other English datasets as well. However, the model performed poorly on highly augmented images. The authors achieved 96.44\% accuracy on the non-augmented portion of the dataset. Hence, efficient preprocessing and augmentation techniques can be undertaken to uplift the overall performance. Reference \cite{0003noor2018handwritten} achieved 96.79\% accuracy on this dataset by ensembling two CNN architectures consisting of 12 and 9 convolutional layers respectively. The number of samples in the training set was increased by applying different augmentation techniques to tackle the noisy images. This resulted in a performance boost compared to the model trained with non-augmented images. However, the 12-layer CNN architecture was able to achieve 96.33\% accuracy, which is comparable to the 96.73\% accuracy achieved by the ensemble model. This 0.46\% improvement comes at a cost of increased computational resources used by the second model. 

One of the most recent works on the NumtaDB dataset has proposed a custom CNN model named `DeepBanglaNet’ which introduces two special types of operations named `Analogous’ and `Changer’ block for better recognition \cite{0013mahmud2020deepbanglanet}. The Analogous block extracts features from a deeper level keeping the dimension of the activation map the same and can also ensure proper gradient flow using the residual connections. On the other hand, the Changer block can modify the dimension of the feature map to extract more generalized features. This work currently holds the state-of-the-art position with an accuracy value of 99.43\%.

Reference \cite{0006rabby2019bangla} proposed a lightweight model consisting of 4 convolutional layers, which achieved a competitive accuracy with a faster execution time. The authors achieved an accuracy of 99.58\%, 98.58\%, and 92.65\% on the ISI-HBN, BanglaLekha-Isolated, and CMATERdb 3.1.1 datasets, respectively. However, the authors tampered with the dataset by fixing some of the incorrect labelings and deleting some images which might have boosted the performance of the models. This makes it difficult to compare it with other works on these datasets. 
Reference \cite{0012islam2019sankhya} compared the performance of different ML algorithms such as $k$NN, SVM, Random Forest, Multilayer Perceptron (MLP), and a custom-designed CNN architecture on the Ekush, CMATERdb, NumtaDB, and BDRW datasets. All the algorithms were fine-tuned to get the best possible hyperparameters. The best performance was achieved by the six-layer CNN model inspired by the LeNet architecture \cite{0505lecun1989backpropagation}. The CNN model outperformed the traditional ML algorithms by at least 4\% across multiple datasets. Yet, there was room for improvement in the performance of the NumtaDB and BDRW datasets as shown in \tableautorefname~\ref{tab:customCNNArchitecture}. Specifically, all the algorithms faced difficulties recognizing the samples of the BDRW dataset due to its high variance and diversified background.

Reference \cite{0025razik2017sustbhand} achieved 99.4\% accuracy on a self-curated dataset using a CNN model consisting of two convolutions and one FC layer. Such a high accuracy with this simple model can be attributed to the absence of challenging and diversified digit samples in the dataset. In another work \cite{0015sikder2020bangla}, authors proposed a CNN model containing six blocks, each consisting of convolution, pooling, and local normalization layers. The authors achieved 99.44\% validation accuracy on the BHAND dataset, outperforming some of the state-of-the-art CNN architectures in the literature.

\subsubsection{Pre-trained CNN-based Models}

Pre-pretrained CNN-based architectures are used in BHDR via transfer learning. Transfer learning refers to the use of an architecture that is trained to extract features using the training data of one domain and reused as a feature extractor in another domain \cite{0554tan2018asurvey}. Most of the transfer learning-based feature extractors that are used in image classification, namely AlexNet \cite{0521krizhevsky2012imagenet}, VGG \cite{0520simonyan2015very}, GoogLeNet \cite{0524szegedy2015going}, ResNet \cite{0518he2016deep}, Xception \cite{0159chollet2017xception} originated from the ImageNet Large Scale Visual Recognition Challenge (ILSVRC) \cite{0522russakovsky2015imagenet}. Along with them, MobileNet \cite{0523howard2017mobilenet} and CapsuleNet \cite{0525sabour2017dynamic} have also been used in BHDR. Studies have shown that fine-tuning these architectures for downstream tasks such as digit and character recognition can achieve satisfactory performance \cite{0555pramanik2019astudy, 0556ghosh2021performance}. A summary of the proposed architectures are shown in \tableautorefname~\ref{tab:transferArchitecture}.

\begin{table*}[tb]
    \centering
    \caption{Performances of the Transfer Learning and Ensemble-based Architectures}
    \label{tab:transferArchitecture}
    \begin{tabular}{C{1.4cm} C{0.7cm} C{3.1cm} C{1.7cm} C{2cm} C{2.5cm} C{1.3cm}}
        \toprule
        \textbf{Reference} & \textbf{Year}  & \textbf{Best Performing Model} & \textbf{Number of Layers} & \textbf{Parameter Count (millions)} & \textbf{Dataset} & \textbf{Accuracy (\%)}\\
        \midrule
        \cite{0009zunair2018unconventional} & 2018 & VGG16 & 21 & 134.7 & NumtaDB & 97.09\\
        \cite{0022basri2020bangla} & 2020 & AlexNet & 8 & 61 & NumtaDB & 99.01\\
        \cite{0024rahman2019convolutional} & 2019 & Modified VGG-11 & 32 & 7.7 & NumtaDB & 99.25\\
        & & & & & ISI-HBN & 99.80\\
        & & & & & CMATERdb 3.1.1 & 99.66\\
        \cite{0007mamun2018bangla} & 2018 & Ensemble of Xception models & 126 & 126 & NumtaDB & 96.69\\
        \cite{0002hasan2018recognition} & 2018 & Ensemble of ResNet-34 and ResNet-50 & 34 and 50 & 21.5 and 23.9 & NumtaDB & 99.34\\
        \bottomrule
    \end{tabular}
\end{table*}

Although the pretrained models from ImageNet Challenge are available for quite a while, \cite{0009zunair2018unconventional} was one of the first to study the use of transfer learning methods in BHDR. The authors used the VGG-16 network and achieved an accuracy of 97.09\% on the NumtaDB dataset. The number of trainable parameters was reduced by freezing some of the layers of the architecture. The authors experimented by freezing five different combinations of layers from the 21 layers available in VGG-16. They are: freezing all layers except the final one, freezing none of the layers, freezing only the softmax layer, freezing everything except layer 16-20, and freezing only layer 16-20. The fifth combination achieved the highest accuracy of 97.09\%. Freezing layers in this manner reduced the number of trainable parameters up to half of the original model.

On the same dataset, \cite{0022basri2020bangla} experimented with four pretrained architectures, namely, AlexNet, MobileNet, Goog\-LeNet, and CapsuleNet. Among them, AlexNet was able to achieve the highest accuracy of 99.01\%. According to the authors, although AlexNet had the highest number of parameters among the architectures considered, it required the least computation time to recognize digits. 

Reference \cite{0024rahman2019convolutional} compared the performance of LeNet-5, ResNet-50, VGG-11, and VGG-16 and found the modified VGG-11 architecture to have the best performance achieving 99.80\%, 99.66\%, 99.25\% accuracy on ISI-HBN, CMATERdb, and NumtaDB datasets, respectively. They introduced batch normalization and zero-padding layers to the vanilla VGG-11 architecture. The number of filters in the convolution layers and the neurons in the fully-connected layers were tuned to reduce the number of trainable parameters. The modified architecture had 7.7M parameters, which is significantly less compared to the 28M parameters in the vanilla architecture.

\subsubsection{Ensemble of Multiple Models}

Combining the results obtained by individual classifiers in an ensemble architecture can be done in several manners. 
A few of the usual approaches are max-voting, probability average, etc. In a max-voting ensemble, among the class labels obtained from the individual classifiers, the highest occurring label is selected as the final label for the input sample. In a classifier network, the probabilities for each label for a particular sample are passed through an argmax layer which provides this class label. These probabilities are utilized by the probability averaging ensemble. The class-wise probability values across all models are aggregated for each class and divided by the total number of classifier networks. This provides the average probability for each class, combining the contribution from all the models. Then, the probability values are passed through an argmax layer to predict the final class label. A summary of the ensemble-based approaches can be found in 
\tableautorefname~\ref{tab:transferArchitecture}.

Reference \cite{0007mamun2018bangla} used a max-voting ensemble architecture of three pretrained Xception architectures on the NumtaDB dataset. According to the depthwise separable convolution blocks used in Xception architecture were suitable for reducing the parameter count. Additionally, considering the large size of the model, dropout and L2 regularization were used to avoid overfitting. 
To train the individual architectures, different preprocessed training images were used as input. The authors achieved an accuracy of 96.694\% using the ensemble architecture, whereas the individual models achieved 96.499\%, 96.305\%, and 96.125\%. Nevertheless, the authors reported weighted average accuracy, where the significantly distorted images of the test set were weighted significantly lower, misrepresenting the actual performance of the models.

Reference \cite{0002hasan2018recognition} used an ensemble architecture of the popular pretrained models, ResNet34 and ResNet-50, that achieved an accuracy of 99.3359\% on the NumtaDB dataset. Each of the models individually achieved an accuracy of 98.686\% and 99.094\%, respectively. To achieve the final accuracy, the authors used a weighted voting ensemble of six architectures, where five of which were ResNet34 and one of them was ResNet50. Each model had a different set of hyperparameters, training, and validation images. However, the accuracy achieved by a single ResNet-50 being close to the ensembled model negates the necessity of the huge increase in computational resources to achieve a meager 0.2419\% increase in accuracy. However, both of the works have ensembled the variants of the same baseline CNN model. The impact of ensembling different types of architecture can also be investigated. Analysis should be conducted on how such ensembling techniques impact the computation and time constraints compared to the performance improvement it provides.

\subsubsection{Choice of Hyperparameters}
Hyperparameters are the parameters whose values are set before the learning process and not updated during training. Properly tuning them can result in faster convergence to the global minima  \cite{0540liao2022empirical}. The trend in hyperparameters selection during the training of the CNN models is shown in \tableautorefname~\ref{tab:customCNNArchitectureHyper}.

\begin{table*}[tb]
    \centering
    \caption{Hyperparameters of the CNN-based Architectures}
    \label{tab:customCNNArchitectureHyper}
    \begin{tabular}{C{1.4cm} C{1.5cm} C{1.2cm} C{3cm} C{1.1cm} C{1.1cm} C{1.3cm} C{2.5cm}}
        \toprule
        \textbf{Reference} & \textbf{Input Dimension} & \textbf{Pooling} & \textbf{Dropout} & \textbf{Batch Size} & \textbf{Epoch} & \textbf{Learning Rate} & \textbf{Optimizer}\\
        \midrule
        \cite{0043akhand2015bangla} & $28\times28\times1$ & $-$ & N/A & $-$ & 80 & $-$ & $-$\\
        \cite{0189chowdhury2016towards} & $32\times32\times1$ & Max & N/A & $-$ & $-$ & $-$ & $-$\\
        \cite{0029akhand2016convolutional} & $28\times28\times1$ & $-$ & N/A & 100 & 310 & $-$ & $-$\\
        \cite{0038akhand2016convolutional} & $28\times28\times1$ & Average & N/A & 50 & 300 & 1 & $-$\\
        \cite{0039akhand2016multiple} & $28\times28\times1$ & $-$ & N/A & 50 & 500 & $-$ & $-$\\
        \cite{0026akhand2018convolutional} & $28\times28\times1$ & $-$ & N/A & 50 & 300 & 1 & $-$\\
        \cite{0152sharif2017evil} & $28\times28\times1$ & Max & 0.5 after every conv and FC layer & 100 & 100 & $-$ & Adadelta\\
        \cite{0021sufian2020bdnet} & $32\times32\times3$ & Average & 0.9 & 32 & 150/125 & 0.009 & SGD\\
        \cite{0050sharif2016hybrid} & $28\times28\times1$ & Max & N/A & $-$ & 100 & $-$ & Adadelta\\
        \cite{0004alom2017handwritten} & $32\times32\times1$ & Max & 0.5 in between convolution layers & $-$ & 30 & $-$ & $-$\\
        \cite{0016saha2019bangla} & $32\times32\times3$ & Max & N/A & $-$ & 40 & $-$ & Adam\\
        \cite{0008paul2018image} & $28\times28\times1$ & $-$ & 0.5 in between convolution layers & 64 & 30 & 0.003 & $-$\\
        \cite{0001shawon2018bangla} & $32\times32\times1$ & Max & 0.2 after FC1 & 64 & 30 & 0.001 & Adam\\
        \cite{0003noor2018handwritten} & $64\times64\times1$ & Max & 0.2 before softmax & $-$ & 30 and 6 & 0.001 and 0.0001 & SGD in Model1 and Adam in Model2\\
        \cite{0013mahmud2020deepbanglanet} & $128\times128\times3$ & Average & N/A & 32 & 100 & 0.0001 & Adam \\
        \cite{0025razik2017sustbhand} & $28\times28\times1$ & Max & N/A & $-$ & 4000 & $-$ & $-$\\
        \cite{0015sikder2020bangla} & $32\times32\times1$ & Max & 0.5 after first 2 FC layers & $-$ & 19550 & 0.001 & RMSprop\\
        \cite{0006rabby2019bangla} & $28\times28\times1$ & Max & 0.25 after convs and 0.5 before softmax & 86 & 30 & 0.001 with LR Reduction & Adam\\
        \cite{0012islam2019sankhya} & $28\times28\times1$ & Max & 0.2 before output & $-$ & $-$ & $-$ & Adam\\
        
        \cite{0007mamun2018bangla} & $71\times71$ & Max & 0.25 before every residual block & 124 & $-$ & 0.001 & Adam\\
        \cite{0009zunair2018unconventional} & $41\times41$ & Max & N/A & 16 & 50 & 0.0001 & AdaDelta\\
        \cite{0002hasan2018recognition} & $-$ & $-$ & N/A & $-$ & 2-3 & 0.0055 - 0.01 & $-$\\
        \cite{0024rahman2019convolutional} & $32\times32$ & Max, Global Average & 0.2 after 6th conv layer, 0.25 and 0.4 respectively after the two FC layers & 128 & 200 & 0.001 & Adam\\
        
        \bottomrule

        \multicolumn{8}{l}{\footnotesize $-$ denotes information was not available in the cited literature}\\
        \multicolumn{8}{l}{\footnotesize N/A denotes dropout layer was not used in the cited literature}\\
    \end{tabular}
\end{table*}



\paragraph{Input Dimension}
A common practice that has been observed throughout the existing literature in this domain is to convert the input images of RGB color space into binary or grayscale to reduce the channel dimension  \cite{0001shawon2018bangla, 0004alom2017handwritten, 0006rabby2019bangla, 0012islam2019sankhya, 0025razik2017sustbhand, 0029akhand2016convolutional, 0038akhand2016convolutional, 0043akhand2015bangla, 0050sharif2016hybrid}. This results in a reduction of computational cost during training and inference without affecting the accuracy. 
Another common practice was to reduce the input dimension at the very beginning of the model. Here, the input dimension should be selected in such a way that it contains as fewer pixels as possible but still maintains the clarity to convey enough information to the model for digit recognition. Taking the resolution too high can increase the computational cost, and too low can distort the digits. So this trade-off should be considered when choosing the input size.


\paragraph{Number of Convolution Layers}
The main building block of a CNN architecture is the convolution layer which applies a set of filters to create activation maps containing meaningful information utilizing the spatial information of the digit image. Starting with the low-level features produced by the earlier layers, the deeper convolution layers combine them to produce a deeper meaning of the data \cite{0530rawat2017deep}.
In the BHDR literature, the number of convolution layers in different models varied with the complexity of the dataset. If the dataset is simpler, i.e., had comparatively simple backgrounds and fewer variations in handwriting, meaningful features could be learned with only a few convolution layers \cite{0004alom2017handwritten, 0029akhand2016convolutional, 0038akhand2016convolutional, 0039akhand2016multiple, 0025razik2017sustbhand, 0043akhand2015bangla, 0050sharif2016hybrid}. Besides, some works manually cropped the digit regions, which also enabled the simpler CNN models to gain higher accuracy \cite{0029akhand2016convolutional, 0038akhand2016convolutional, 0039akhand2016multiple}. However, such rigorous preprocessing is not recommended considering the variety in real-life handwritten numerals. As the size, variation, and background of the samples in the dataset got diversified, the authors used complex models with more layers. For example, since the NumtaDB dataset contains different challenging samples, most of the works have utilized CNN models with more than six layers to train their models to achieve acceptable performance on this dataset \cite{0001shawon2018bangla, 0003noor2018handwritten, 0012islam2019sankhya, 0015sikder2020bangla, 0021sufian2020bdnet, 0023alnasim2020comparative, 0020hasan2020new, 0013mahmud2020deepbanglanet}.

\paragraph{Number of Fully Connected (FC) Layers}
The fully-connected layers take the flattened output produced by the CNN block, representing meaningful features extracted from the input. The number of FC layers varied in the BHDR literature based on the size of the flattened layer. If the number of nodes in the flatten layer was larger, one or two dense layers were added before the final prediction layer to consider further combinations of these features for robust digit recognition \cite{0001shawon2018bangla, 0003noor2018handwritten, 0004alom2017handwritten, 0006rabby2019bangla, 0012islam2019sankhya, 0013mahmud2020deepbanglanet, 0015sikder2020bangla, 0023alnasim2020comparative, 0020hasan2020new, 0152sharif2017evil}. However, if the number of nodes in the flatten layer was smaller, it was directly fed to the final output layer \cite{0025razik2017sustbhand, 0029akhand2016convolutional, 0038akhand2016convolutional, 0039akhand2016multiple, 0043akhand2015bangla, 0186mukhoti2020handwritten}. 

\paragraph{Pooling and Batch Normalization Layers}
Pooling layers reduce the number of parameters and computations in the network by downsampling the spatial size of the activation. It works on each feature map independently. In most of the BHDR models, the authors have used a Pooling layer after every one or two convolution layers. It was observed that if the number of convolution layers was higher, one pooling layer was used after every two layers \cite{0001shawon2018bangla, 0003noor2018handwritten, 0012islam2019sankhya, 0015sikder2020bangla}. Otherwise, pooling was applied after every convolution layer \cite{0026akhand2018convolutional, 0043akhand2015bangla}. Applying pooling layers after every convolution layer in deeper models aggressively downsamples the feature maps, hampering the overall performance of the digit recognition models. Max pooling was the most popular choice in the literature, but a few works also used average pooling \cite{0013mahmud2020deepbanglanet, 0038akhand2016convolutional}. Some works also used Batch Normalization layers \cite{0586ioffe2015batch} before convolution layers to speed up the learning process by normalizing the input layers \cite{0001shawon2018bangla, 0015sikder2020bangla}.

\paragraph{Size of Filters}
Filters are the learned weights of CNN architecture that are applied over the input to capture patterns. Most of the existing CNN architectures for Bengali numeral recognition have gradually increased the number of filters as they went towards the deeper layers to capture combinations of the features as much as possible. Common choices for filter size were $3\times3$ and $5\times 5$. The use of bigger filters incurs more computation, while also aggressively downsampling the data. On the contrary, a filter of any size can be mimicked using a consecutive number of small filters like $3\times3$, resulting in less amount of computation \cite{0535luo2016understanding}. 
In the case of a larger input dimension, a $7\times7$ filter has also been utilized to rigorously downsample the feature space only in the first layer \cite{0013mahmud2020deepbanglanet}. The authors set the size of the filter in the pooling layer to be $2\times 2$.

\paragraph{Dropout Rate}
The dropout technique is used to introduce regularization in the model by randomly switching off nodes during the training phase \cite{0551srivastava2014dropout}. Some of the works used dropout layers before the final output layer \cite{0001shawon2018bangla, 0003noor2018handwritten, 0012islam2019sankhya, 0020hasan2020new}. In some other works, if the model contained multiple FC layers, dropout was used after each of them with different probabilities \cite{0015sikder2020bangla}. In both scenarios, the dropout layer was not used in the intermediate layers. The possible reason for omission might be to preserve high-level features in the intermediate layers \cite{0536garbin2020dropout}. 
However, several works also achieved a positive result by utilizing dropout after both the convolution layers and FC layers. For example, \cite{0006rabby2019bangla} applied a dropout of 25\% after every two convolution layers and 50\% before the final FC layer. \cite{0152sharif2017evil} applied 50\% dropout after every convolution and FC layers. \cite{0004alom2017handwritten} and \cite{0008paul2018image} also found the usage of dropout useful in between the convolution layers.

\paragraph{Optimizer}
Optimizers play a vital role to update the weights of the model in order to reduce the overall loss. In the present literature, the most popular choice has been the Adam optimizer \cite{0507diederik2015adam}, since it provides faster convergence in most of the cases \cite{0001shawon2018bangla, 0006rabby2019bangla, 0012islam2019sankhya, 0013mahmud2020deepbanglanet,  0016saha2019bangla}. However, other optimizers have also shown promising results, such as Adadelta in \cite{0009zunair2018unconventional, 0050sharif2016hybrid, 0152sharif2017evil}, RMSprop in \cite{0015sikder2020bangla}, and SGD in \cite{0152sharif2017evil}. In the ensemble model proposed in \cite{0003noor2018handwritten}, one model was trained with Adam, and another model was trained with an SGD optimizer.

\paragraph{Activation Function}
The activation function helps the performance of the architecture by introducing non-linearity into the output of a neuron \cite{0552ding2018activation}. In the existing literature of BHDR, ReLU \cite{0513nair2010rectified} has been widely used as the activation function \cite{0001shawon2018bangla, 0013mahmud2020deepbanglanet, 0016saha2019bangla, 0020hasan2020new}. It removes the negative values from the activation map by setting them to zero. Thus, it increases the nonlinear properties of the decision function and removes the chances of vanishing gradient without affecting the receptive fields of the convolution layers \cite{0537dubey2019comparative}. 

\paragraph{Choice of Learning Rate}
Learning Rate (LR) helps the optimization algorithm to control the step size for each iteration along the downward slope. Large learning rates might not hold for deep neural architecture since the model learns suboptimal solutions and might not converge to the global minima \cite{0553wu2019demystifying}. Using an adaptive learning rate might be handy in such scenarios \cite{0538wen2021convolutional}. Some popular choices of LR in BHDR literature are 0.001 \cite{0001shawon2018bangla, 0006rabby2019bangla, 0007mamun2018bangla, 0015sikder2020bangla, 0024rahman2019convolutional}, 0.0001 \cite{0009zunair2018unconventional, 0013mahmud2020deepbanglanet, 0020hasan2020new, 0023alnasim2020comparative}, etc. 
Reference \cite{0006rabby2019bangla} manually monitored the results up to 30 epochs, manually reduced the learning rate, and continued training for 5-10 more epochs. This process was repeated a couple of times.
Reference \cite{0021sufian2020bdnet} used an LR of 0.009 till 80 epochs and then reduced it by a factor of 0.15 until the model converged. Reference \cite{0003noor2018handwritten} used two different LRs (0.001 and 0.0001) to train the ensembled CNN networks.

\paragraph{Batch Size}
Batch size denotes the number of samples that will be passed through the model before updating weights once. Mini-batch is usually preferred for large datasets. If the optimization algorithms work with the entire batch for every update, the overall training process slows down \cite{0539kandel2020effect}. In addition, smaller batch size is used due to the limitation of memory. Batch size does not contribute much to the performance, apart from increasing the training speed. Some of the common choices of batch sizes in the present literature are 32 in \cite{0013mahmud2020deepbanglanet, 0021sufian2020bdnet}, 50 in \cite{0038akhand2016convolutional, 0039akhand2016multiple}, 64 in \cite{0001shawon2018bangla, 0020hasan2020new}, 86 in \cite{0006rabby2019bangla}, 100 in \cite{0152sharif2017evil}, etc.

\paragraph{Number of Epochs}
The number of epochs denotes the count of completed passes through the dataset by the optimizer during training. Choosing the optimal value for this hyperparameter depends highly on the architecture \cite{0540liao2022empirical}. 
An inverse proportional relationship between model depth and the number of epochs can be seen in the existing BHDR literature.
For example, models with more than six convolution layers required 30 epochs to converge \cite{0001shawon2018bangla, 0003noor2018handwritten, 0020hasan2020new, 0023alnasim2020comparative}. However, several authors trained deeper models for even more epochs, such as \cite{0016saha2019bangla} trained for 40 epochs, \cite{0043akhand2015bangla} for 80 epochs, \cite{0013mahmud2020deepbanglanet, 0152sharif2017evil} for 100 epochs. On the other hand, if the model is shallow, it took longer epochs to converge. For example, \cite{0038akhand2016convolutional}, \cite{0029akhand2016convolutional} and \cite{0039akhand2016multiple} trained the model for around 300 epochs to converge. Similarly, \cite{0025razik2017sustbhand} trained the model for 4000 epochs since the model had to learn from around 80000 images with a simple 2-layer CNN. In the case of transfer learning-based approaches, it is observed that they usually require a lower number of epochs than custom CNN-based architecture that is trained from scratch. This is because these architectures are already pretrained to extract features from a huge set of image samples. 

Instead of training the model for a fixed number of epochs, the early stopping approach can be considered to limit the maximum number of epochs required \cite{0541bai2021understanding}. In this method, regardless of the number of epochs selected, the training is stopped as soon as the model converges. Such an approach was adopted by \cite{0050sharif2016hybrid}, where the authors initially set the number of epochs to be 100, however, the training was stopped at 55 epochs as soon as it converged. Similarly, \cite{0021sufian2020bdnet} set the number of epochs to be 150, but the model converged at 125 epochs. Another approach can be varying the number of epochs based on the learning rate, giving the model enough opportunity to learn as long as possible \cite{0006rabby2019bangla}.

\subsection{Recurrent Neural Network-based Architectures}
\begin{figure}[htb]
    \centering
    \includegraphics[width=\columnwidth]{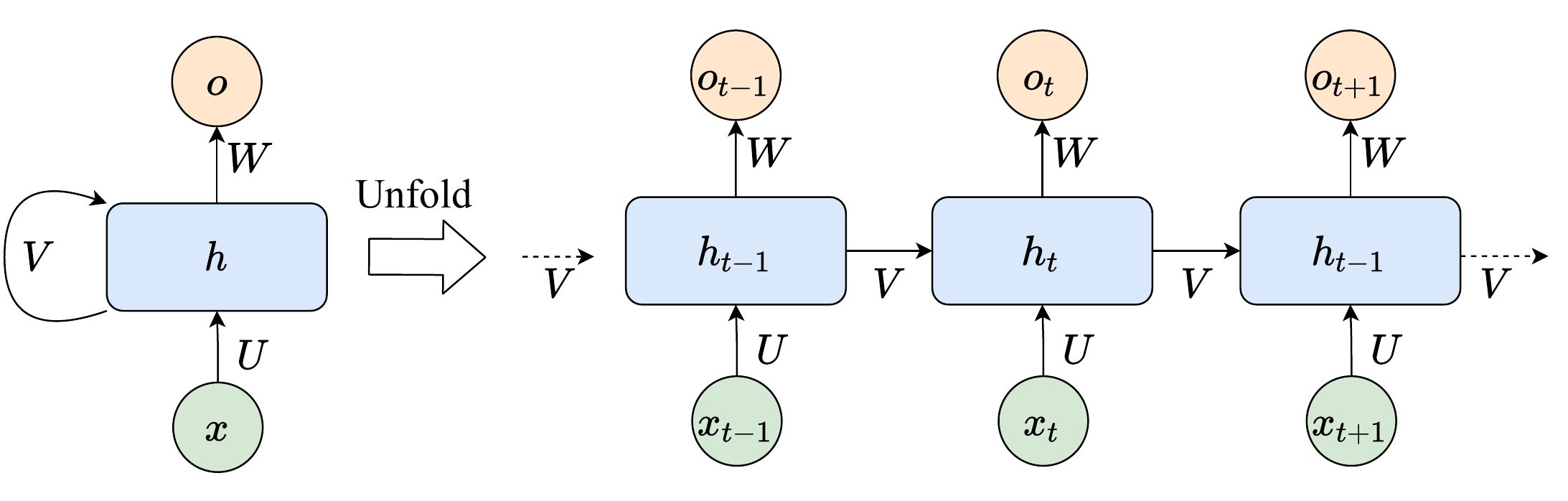}
    \caption{Generalized Recurrent Neural Network-based architecture for handwritten digit recognition}
    \label{fig:basicRNN}
\end{figure}

A neural network that is specialized for processing a sequence of data for tasks involving sequential inputs such as speech and language is referred to as a recurrent neural network (RNN). The output of a specific element in an RNN-based architecture is dependent not only on that element but also on the previous elements in the sequence, which together serve as a `memory' for the architecture as shown in Figure \ref{fig:basicRNN}.

Previous research works have proven CNN-based architectures to be very effective for image classification tasks. However, various authors reported that CNN-based architectures tend to misclassify specific Bengali digits where the shape and size are similar (such as `$\bnone$' and `$\bnnine$') \cite{0037ahmed2019recognizing, 0022basri2020bangla, 0182hye2020extraction}. To improve the overall performance, the efficiency of RNN-based architectures has also been explored for its ability to learn contextual information through recurrent connections. Specifically, a variation of RNN named Long Short Term Memory (LSTM) based networks is mostly used in this specific task. Unlike RNN and other DL methods which use gradient-based learning, LSTM is proven to be less prone to the vanishing gradient problem which contributes to the overall performance \cite{0533pascanu2013difficulty}, \cite{0534bayer2015learning}. Although LSTM is well-known in classifying time-series data, several authors have reported satisfactory performance of LSTM-based networks in this domain as well \cite{0559sun2016deepLSTM,0560wigington2017data}. These models have the ability to extract features utilizing handwriting strokes which can be useful to differentiate the numerals having similar shapes and sizes. A summary of the RNN-based architectures is provided in \tableautorefname~\ref{tab:RNNarchitecture}.

\begin{table*}[tb]
    \centering
    \caption{Performances of the Recurrent Neural Network-based Architectures}
    \label{tab:RNNarchitecture}
    \begin{tabular}{C{1.4cm} C{0.7cm} C{1.8cm} C{1.1cm} C{0.8cm} C{0.8cm} C{1.1cm} C{1.4cm} C{2cm} C{1.3cm}}
       \toprule
       \textbf{Reference} & \textbf{Year} & \textbf{LSTM Layers (Hidden Units)} & \textbf{Dropout} & \textbf{Epoch} & \textbf{Batch Size} & \textbf{Learning Rate} & \textbf{Optimizer} & \textbf{Loss Function}& \textbf{Accuracy (\%)} \\ \midrule
       \cite{0040ahmed2016handwritten} & 2016 & 2x (50, 100) & N/A & 460 & 50 & $-$ & $-$ & $-$& 98.25 \\
       \cite{0037ahmed2019recognizing} & 2019 & 2x (100, 50) & 0.2 & 370 & 50 & 0.02 & Adadelta & Categorical Cross-Entropy & 98.25\\
       \cite{0182hye2020extraction} & 2020 & 2x (50, 30) & N/A & 1500 & $-$ & $-$ & $-$ & $-$  & 98.03\\
       \bottomrule
       \multicolumn{10}{l}{\footnotesize $-$ denotes information was not available in the cited literature}\\
       \multicolumn{10}{l}{\footnotesize N/A denotes dropout layer was not used in the cited literature}\\
    \end{tabular}
\end{table*}

Reference \cite{0040ahmed2016handwritten} proposed one of the first works in BHDR using LSTM-based architectures. Their comparison showed that the proposed pipeline outperformed some pre-dominant architectures till that time, namely, SVM, MLP, CNN, etc., and achieved an accuracy of 98.25\% on the ISI-HBN dataset. The proposed architecture consisted of two LSTM layers with 50 and 100 hidden units respectively, with a feed-forward layer in between them. Both layers used Tanh activation functions. The LSTM layers were followed by a reshape layer and a dense layer. 
A subsequent work by the same authors also experimented with three configurations of LSTM architectures and showed that an LSTM with two layers performed better than the one having a single layer \cite{0037ahmed2019recognizing}. After experimenting with different choices of learning rates and batch sizes, the authors concluded that smaller values for these hyperparameters resulted in better performance. The smaller learning rate allowed the optimizer to take smaller steps to apply the gradients. On the other hand, the smaller batch size provided regularizing effect  \cite{0516bengio2012practical, 0514keskar2017largebatch, 0515masters2018revisiting}.

Although the intention of both the works was to demonstrate the superiority of LSTM-based architectures, several existing works using custom CNN-based architectures on the same dataset achieved a better performance \cite{0029akhand2016convolutional, 0038akhand2016convolutional, 0039akhand2016multiple}. 
Again, in addition to experimentation with single-layer and two-layer LSTMs in \cite{0037ahmed2019recognizing}, the authors could have tested how the models perform when the number of LSTM layers is increased. The impact on computational resources compared to other architectures could also have been explored.

In another work, the authors took a unique preprocessing approach that converted the input image into a directed acyclic graph with a fixed number of equidistant points \cite{0182hye2020extraction}. Then it was fed to an LSTM architecture containing two LSTM layers with 50 and 30 units respectively. 

The proposed architecture achieved 98.03\% accuracy on the ISI-HBN dataset, which was 2.58\% higher than the performance achieved by applying a single layer. Nevertheless, exploring the impact of LSTM-based architectures on reducing computation resources along with ensuring a comparable performance can be useful. On the other hand, the claim that LSTM can alleviate the difficulty of correctly classifying Bengali digits with similar shapes and sizes is not put to test in any of the existing literature. Future research can focus on these things as well.

\section{Broader Applications of Handwritten Digit Recognition}\label{sec:broader}
In recent times, some works have proposed handwritten digit recognition models with a broader scope than only classifying numerals. Such systems can be utilized to build useful tools focusing on different real-life applications. 

\subsection{Handwritten Mathematical Expression Evaluation}
Reference \cite{0169shuvo2021mathnet} proposed a CNN-based architecture named `MathNet' to recognize different components of handwritten mathematical equations containing Bengali digits. The authors trained the model using a dataset of 32,400 images consisting of 10 classes of Bengali handwritten digits and 44 other classes belonging to operators from algebra, set-symbols, limit, calculus, symbols of comparison, delimiters, etc. The dataset included 6,000 numeral images from the `Ekush' dataset and the symbol images were self-collected.
Although the model achieved high performance, the classes were assumed to be isolated and no segmentation was proposed, which leaves the opportunity for future improvements. Besides, the task could use more challenging images for numeral detection like `NumtaDB' instead of `Ekush' for digit images, which is more likely to represent real-world scenarios. 

This work was further extended in \cite{0177hasan2021bangla} by proposing a complete pipeline for handwritten Bengali mathematical expression recognition and simplification. It utilized an existing line and character segmentation technique and applied the `MathNet' model for classification. Nevertheless, the work is still in a very preliminary state and less likely to generalize in real-world examples due to certain strict assumptions by the authors. For example, the line segmentation technique assumes that there will be a certain amount of gap between two lines. But in a real-world scenario, the handwriting might not be in such an ideal form. 

However, this area has a long way to go compared to the present state of mathematical expression recognition on other scripts \cite{0276zhelezniakov2021online}. Advanced techniques like stroke extraction \cite{0277chan2020stroke} are found to be useful on the publicly available mathematical expression dataset CROHME \cite{0279mouchere2016icfhr}. Curating such benchmark datasets for Bengali handwritten mathematical expression recognition along with a robust pipeline might be useful for many practical applications.

\subsection{Handwritten Digit Generation}
Despite the existence of several works on handwritten character generation in recent times in different languages, the field of Bengali handwritten digit generation is still comparatively unexplored \cite{0542chang2018generating,  0543liu2019multi, 0544alonso2019adversarial}. \cite{0015sikder2020bangla} proposed a semi-supervised Generative Adversarial Network (SGAN) \cite{0562odena2016semisupervised} for Bengali handwritten digit generation. Both the generator and the discriminator blocks had a series of convolution and fully-connected layers. The model had a high loss of 0.368 and 0.694 in the discriminator and the generator, respectively.

Reference \cite{0048haque2019onkogan} trained a DCGAN-based architecture \cite{0563radford2015unsupervised} to generate Bengla handwritten digits from random noise. The authors experimented with four publicly available datasets: CMATERdb, BanglaLekha-Isolated, ISI-HBN, and Ekush for this task. Once the model was trained with a merged dataset of 58,827 images, it achieved a loss of 0.647. The final output was promising, but still, there is plenty of room for improvement.

One negative aspect of both works is that the generated datasets were not tested for digit recognition. Experiments should be designed where these generated images are used in the training phase of a digit recognition model to understand whether these systems are capable of generating adverse samples to improve the performance of the recognition models.

\subsection{Handwritten Digit Detection}
Reference \cite{0041tajrean2019handwritten} proposed a pipeline to detect Bengali handwritten digits using a region proposal network. An architecture combining Faster-RCNN \cite{0545ren2017faster} with InceptionV2 \cite{0546szegedy2015rethinking} was used for the detection and classification of digits. Due to the absence of any publicly available dataset suitable for this task, the authors generated a dataset using isolated images from the NumtaDB dataset to simulate the effect of a page with handwritten digits on it.

Despite the model achieving high performance, its effectiveness of this in real-life applications remains untested. The images in the generated dataset were placed randomly with the strict assumption that the digits should not be too densely packed, which might not be the case in real-world handwriting. Again, the images were taken from a comparatively less challenging portion of the NumtaDB dataset. Further experiments are needed to understand how it performs on images with a high degree of variability. Additionally, strict distance-based rules were applied to understand if two digits belong to the same number, which might not work in real-life scenarios.

\subsection{Multilingual Digit Recognition}
At present, most of the solutions are designed for single script numeral recognition. However, with the ever-growing international correspondence, especially in countries where multiple languages are being used in daily activities, the capability of a model to recognize multiple scripts is highly beneficial.

Reference \cite{0178haque2021quantitative} worked to recognize different Indic numerals in a single document page. The authors concentrated on the four most frequent scripts in the Indian subcontinent: Bengali, English, Hindi, and Oriya. Different state-of-the-art Deep CNN architectures were employed to solve the task, and Inception-v4 was the best performing model among them. However, the experiments were performed on individual datasets of each script, which conflicts with the goal of building a script invariant handwritten digit identification system.
Similarly, \cite{0181gupta2021cnnbased} proposed a CNN-based approach for multilingual numeral recognition, but the experiments were limited to applying the model separately to each dataset.
Future endeavors should concentrate on mixing samples from all the scripts to generate a combined dataset and then judge the performance of the model. Since different scripts have high inter-class similarities (such as Eight ($8$) in English and Four ($\bnfour$) in Bengali), it can pose a huge challenge for the models to recognize such occurrences. 

In this regard, \cite{0236singh2021anew} proposed a novel feature extraction technique named `Symbolization of binary images (SBI)' for the recognition of handwritten numerals in different scripts. The authors achieved above 95\% accuracy with the SVM classifier on each of the four popular scripts: Arabic, Bengali, Devanagari, and Latin. When one of the scripts was mixed pairwise with Latin, the performance was above 91\%. Finally, once all four scripts were mixed, the model still had a 90.98\% recognition rate, justifying its capability to recognize the numerals invariant of the script class. 
In another work, the authors introduced  the `Quadrangular Transition Count' feature extraction technique which resulted in satisfactory performance in script invariant handwritten digit recognition
 \cite{1004singh2018script}.

However, this domain of mixed numeral recognition has a long way to go since distinguishing these high inter-script similarities poses an extremely difficult challenge even for a human being. Using separate models for language detection and digit recognition has been proven to be useful in multi-script images \cite{0181fateh2021multilingual}. Such approaches can be adopted for mixed numeral detection of Bengali and other scripts. Furthermore, benchmark datasets containing mixed script handwritten documents can be proposed in this regard to provide baselines and make the performance of different models comparable \cite{1011singh2018benchmark}.

\subsection{Handwritten digit string recognition (HDSR)}
Recognizing handwritten digits from different strings is a challenging task since it has to deal with several complex scenarios such as the presence of noise, broken digits, digits touching each other, etc. In this regard, some recent works have introduced segmentation-free approaches \cite{0271neto2020hdsrflor,0272hochuli2021acomprehensive,  0273aly2019unknownlength}. Reference \cite{0210Kusetogullari2021digitnet} utilized different techniques of HDSR to detect and recognize handwritten English digits from historical handwritten document images. However, this field is completely unexplored in the context of Bengali handwritten string recognition. Developing such systems can be extremely useful in application domains like processing bank checks, postal codes, handwritten forms, etc. Application in these domains can help build assistive technology for visually impaired people \cite{1013sabab2016blind}.

\section{Research Gaps and Future Directions}\label{sec:futureworks}
Based on the experience of detailed literature review, some research gaps have been identified in this field, which requires intensive research to contribute more. Therefore, a list of future research directions is suggested below:
\begin{itemize}
    
    \item Most of the existing works assume the digits to be in isolated form. A complete pipeline is necessary to check how this can be applied to entire documents. Even though a few works in BHDR have focused on it, they are constrained by a number of challenges. To extract digits from a handwritten document, a well-designed line and digit segmentation algorithm can be proposed. Grouping the digits belonging to a single number is another challenge in this regard.
    
    \item Although lots of preprocessing techniques have been proposed in the existing literature, they are mostly considered localized solutions for specific datasets. These techniques might not hold up against real-life samples if they contain different types of noises, background clutter, occlusion, etc.  For this reason, instead of preprocessing the dataset, it would be better to focus on augmentation to ensure that the models learn to recognize digits in various scenarios.
    
    \item Despite the main focus of most of the works have been on improving the accuracy, lightweight models can be proposed that are suitable for low-powered devices such as smartphones and embedded systems. An in-depth analysis of training time and inference time for such models is yet to be explored. Such analysis of the time complexities of different methodologies can be extremely useful while picking the suitable model satisfying the resource constraints across different applications.


    \item Instead of focusing on digit classification only, future research efforts can be concentrated on extracting age, gender, location, educational status, etc. information from handwritten digits. These endeavors can be facilitated by various modalities provided in Ekush and BanglaLekha-Isolated datasets. Extraction of such information can facilitate forensic investigation of different handwritings.
    
    \item Most of the existing literature in BHDR has not considered a detailed ablation study of the proposed pipeline. An in-depth analysis of the contribution of different components of the model to the overall result may not only shed light on the pros and cons of the proposed model but also provide insights for future researchers to work on. Furthermore, proper benchmarking attempts are required to compare the models with previous literature. The heterogeneity of reported results makes it difficult to position a model with respect to earlier works.
    
    \item To further understand the generalization capability of the models, one approach can be training the model on one dataset and testing it on another. Few such works can be seen in the existing literature. The merit of this experiment is that samples of the same dataset can have similar biases within them. Even when the test set remains completely unseen, the model can get a sense of the variety of the images in the training phase, which can affect the classification accuracy. Validating the model on a completely different dataset can alleviate this issue.
    
    \item Several existing works on BHDR utilize multiple classifiers to provide a comparative understanding of their performance of them with a view to identifying the best one. However, while working with multiple datasets across different classifiers, it is difficult to provide a generalized comparison. To address the issue, various parametric and non-parametric statistical analyses can be performed to standardize the experimental results of the classifiers which can provide better insight into their performance \cite{1001singh2016significance, 1002singh2015statistical}.
    
    \item With the rapid growth in the usage of smart devices, storing data from online handwriting has become a common phenomenon. To convert this written form of information into editable electronic content, Online Handwritten Recognition has become an active area of research in recent times \cite{0215singh2021online, 1009sen2018appliaction}. Although several research works have been done on several scripts such as, Arabic \cite{0104alhelali2017arabic}, Chinese \cite{1010ren2019recognizing}, Devanagari \cite{1010ghosh2015novel}, etc, it is yet to be explored with Bengali numeral. Research attempts can be made to propose such online handwritten digit recognition systems suitable for end-level smart devices.
    
    \item One of the key challenges in ensuring the adoption of the deep learning-based BHDR pipelines in various applications is the lack of transparency \cite{1006samek2019towards}. In the case of the machine learning-based models, the use of handcrafted features usually provided better visualization of what worked and what did not. On the other hand, DL-based works on BHDR often treat the model architecture as a ``black boxes''. In this regard, adopting recent trends in explainable artificial intelligence can provide better insight into how the models identify the digits. This interpretability of the models can ensure effective performance \cite{1008gunning2019xai} and improve the user experience by helping the end users to trust the decision-making process of the BHDR pipelines \cite{1007alizadeh2001dont}.
    
    \item Research attempts can be concentrated on generating handwritten digits, which can add value to the overall performance of the classifier. On this note, in recent times, there has been a trend in proposing models having fewer dependencies on very large datasets \cite{0548wang2020generalizing}. The usability of different few-shot learning-based approaches can be explored in the field of Bengali handwritten digit recognition \cite{0547dhillon2019abaseline, 0549souibgui2021few}.
    
    \item In recent times, some broader applications of BHDR such as mathematical expression recognition, numeral recognition with a mixture of digits from multiple scripts, detecting digits from handwritten strings, etc., have gained interest. The status quo of such works in the field of BHDR is still in a preliminary state. Such application fields can further be explored by curating publicly available benchmark datasets, preparing baselines, and proposing robust models for each of such use cases.
    
\end{itemize}

\section{Conclusion}\label{sec:future}

In this paper, the characteristics and inherent ambiguities of Bengali handwritten digits along with comprehensive insights on the state-of-the-art works that are proposed in the last two decades have been analyzed. Even though a lot of work has been done in this domain, addressing these ambiguities still remains a major challenge for future researchers. In this regard, the use of contextual information such as texts, images, etc. surrounding the digits to be recognized might be an interesting research avenue. The discussion regarding the available benchmark datasets highlights the usefulness of different modalities such as gender, age, aesthetic quality index, etc. available with the dataset in various biometric applications. Again, utilization of these benchmark datasets in future works can provide better insight into the comparative performance of different models. Furthermore, the assessment of the preprocessing and augmentation techniques used over the years indicates heavy reliability in targeted preprocessing resulting in BHDR pipelines with reduced generalization capabilities. This issue requisites a shift from preprocessing to augmentation that can result in robust pipelines capable of performing well in unknown scenarios.

The last two decades have seen a huge shift in the recognition process transforming from classical machine learning-based approaches to the deep learning-based techniques. Despite the fact that deep learning-based approaches helped get rid of cumbersome handcrafted feature-based approaches increasing the generalization capability of the recognition pipelines, there are still a lot to explore in terms of model architectures and hyperparameters of the deep learning-based approaches. Careful consideration should be provided to the overall training and testing process to better analyze the performance of the models. In addition to exploring the newer and better architectures, exploitation of the existing ones are also necessary to harness their full potential.


Finally, the use of BHDR pipelines in real-life scenarios, such as mathematical expression evaluation, multilingual digit, recognition, handwritten digit string recognition, etc. can further empower seamless connection and cooperation between the physical and the digital world. To facilitate that, this paper will serve the purpose of a compendium for researchers who are interested in the science behind offline BHDR, instigating the exploration of newer avenues of relevant research, which may lead to better recognition of offline Bengali handwritten digits in different application areas.


\section*{Acknowledgment}
The authors are thankful to Mohammad Ishrak Abedin, Department of Computer Science and Engineering, Islamic University of Technology for his valuable time and support in proofreading the article.

\section*{Conflict of Interest}
The authors declare that there is no conflict of interest.

\bibliographystyle{IEEEtran}
\bibliography{ref}

\begin{IEEEbiography}[{\includegraphics[width=1in,height=1.25in,clip,keepaspectratio]{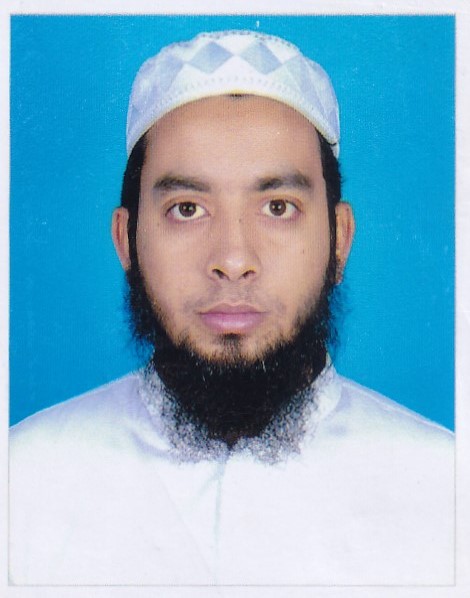}}]{A.B.M. Ashikur Rahman} completed his B.Sc. in Engineering degree in Computer Science and Engineering (CSE) from Islamic University of Technology (IUT) in 2014 and M.Sc. form the same institution in 2018.

He is currently working as an Assistant Professor in the Department of Computer Science and Engineering, IUT. His research interest includes the application of computer vision techniques in medical image analysis and human biometrics.
\end{IEEEbiography}

\begin{IEEEbiography}[{\includegraphics[width=1in,height=1.25in,clip,keepaspectratio]{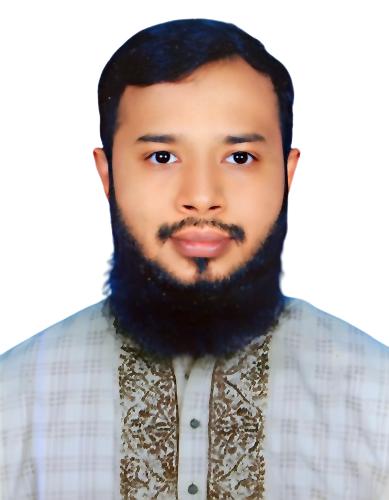}}]{Md. Bakhtiar Hasan} completed his B.Sc. Engg. and M.Sc. Engg. degree in Computer Science and Engineering (CSE) from Islamic University of Technology (IUT) in 2018 and 2022, respectively.

Since 2019, he has been working as a Lecturer in the Department of Computer Science and Engineering, IUT. His research interest includes the use of deep learning and computer vision techniques in human biometrics and smart agriculture.

Mr. Hasan received Huawei Seeds for the Future scholarship in 2018. He was awarded IUT Gold Medal in recognition of his outstanding performance in the pursuit of the B.Sc. Engg. in CSE degree in 2018.
\end{IEEEbiography}

\begin{IEEEbiography}[{\includegraphics[width=1in,height=1.25in,clip,keepaspectratio]{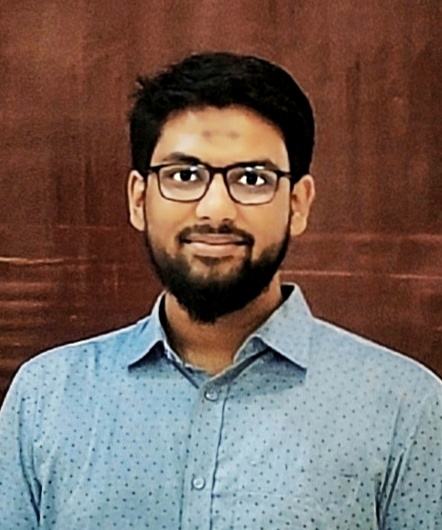}}]{Sabbir Ahmed} was born in Dhaka, Bangladesh, in 1996. He graduated from Islamic University of Technology (IUT), Gazipur, Bangladesh, in 2017 with a B.Sc. Engg. degree (IUT Gold Medalist) in Computer Science (CS) and is pursuing his M.Sc. degree in CS from the same institution.

Since 2018, he has been working as a Lecturer in the Department of Computer Science and Engineering, IUT. His research interests include pattern recognition, deep learning in computer vision, and intelligent agriculture.
\end{IEEEbiography}

\begin{IEEEbiography}[{\includegraphics[width=1in,height=1.25in,clip,keepaspectratio]{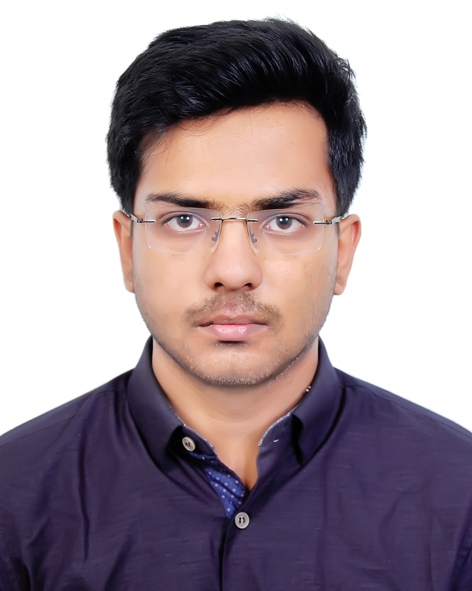}}]{Tasnim Ahmed} was born in Kushtia, Bangladesh in 1997. He received his B.Sc degree in Computer Science and Engineering from the Islamic University of Technology, Gazipur, Bangladesh and currently he is pursuing the M.Sc degree.

Since 2020, he has been working as a full-time Lecturer with the Computer Science and Engineering Department, Islamic University of Technology, Gazipur, Bangladesh. His research interests include computer vision, natural language processing, bioinformatics, and software engineering.
\end{IEEEbiography}

\begin{IEEEbiography}[{\includegraphics[width=1in,height=1.25in,clip,keepaspectratio]{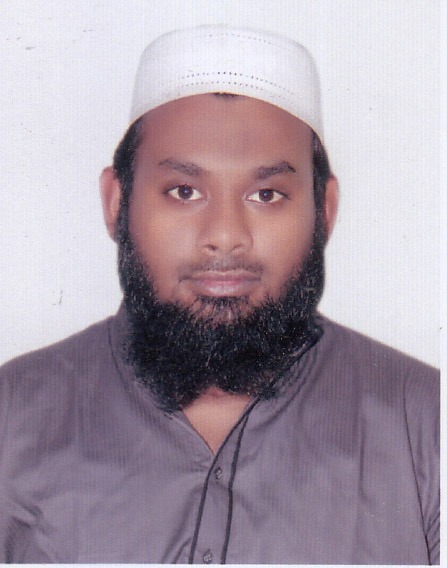}}]{Md. Hamjajul Ashmafee} received his B.Sc. Engg. and M.Sc. Engg. in Computer Science and Engineering (CSE) degree from Islamic University of Technology (IUT) in 2015 and 2022, respectively. 

Since 2016, he has been working as a full-time Lecturer with the Computer Science and Engineering Department, Islamic University of Technology. His research interests include data analytics and machine learning.
\end{IEEEbiography}

\begin{IEEEbiography}[{\includegraphics[width=1in,height=1.25in,clip,keepaspectratio]{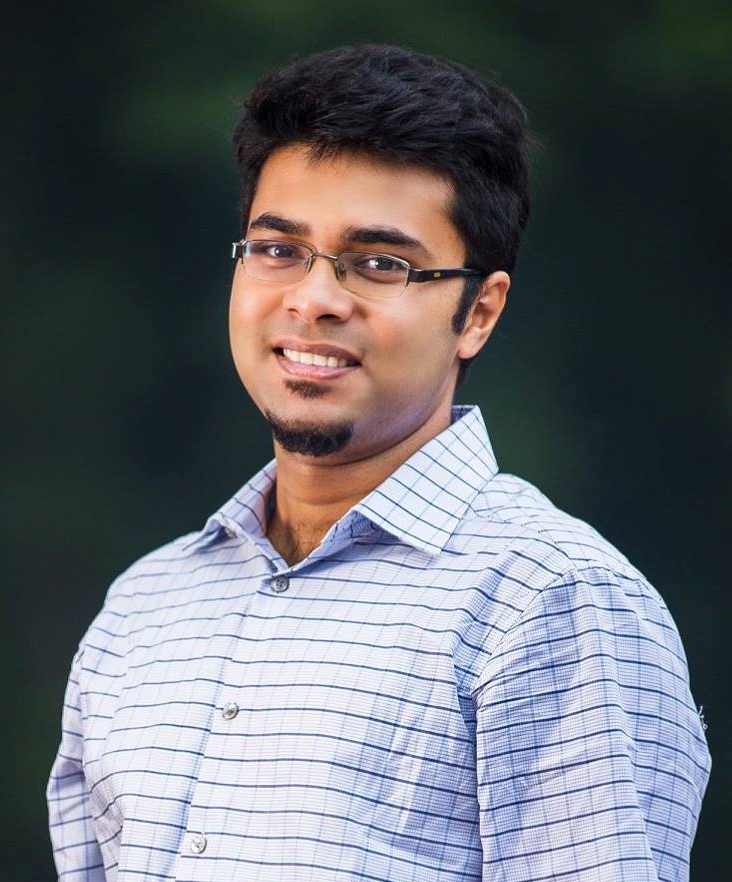}}]{Mohammad Ridwan Kabir} received the B.Sc. Engg. and M.Sc. Engg. degree in computer science and engineering from the Islamic University of Technology (IUT), Boardbazar, Gazipur, Bangladesh, in 2017 and 2022, respectively.

Since 2018, he has been working as a Lecturer with the Department of Computer Science and Engineering, IUT. He worked as a Lecturer with the Department of Computer Science Engineering, BRAC University, Dhaka, Bangladesh. His research interests include human–computer interaction, computer vision, assistive technology, machine learning, data analysis and visualization, embedded system development, and wearable devices.

Mr. Kabir and his team received the Runners Up title (Project Showcasing) at the National ICT Fest, in 2016, the Champions title (Project Showcasing) at Esonance, in 2017, the Top 5 Innovative Projects awards at BASIS Soft Expo, Bangladesh, in 2020, and the BASIS National ICT Awards, Bangladesh, in 2021. Furthermore, he has received the WINNERS title (Research and Development) in the prestigious APICTA Awards 2020–2021 (host country: Malaysia) from Bangladesh.
\end{IEEEbiography}

\begin{IEEEbiography}[{\includegraphics[width=1in,height=1.25in,clip,keepaspectratio]{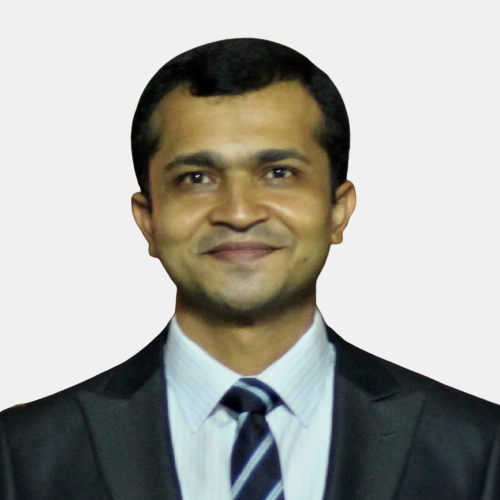}}]{Md. Hasanul Kabir} (M’17) received the B.Sc. degree in computer science and information technology from the Islamic University of Technology, Bangladesh, and the Ph.D. degree in computer engineering from Kyung Hee University, South Korea.

He is currently a Professor in the Department of Computer Science and Engineering, Islamic University of Technology. His research interests include feature extraction, visual question answering, and sign language interpretation by combining image processing, machine learning, and computer vision.
\end{IEEEbiography}

\EOD

\end{document}